\documentclass[journal]{IEEEtran}

\usepackage[OT1]{fontenc} 
\usepackage{cite}
\usepackage{array}
\usepackage{textcomp}
\usepackage{color}
\usepackage{colortbl}
\usepackage{diagbox} 
\usepackage{graphicx}
\usepackage{amsmath}
\usepackage{algorithmic}
\usepackage{subfigure}
\usepackage{stfloats}
\usepackage{url}
\usepackage{multirow}
\usepackage{multicol}
\usepackage{amssymb}
\usepackage{booktabs}
\usepackage{arydshln}
\usepackage{slashbox}
\usepackage{makecell}
\usepackage{rotating}
\usepackage{threeparttable}
\usepackage[colorlinks=true,linkcolor=blue]{hyperref}
\usepackage{pifont}
\usepackage{amssymb}
\usepackage{mathrsfs}

\definecolor{r00}{RGB}{255, 255, 255}
\definecolor{r01}{RGB}{255, 253, 253}
\definecolor{r02}{RGB}{255, 252, 252}
\definecolor{r03}{RGB}{255, 251, 251}
\definecolor{r04}{RGB}{255, 250, 250}
\definecolor{r05}{RGB}{255, 249, 249}
\definecolor{r06}{RGB}{255, 247, 247}
\definecolor{r07}{RGB}{255, 246, 246}
\definecolor{r08}{RGB}{255, 245, 245}
\definecolor{r09}{RGB}{255, 244, 244}
\definecolor{r10}{RGB}{255, 243, 243}
\definecolor{r11}{RGB}{255, 241, 241}
\definecolor{r12}{RGB}{255, 240, 240}
\definecolor{r13}{RGB}{255, 239, 239}
\definecolor{r14}{RGB}{255, 238, 238}
\definecolor{r15}{RGB}{255, 237, 237}
\definecolor{r16}{RGB}{255, 235, 235}
\definecolor{r17}{RGB}{255, 234, 234}
\definecolor{r18}{RGB}{255, 233, 233}
\definecolor{r19}{RGB}{255, 232, 232}
\definecolor{r20}{RGB}{255, 231, 231}
\definecolor{r21}{RGB}{255, 229, 229}
\definecolor{r22}{RGB}{255, 228, 228}
\definecolor{r23}{RGB}{255, 227, 227}
\definecolor{r24}{RGB}{255, 226, 226}
\definecolor{r25}{RGB}{255, 225, 225}
\definecolor{r26}{RGB}{255, 223, 223}
\definecolor{r27}{RGB}{255, 222, 222}
\definecolor{r28}{RGB}{255, 221, 221}
\definecolor{r29}{RGB}{255, 220, 220}
\definecolor{r30}{RGB}{255, 219, 219}
\definecolor{r31}{RGB}{255, 217, 217}
\definecolor{r32}{RGB}{255, 216, 216}
\definecolor{r33}{RGB}{255, 215, 215}
\definecolor{r34}{RGB}{255, 214, 214}
\definecolor{r35}{RGB}{255, 213, 213}
\definecolor{r36}{RGB}{255, 211, 211}
\definecolor{r37}{RGB}{255, 210, 210}
\definecolor{r38}{RGB}{255, 209, 209}
\definecolor{r39}{RGB}{255, 208, 208}
\definecolor{r40}{RGB}{255, 207, 207}
\definecolor{r41}{RGB}{255, 205, 205}
\definecolor{r42}{RGB}{255, 204, 204}
\definecolor{r43}{RGB}{255, 203, 203}
\definecolor{r44}{RGB}{255, 202, 202}
\definecolor{r45}{RGB}{255, 201, 201}
\definecolor{r46}{RGB}{255, 199, 199}
\definecolor{r47}{RGB}{255, 198, 198}
\definecolor{r48}{RGB}{255, 197, 197}
\definecolor{r49}{RGB}{255, 196, 196}
\definecolor{r50}{RGB}{255, 195, 195}
\definecolor{r51}{RGB}{255, 193, 193}
\definecolor{r52}{RGB}{255, 192, 192}
\definecolor{r53}{RGB}{255, 191, 191}
\definecolor{r54}{RGB}{255, 190, 190}
\definecolor{r55}{RGB}{255, 189, 189}
\definecolor{r56}{RGB}{255, 187, 187}
\definecolor{r57}{RGB}{255, 186, 186}
\definecolor{r58}{RGB}{255, 185, 185}
\definecolor{r59}{RGB}{255, 184, 184}
\definecolor{r60}{RGB}{255, 183, 183}
\definecolor{r61}{RGB}{255, 181, 181}
\definecolor{r62}{RGB}{255, 180, 180}
\definecolor{r63}{RGB}{255, 179, 179}
\definecolor{r64}{RGB}{255, 178, 178}
\definecolor{r65}{RGB}{255, 177, 177}
\definecolor{r66}{RGB}{255, 175, 175}
\definecolor{r67}{RGB}{255, 174, 174}
\definecolor{r68}{RGB}{255, 173, 173}
\definecolor{r69}{RGB}{255, 172, 172}
\definecolor{r70}{RGB}{255, 171, 171}
\definecolor{r71}{RGB}{255, 169, 169}
\definecolor{r72}{RGB}{255, 168, 168}
\definecolor{r73}{RGB}{255, 167, 167}
\definecolor{r74}{RGB}{255, 166, 166}
\definecolor{r75}{RGB}{255, 165, 165}
\definecolor{r76}{RGB}{255, 163, 163}
\definecolor{r77}{RGB}{255, 162, 162}
\definecolor{r78}{RGB}{255, 161, 161}
\definecolor{r79}{RGB}{255, 160, 160}
\definecolor{r80}{RGB}{255, 159, 159}
\definecolor{r81}{RGB}{255, 157, 157}
\definecolor{r82}{RGB}{255, 156, 156}
\definecolor{r83}{RGB}{255, 155, 155}
\definecolor{r84}{RGB}{255, 154, 154}
\definecolor{r85}{RGB}{255, 153, 153}
\definecolor{r86}{RGB}{255, 151, 151}
\definecolor{r87}{RGB}{255, 150, 150}
\definecolor{r88}{RGB}{255, 149, 149}
\definecolor{r89}{RGB}{255, 148, 148}
\definecolor{r90}{RGB}{255, 147, 147}
\definecolor{r91}{RGB}{255, 145, 145}
\definecolor{r92}{RGB}{255, 144, 144}
\definecolor{r93}{RGB}{255, 143, 143}
\definecolor{r94}{RGB}{255, 142, 142}
\definecolor{r95}{RGB}{255, 141, 141}
\definecolor{r96}{RGB}{255, 139, 139}
\definecolor{r97}{RGB}{255, 138, 138}
\definecolor{r98}{RGB}{255, 137, 137}
\definecolor{r99}{RGB}{255, 136, 136}

\definecolor{red}{RGB}{0,0,0}

\begin{document}
\title{\LARGE \bf
	Are We Hungry for 3D LiDAR Data for Semantic Segmentation?\\ \textcolor{red}{A Survey and Experimental Study}}
%
%
%

\author{Biao~Gao,~\IEEEmembership{Member,~IEEE,}
		Yancheng~Pan,~\IEEEmembership{Member,~IEEE,}
		Chengkun~Li,~\IEEEmembership{Member,~IEEE,}
		Sibo~Geng,~\IEEEmembership{Member,~IEEE,}
		Huijing~Zhao,~\IEEEmembership{Member,~IEEE,}
	
\thanks{B.Gao and Y.Pan are both the first authors of this paper. This work was supported in part by the National Natural Science Foundation of China under Grant 61973004 and in part by the Development Program of China under Grant 2017YFB1002601. B.Gao, Y.Pan, C.Li, S.Geng and H. Zhao are with the Key Lab of Machine Perception (MOE), Peking University, Beijing, China. Contact: H.Zhao, zhaohj@cis.pku.edu.cn.}

}

\maketitle

\begin{abstract}
	
\textcolor{red}{
3D semantic segmentation is a fundamental task for robotic and autonomous driving applications. Recent works have been focused on using deep learning techniques, whereas developing fine-annotated 3D LiDAR datasets is extremely labor intensive and requires professional skills. The performance limitation caused by insufficient datasets is called data hunger problem. This research provides a comprehensive survey and experimental study on the question: are we hungry for 3D LiDAR data for semantic segmentation? The studies are conducted at three levels. First, a broad review to the main 3D LiDAR datasets is conducted, followed by a statistical analysis on three representative datasets to gain an in-depth view on the datasets' size and diversity, which are the critical factors in learning deep models. Second, a systematic review to the state-of-the-art 3D semantic segmentation is conducted, followed by experiments and cross examinations of three representative deep learning methods to find out how the size and diversity of the datasets affect deep models' performance. Finally, a systematic survey to the existing efforts to solve the data hunger problem is conducted on both methodological and dataset's viewpoints, followed by an insightful discussion of remaining problems and open questions To the best of our knowledge, this is the first work to analyze the data hunger problem for 3D semantic segmentation using deep learning techniques that are addressed in the literature review, statistical analysis, and cross-dataset and cross-algorithm experiments. We share findings and discussions, which may lead to potential topics in future works. }

\end{abstract}


%

\section{Introduction}
%

\IEEEPARstart{T}{oday}, LiDAR has become the main sensor in many robotic \cite{thrun2006stanley} \cite{patz2008practical}, mobile mapping \cite{zhang2014loam} \cite{hess2016real} and autonomous driving \cite{li2016vehicle} \cite{chen2017multi} systems. 3D LiDAR data, captured from either a static viewpoint \cite{hackel2017semantic3d} or a mobile platform \cite{behley2019semantickitti} during a dynamic procedure, provide a copy of the real world with rich 3D geometry in true size, which can be represented in the format of either 3D point clouds \cite{rusu20113d} \cite{qi2017pointnet} or 2D grids \cite{wu2018squeezeseg}, e.g., range image, using a static or a sequence of data frames. Semantic segmentation \cite{garcia2017review} \cite{yu2018methods} is a fundamental task of scene understanding, which divides a whole piece of input data into different semantically interpretable categories according to a meaningful taxonomy in the real world. With the widespread use of LiDAR sensors in various applications, semantic segmentation of 3D LiDAR data \cite{yuxing2019review} \cite{yulan2019review} is attracting increasing attention. Hereinafter, we refer to {\bf 3D semantic segmentation} to emphasize the works addressing the features of 3D LiDAR data, and {\bf semantic segmentation} for those of potentially general purpose.

Semantic segmentation has been studied for decades. A comprehensive review of early works up to 2014 is given in \cite{zhu2016beyond}. We refer to these works as {\bf traditional methods}, which are characterized by using handcrafted features and bottom-up procedures. Inspired by the amazing success of deep learning techniques \cite{lecun2015deep} \cite{schmidhuber2015deep}, recent semantic segmentation works have focused on using deep neural networks to learn a richer feature representation, and model the mapping from input data to semantic labels in an end-to-end procedure \cite{long2015fully}, which are referred to as {\bf deep learning methods} hereinafter. However, compared to traditional methods, deep learning methods face a considerable challenge of requiring large quantities of manually labeled data in training \cite{chen2014big}. The quantity, quality and diversity of training data have a considerable influence on the generalization performance of deep learning models \cite{torralba2011unbiased} \cite{sun2017revisiting}.

The performance limitation caused by insufficient training data is called the {\bf data hunger} effect, \textcolor{red}{which is reflected on both data size and diversity.} As noted by G.Marcus in \cite{marcus2018deep}, against the background of considerable progress and enthusiasm, the data hunger problem was his first concern among the ten challenges faced by the current deep learning systems. For 3D semantic segmentation tasks, 3D LiDAR data with point-wise annotation are required, where S3DIS \cite{armeni20163d}, Semantic3D \cite{hackel2017semantic3d}, and SemanticKITTI \cite{behley2019semantickitti} are among the most popular datasets. These datasets are annotated fully or partially by human operators, which is time consuming, human intensive, and requires special skill and software, e.g., the operators are trained to handle professional software to visualize and annotate 3D point clouds, which are much harder to interpret than 2D images. \textcolor{red}{Due to these difficulties, the publicly available datasets for 3D semantic segmentation are very limited in both data size and diversity compared with those of 2D images \cite{everingham2015pascal} \cite{deng2009imagenet}. Therefore, 3D semantic segmentation may face even severe data hunger problem.}

In this research, we seek to answer the following questions. Are we hungry for 3D LiDAR data for semantic segmentation using deep learning techniques? \textcolor{red}{Further, how serious is the problem on the aspect of both data size and diversity? What impact will the problem have on training deep 3D semantic segmentation models? What measures could be taken to solve the problem on the aspects of both methodology and dataset developing, and what are the remaining questions that need to be answered in future studies?}

To answer the questions, the following steps are taken in this work. \textcolor{red}{Section \ref{sec:2} reviews the existing 3D LiDAR datasets, and through statistical analysis on three representative datasets, an in-depth view of data size and diversity is gained. Section \ref{sec:3} reviews the existing methods of 3D semantic segmentation, through which, three representative methods using deep learning methods are selected in Section \ref{sec:4}, and experiments and cross examinations are conducted by training and testing on the datasets to find how datasets influence model performance. Section \ref{sec:5} reviews the efforts that have been conducted or potentially could be used to solve the data hunger problem of 3D semantic segmentation, followed by discussion on future topics and open questions in Section \ref{sec:6}.}


A number of surveys are relevant to this work. \cite{nguyen20133d} \cite{grilli2017review} review early methods of 3D point cloud segmentation and classification in the literature. \cite{garcia2017review} \cite{yu2018methods} \cite{lateef2019survey} review methods and datasets using for semantic segmentation. Furthermore, \cite{vodrahalli20173d} \cite{ioannidou2017deep} \cite{yuxing2019review} \cite{griffiths2019review} \cite{yulan2019review} \cite{Saifullahi2020review} review the deep learning methods for 3D semantic segmentation task. In addition, \cite{feng2020deep} reviews multi-modal methods used for semantic segmentation and detection. However, these surveys focused on summarizing and classifying the existing methods, and none of them emphasize 3D datasets or the data hunger problem. To the best of our knowledge, this is the first work to provide an in-depth survey and experimental study on the data hunger problem for 3D semantic segmentation using deep learning techniques.
The main contributions of our work are as follows:

\begin{figure*}[b]
	\centering
	\includegraphics[scale=0.24]{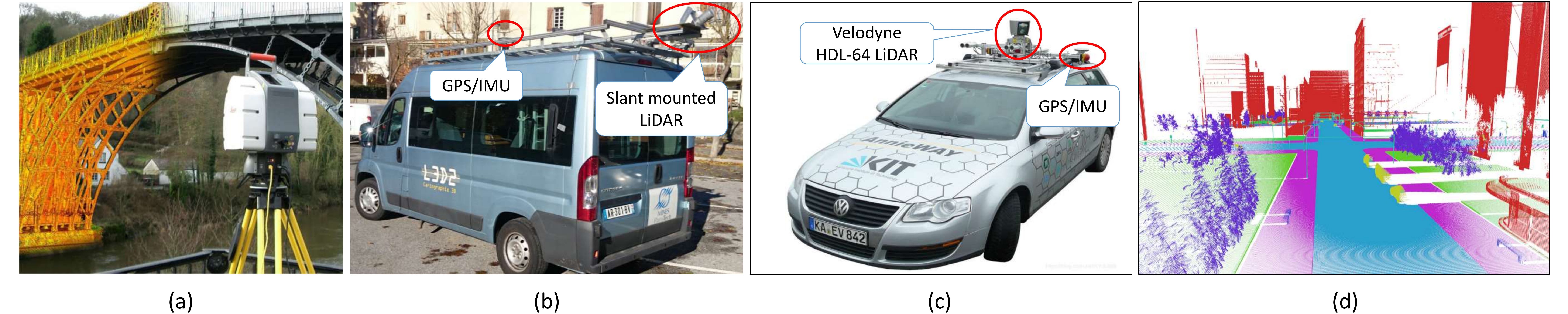}
	\caption{Typical 3D LiDAR acquisition systems. (a) a terrestrial laser scanner, collects data from static viewpoints for \textit{static datasets}, (b) a MLS (mobile laser scanning) system \cite{roynard2018paris}, collects data of mainly static scene objects for \textit{static datasets}, (c) an autonomous driving system \cite{geiger2012we}, collects 3D LiDAR streams for \textit{sequential datasets}, (d) a simulation system \cite{griffiths2019synthcity} for \textit{synthetic datasets}.}
	\label{fig:1}
\end{figure*}

\begin{itemize}
	\item A broad review of the existing 3D datasets is provided that are divided into static, sequential and synthetic datasets according to the data acquisition methods and their main applications, and an organized survey of 3D semantic segmentation methods is given with a focus on the latest research trend using deep learning techniques.
	\item An in-depth view to the data hunger problem on the aspects of data size and diversity is gained through statistical analysis on three representative 3D datasets, and \textcolor{red}{the impact of the problem on the performance of deep learning models is studied through experiments and cross examinations using three representative semantic segmentation methods.}
	\item A systematic survey of the efforts to solve the data hunger problem is given on the aspects of both methodologies that require less on fine annotated data, and data annotation methods that are less labor intensive. \textcolor{red}{An insightful discussion of remaining problems and open questions is given, leading to potential topics in future works.}
\end{itemize}


\section{3D LiDAR Datasets and Statistical Analysis}	\label{sec:2}

\begin{table*}[t]
	\centering
	\setlength{\tabcolsep}{0.4mm}
	\renewcommand\arraystretch{1.2}
	\caption{3D LIDAR datasets with comparison to representative IMAGE AND RGB-D ones*}
	\label{tab:1}
	\begin{threeparttable}
		\begin{tabular}{c|c|c|ccccccccccc} 
			\hline
			sensor                                        & data type                    & anno.                   & dataset                                                        & frames                   & points/pixels            & classes               & scene$^1$               & ins.                            & seq.                            & \textcolor{red}{sensor type}$^2$ & year                    & organization                                & \textcolor{red}{license$^3$}   \\ 
			\hline
			\multirow{21}{*}{\rotatebox{90}{3D LiDAR} }   & \multirow{7}{*}{static}      & \multirow{7}{*}{point}  & Oakland\cite{munoz2009contextual}                             & 17                       & 1.6M                     & 44                    & o                    & $\times$                        & $\times$                        & \textcolor{red}{SL}          & 2009                    & CMU                                         & \textcolor{red}{RO}                    \\
			&                              &                         & Paris-rue-Madame\cite{serna2014paris}                         & 2                        & 20M                      & 17                    & o                    & \checkmark                    & $\times$                        & \textcolor{red}{V32}         & 2014                    & MINES ParisTech                             & \textcolor{red}{ND-3.0 }               \\
			&                              &                         & TerraMobilita/IQmulus\cite{vallet2015terramobilita}           & 10                       & 12M                      & 15                    & o                    & \checkmark                    & $\times$                        & \textcolor{red}{MLS}         & 2015                    & Univ. of Paris-Est                          & \textcolor{red}{ND-3.0 }               \\
			&                              &                         & S3DIS \cite{armeni20163d}                                     & 5                        & 215M                     & 12                    & i                    & $\times$                        & $\times$                        & \textcolor{red}{MS}          & 2016                    & Stanford Univ.  & \textcolor{red}{/ }                    \\
			&                              &                         & \textcolor{red}{TUM City Campus$^4$\cite{gehrung2017approach}}  & \textcolor{red}{631 }    & \textcolor{red}{41M }    & \textcolor{red}{8 }   & \textcolor{red}{o }  & \textcolor{red}{\checkmark }  & \textcolor{red}{\checkmark }  & \textcolor{red}{V64*2}       & \textcolor{red}{2016 }  & \textcolor{red}{TUM }                       & \textcolor{red}{SA-4.0 }               \\
				&                              &                         & Semantic3D\cite{hackel2017semantic3d}                         & 30                       & 4009M                    & 8                     & o                    & $\times$                        & $\times$                        & \textcolor{red}{TLS}         & 2017                    & ETH Zurich                                  & \textcolor{red}{SA-3.0}                \\
				&                              &                         & Paris-lille-3D \cite{roynard2018paris}                        & 3                        & 143M                     & 50                    & o                    & \checkmark                    & $\times$                        & \textcolor{red}{V32}         & 2018                    & MINES ParisTech                             & \textcolor{red}{ND-3.0}                \\ 
				\cline{2-14}
				& \multirow{12}{*}{sequential} & \multirow{4}{*}{point}  & Sydney Urban \cite{de2013unsupervised}                        & 631                      & /                        & 26                    & o                    & \checkmark                    & \checkmark                    & \textcolor{red}{V64}         & 2013                    & ACFR                                        & \textcolor{red}{/ }                    \\
				&                              &                         & SemanticKITTI\cite{behley2019semantickitti}                   & 43552                    & 4549M                    & 28                    & o                    & \checkmark                    & \checkmark                    & \textcolor{red}{V64}         & 2019                    & Univ. of Bonn                               & \textcolor{red}{SA-3.0 }               \\
				&                              &                         & SemanticPOSS\cite{pan2020semanticposs}                        & 2988                     & 216M                     & 14                    & o                    & \checkmark                    & \checkmark                    & \textcolor{red}{HP}          & 2020                    & Peking Univ.                                & \textcolor{red}{SA-3.0 }               \\
				&                              &                         & \textcolor{red}{A2D2\cite{geyer2020a2d2} }                    & \textcolor{red}{41277 }  & \textcolor{red}{1238M }  & \textcolor{red}{38 }  & \textcolor{red}{o }  & \textcolor{red}{\checkmark }  & \textcolor{red}{\checkmark }  & \textcolor{red}{V16*5}       & \textcolor{red}{2020 }  & \textcolor{red}{Audi }                      & \textcolor{red}{ND-4.0 }               \\ 
				\cline{3-14}
				&                              & \multirow{8}{*}{3D-box} & KITTI\cite{geiger2012we}                                      & 14999                    & 1799M                    & 8                     & o                    & \checkmark                    & \checkmark                    & \textcolor{red}{V64}         & 2012                    & KIT                                         & \textcolor{red}{SA-3.0 }               \\
				&                              &                         & H3D \cite{patil2019h3d}                                       & 27K                      & /                        & 8                     & o                    & \checkmark                    & \checkmark                    & \textcolor{red}{V64}         & 2019                    & HRI                                         & \textcolor{red}{RO}                    \\
				&                              &                         & nuScenes\cite{caesar2019nuscenes}                             & 40K                      & 2780M                    & 23                    & o                    & \checkmark                    & \checkmark                    & \textcolor{red}{U32}         & 2019                    & nuTonomy                                    & \textcolor{red}{NC}                    \\
				&                              &                         & \textcolor{red}{Lyft L5\cite{lyft2019} }                      & \textcolor{red}{46K }    & \textcolor{red}{9936M }  & \textcolor{red}{9 }   & \textcolor{red}{o }  & \textcolor{red}{\checkmark }  & \textcolor{red}{\checkmark }  & \textcolor{red}{U64+U40*2}   & \textcolor{red}{2019 }  & \textcolor{red}{Lyft Inc. }                 & \textcolor{red}{SA-4.0 }               \\
				&                              &                         & \textcolor{red}{Argoverse\cite{chang2019argoverse} }          & \textcolor{red}{22K }    & \textcolor{red}{2354M }  & \textcolor{red}{15 }  & \textcolor{red}{o }  & \textcolor{red}{\checkmark }  & \textcolor{red}{\checkmark }  & \textcolor{red}{VV32*2}       & \textcolor{red}{2019 }  & \textcolor{red}{Argo AI }                   & \textcolor{red}{SA-4.0 }               \\
				&                              &                         & Waymo\cite{sun2019scalability}                                & 230K                     & 40710M                   & 4                     & o                    & \checkmark                    & \checkmark                    & \textcolor{red}{MR+SR*4}     & 2020                    & Waymo LLC                                   & \textcolor{red}{Waymo Lic.}            \\
				&                              &                         & \textcolor{red}{A*3D\cite{pham20203d} }                       & \textcolor{red}{39K }    & \textcolor{red}{5093M }  & \textcolor{red}{7 }   & \textcolor{red}{o }  & \textcolor{red}{\checkmark }  & \textcolor{red}{\checkmark }  & \textcolor{red}{V64}         & \textcolor{red}{2020 }  & \textcolor{red}{I2R }                       & \textcolor{red}{SA-4.0 }               \\
				&                              &                         & \textcolor{red}{DENSE\cite{bijelic2020seeing} }               & \textcolor{red}{13.5K }  & \textcolor{red}{/ }      & \textcolor{red}{4 }   & \textcolor{red}{o }  & \textcolor{red}{\checkmark }  & \textcolor{red}{\checkmark }  & \textcolor{red}{V64}         & \textcolor{red}{2020 }  & \textcolor{red}{Mercedes-Benz }             & \textcolor{red}{NC}                    \\ 
				\cline{2-14}
				& \multirow{2}{*}{synthetic}   & \multirow{2}{*}{point}  & GTA-V \cite{richter2016playing}                               & /                        & /                        & /                     & o                    & $\times$                        & \checkmark                    & \textcolor{red}{/}           & 2018                    & UC, Berkeley                                & \textcolor{red}{RO}                    \\
				&                              &                         & SynthCity\cite{griffiths2019synthcity}                        & 75000                    & 367.9M                   & 9                     & o                    & $\times$                        & \checkmark                    & \textcolor{red}{/}           & 2019                    & UCL                                         & \textcolor{red}{RO}                    \\ 
				\hline\hline
				\multirow{5}{*}{\rotatebox{90}{image/RGB-D} } & \multirow{2}{*}{image}       & \multirow{2}{*}{pixel}  & PASCAL VOC\cite{everingham2015pascal}                         & 9993                     & /                        & 20                    & i/o                  & $\times$                        & $\times$                        & \textcolor{red}{Camera}      & 2015                    & UoL/Microsoft                               & \textcolor{red}{/ }                    \\
				&                              &                         & Cityscapes\cite{cordts2016cityscapes}                         & 24998                    & 52425M                   & 30                    & o                    & \checkmark                    & \checkmark                    & \textcolor{red}{Camera}      & 2016                    & Daimler                                     & \textcolor{red}{NC}                    \\ 
				\cline{2-14}
				& \multirow{3}{*}{RGB-D}       & \multirow{3}{*}{pixel}  & NYU-Depth V2\cite{silberman2012indoor}                        & 1449                     & 445M                     & 894                   & i                    & \checkmark                    & \checkmark                    & \textcolor{red}{MK}          & 2012                    & NYU                                         & \textcolor{red}{/ }                    \\
				&                              &                         & ScanNet\cite{dai2017scannet}                                  & 2500K                    & 768000M                  & 20                    & i                    & \checkmark                    & \checkmark                    & \textcolor{red}{SS}          & 2017                    & Stanford Univ.                              & \textcolor{red}{RO}                    \\
				&                              &                         & ApolloScape$^5$\cite{huang2018apolloscape}                      & 146997                   & 1322973M                 & 25                    & o                    & \checkmark                    & \checkmark                    & \textcolor{red}{VUX*2}         & 2018                    & Baidu Research                              & \textcolor{red}{ApolloScape Lic. }     \\
				\hline
			\end{tabular}
		
		\begin{tablenotes}
			\footnotesize
			\item[*] Abbreviation explanation. anno.: annotation, ins.:instance, seq.:sequential.
			\item[1] \textbf{i} means indoor, \textbf{o} means outdoor  
			\item[2] \textbf{SL}: SICK LMS, \textbf{V32}: Velodyne HDL-32, \textbf{MLS}: Mobile Laser Scanner, \textbf{MS}: Matterport scanner, \textbf{V64}: Velodyne HDL-64E, \textbf{TLS}: Terrestrial laser scanner, \textbf{HP}: HESAI Pandora, \textbf{V16}: Velodyne VLP-16, \textbf{U32}: (unknown type) 32 channels LiDAR, \textbf{U64}: (unknown type) 64 channels LiDAR,  \textbf{U40}: (unknown type) 40 channels LiDAR,\textbf{VV32}: Velodyne VLP-32, \textbf{MR}: Mid range LiDAR, \textbf{SR}: Short range LiDAR, \textbf{MK}: Microsoft Kinect, \textbf{SS}: Structure sensor, \textbf{VUX}: VUX-1HA laser scanner.
			\item[3] \textbf{ND-3.0}: CC-BY-NC-ND-3.0, \textbf{ND-4.0}: CC-BY-ND-4.0, \textbf{SA-3.0}: CC-BY-NC-SA-3.0, \textbf{SA-4.0}: CC-BY-NC-SA-4.0, \textbf{RO}: research only, \textbf{NC}: non-commercial
			\item[4] TUM City Campus dataset provides sequential LiDAR frames, but it was collected by a MLS system, so it is still categorized as a static dataset.
			\item[5] ApolloScape provides depth data only for static street views without moving objects.
		\end{tablenotes}
	\end{threeparttable}
	\vspace{-4mm}
\end{table*}

Below, we review the publicly available 3D LiDAR datasets, followed by statistical analysis on three representative datasets.

\subsection{3D LiDAR Datasets}

According to data acquisition methods and the main applications, 3D LiDAR datasets (as listed in Table \ref{tab:1}) are divided into three groups: 1)
\textit{Static datasets}: data collected from static viewpoints by terrestrial laser scanners or using MLS (Mobile Laser Scanning) systems that capture mainly static scene objects for applications such as street view, 3D modeling, and virtual realities. 2) \textit{Sequential datasets}: data collected as sequences of frames from vehicular platforms for ADAS (Advanced Driving Assistance System) or autonomous driving applications, which can be further divided into datasets with point-wise or 3D bounding box annotations. 3) \textit{Synthetic datasets}: data collected in a virtual world by simulating any of the above data acquisition systems. In addition, the most popular image and RGB-D datasets are also listed in Table \ref{tab:1} for comparison.

\subsubsection{Static datasets}
Static datasets are most commonly used for point cloud classification tasks. Their main application scenarios include robotics, augmented reality and urban planning.

As shown in Fig. \ref{fig:1}(a), terrestrial laser scanners are usually used to collect static dense 3D LiDAR data from fixed viewpoints. MLS systems such as Fig. \ref{fig:1}(b) capture sequences of LiDAR frames from a moving vehicle. However, the data are generally static which reconstruct a large-scale street view with no motion of dynamic objects.


\subsubsection{Sequential datasets}
Sequential datasets are most commonly used for autonomous driving tasks.
As shown in Fig. \ref{fig:1}(c), autonomous driving systems are exploited to capture the sequences of LiDAR frames with a moving viewpoint on the street.
These datasets usually contain more frames but sparse points than static datasets.
\textcolor{red}{Besides, since the sensor's viewpoint moves along the direction of roads, the LiDAR points of road category usually distribute at certain angular regions, which can be predicted according to the system's setting.}

Recent years, there appear sequential datasets with both point-wise and instance labels, which help research on 3D semantic segmentation \cite{hu2019randla} and \textcolor{red}{panoptic segmentation \cite{kirillov2019panoptic}}.

\subsubsection{Synthetic datasets}

The generation of real datasets is extremely expensive due to the labor intensiveness of data annotation. Synthetic datasets are built through computer simulation, as shown in Fig. \ref{fig:1}(d), which can be large scale and have fine but cheap annotations.
The problem of using such datasets is caused by the large gap between synthetic and real scenes. Synthetic scenes can generally be very realistic, but they lack accuracy in detail. For example, pedestrians in the GTA-V \cite{richter2016playing} dataset have RGB information with rich details, but their physical models are simplified into cylinders, and the resultant point clouds lack the necessary details of real objects. 

\subsubsection{Comparison with Image and RGB-D Datasets}

A few representative image and RGB-D datasets are listed in Table \ref{tab:1}, which have much larger scales. Comparing to image and RGB-D datasets, it can be found that whatever Cityscapes \cite{cordts2016cityscapes} and ApolloScape \cite{huang2018apolloscape} for semantic segmentation in autonomous driving scenes, or ScanNet \cite{dai2017scannet} for indoor scenes, their number of pixels/frames are more sufficient than 3D LiDAR ones.
Although the studies on image and RGB-D still face the data hunger problem, it is more serious in the domain of 3D LiDAR datasets.

\begin{figure*}[ht]
	\centering
	\includegraphics[scale=0.29]{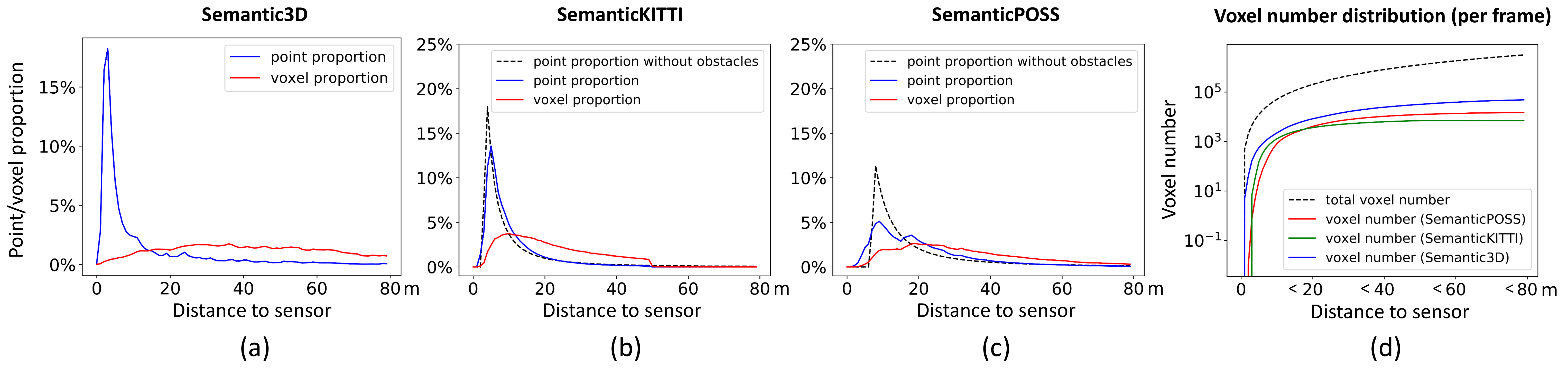}
	\caption{\textcolor{red}{Point/voxel proportion with respect to range distance: (a) Semantic3D (b) SemanticKITTI (c) SemanticPOSS, only LiDAR points with valid labels are counted. Point proportion without obstacles are simulated results for references. Semantic3D's sensor settings are not public, so it is absent in (a). The problem of unbalanced spatial distribution of LiDAR points can be alleviated by voxelization. (d) voxel number distribution with respect to range distance. The difference between the curve "total voxel number" and others is the number of empty voxels for each dataset.}}
	\label{fig:2}
	\vspace{-4mm}
\end{figure*}
\begin{figure*}[ht]
	\centering
	\includegraphics[scale=0.17]{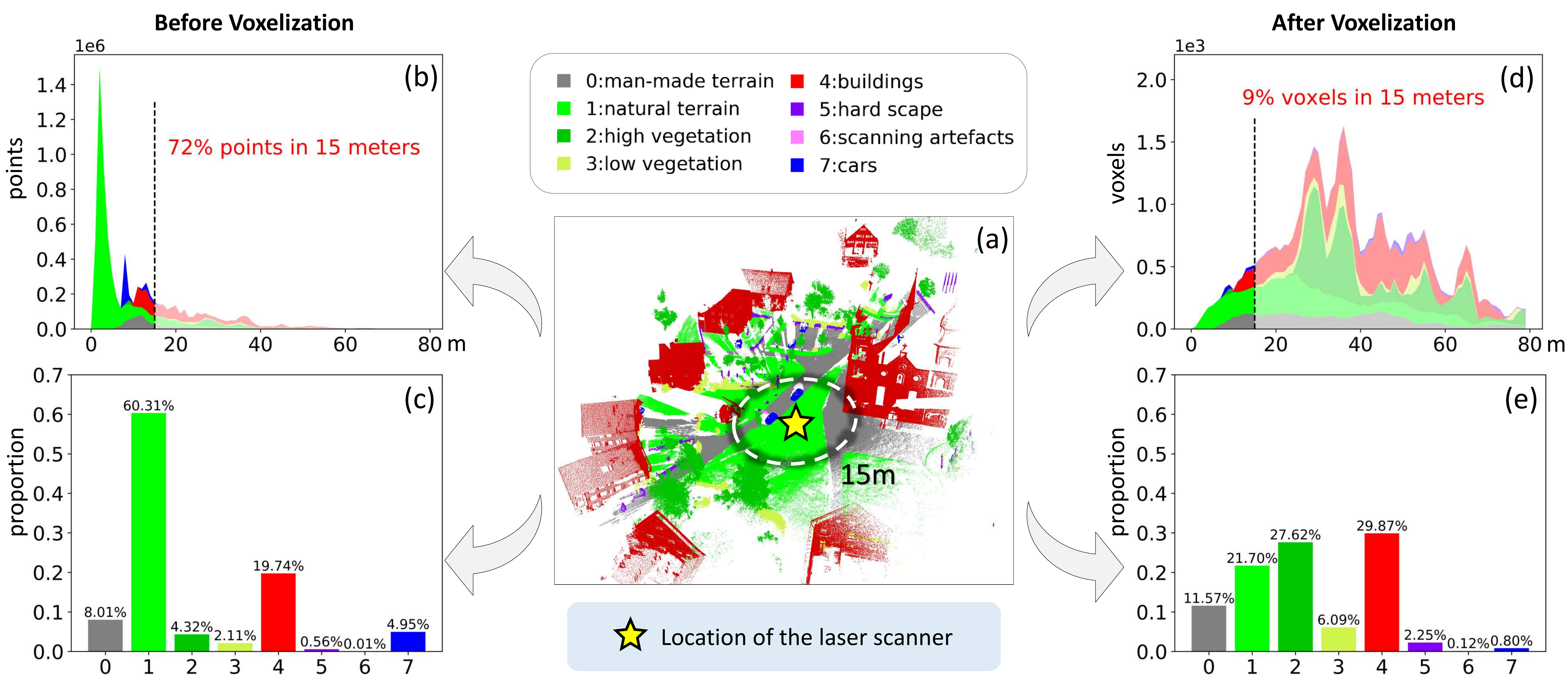}
	\caption{Comparison of point/voxel proportion of a 3D LiDAR frame in Semantic3D. (a) Visualization of the 3D LiDAR frame. (b) Proportion of LiDAR points with respect to distance. (c) Proportion of LiDAR points with respect to categories. (d) Proportion of voxels with respect to distance. (e) Proportion of voxels with respect to categories. Voxels distribute more evenly and category proportion on voxels match more with the visualized scene.}
	\label{fig:3}
	\vspace{-4mm}
\end{figure*}

\begin{figure}[]
	\centering
	\includegraphics[scale=0.258]{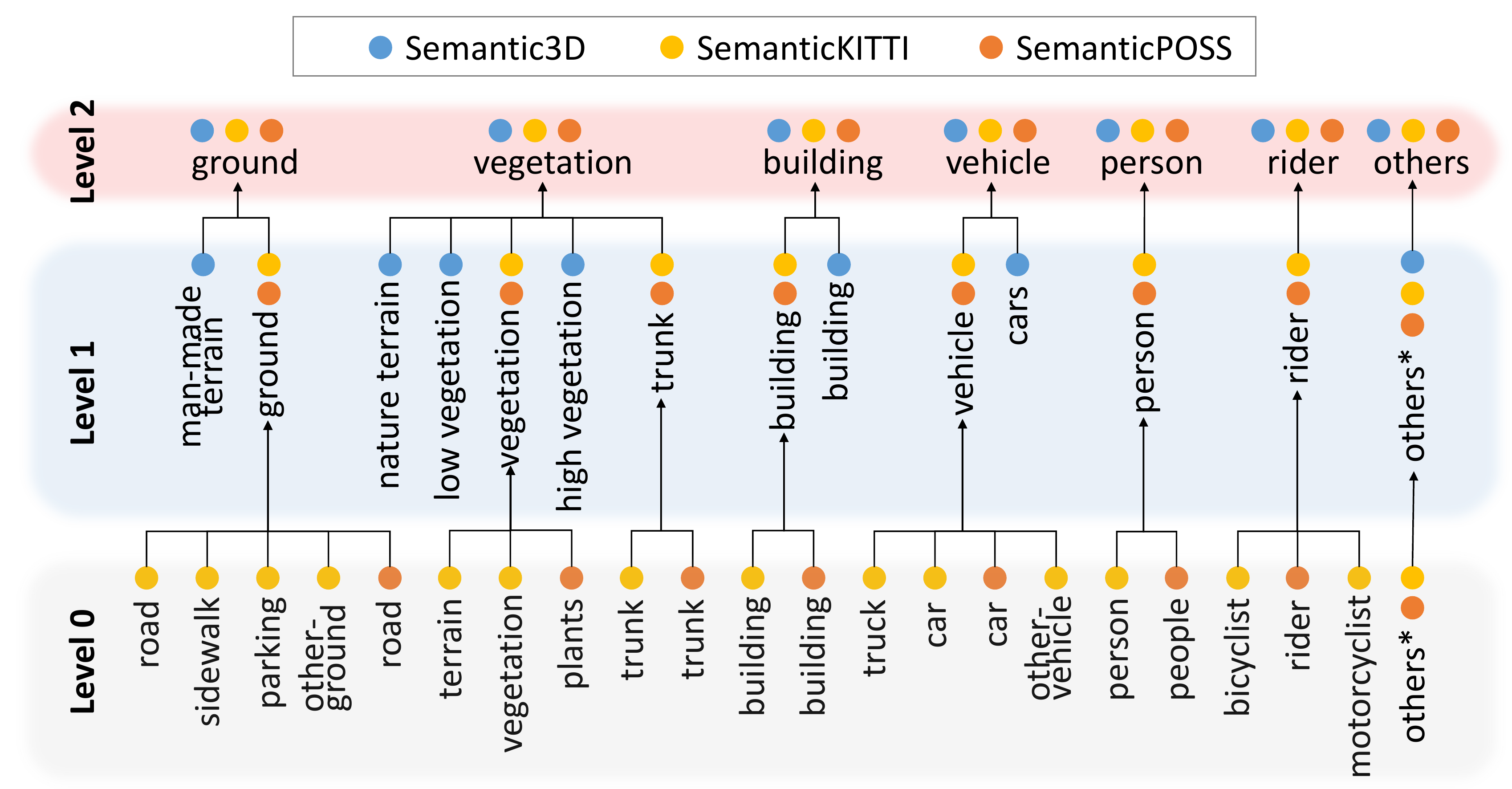}
	\caption{\textcolor{red}{Integration of the different label definitions of datasets. Level 0: some original labels of the datasets. Level 1: merged labels for per-dataset analysis. Level 2: merged labels for cross-dataset analysis. (*Some minor labels are merged as 'others' for simplicity.) }}
	\label{fig:4}
	\vspace{-4mm}
\end{figure}

\begin{figure*}[]
	\centering
	\includegraphics[scale=0.32]{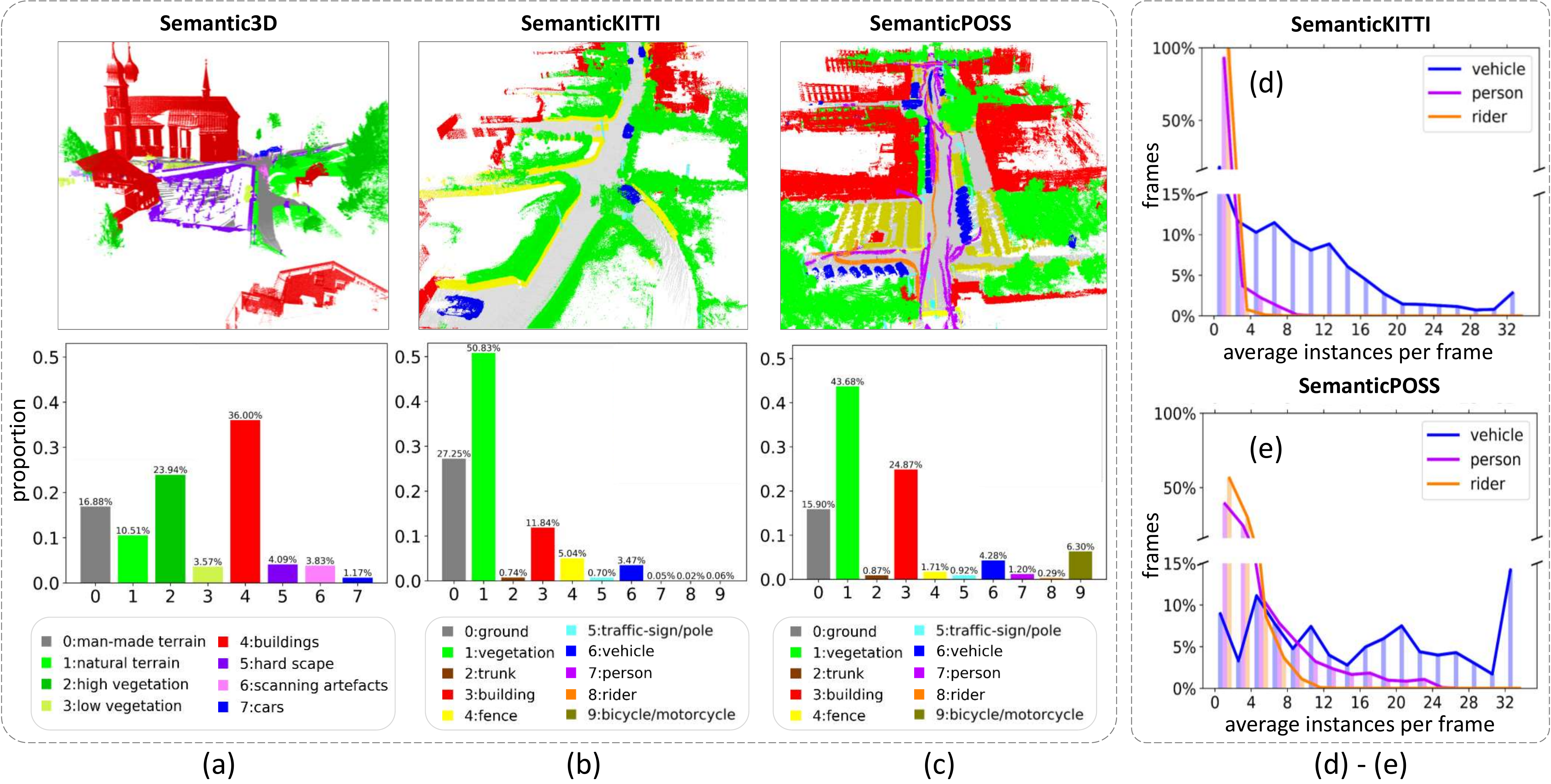}
	\caption{Overall analysis of each dataset. (a-c) Each data set is visualized by a representative scene and a histogram of the scene descriptor of the whole dataset. (d-e) Per-frame average instance number of three kinds of dynamic objects. Semantic3D is absent since it describes mainly static scene and has no instance label.}
	\label{fig:5}
	\vspace{-4mm}
\end{figure*}

\begin{table}
	\centering
	\setlength{\tabcolsep}{2.9mm}
	\renewcommand\arraystretch{1.2}
	\caption{Statistical analysis of POINT/VOXEL PROPORTION with respect to range DISTANCE of 3D LIDAR datasets}
	\label{tab:2}
	\begin{threeparttable}
		\begin{tabular}{c|cccc} 
			\hline
			\multirow{2}{*}{} & \multicolumn{4}{c}{point proportion}   \\ 
			\cline{2-5}
			& $<10m$  & $<30m$  & $<50m$  & $<70m$   \\ 
			\hline
			Semantic3D        & 69.62\% & 90.12\% & 96.39\% & 99.52\%  \\
			SemanticKITTI     & 62.54\% & 95.71\% & 99.94\% & 99.99\%  \\
			SemanticPOSS      & 20.77\% & 78.42\% & 94.12\% & 98.91\%  \\ 
			\hline\hline
			\multirow{2}{*}{} & \multicolumn{4}{c}{voxel proportion}   \\ 
			\cline{2-5}
			& $<10m$  & $<30m$  & $<50m$  & $<70m$   \\ 
			\hline
			Semantic3D        & 4.56\%  & 33.19\% & 65.37\% & 91.77\%  \\
			SemanticKITTI     & 18.37\% & 74.39\% & 99.51\% & 99.92\%  \\
			SemanticPOSS      & 5.41\%  & 51.08\% & 82.01\% & 96.18\%  \\
			\hline
		\end{tabular}
		\begin{tablenotes}
			\footnotesize
			\item[*] This table only considers points with labels and nonempty voxels.
		\end{tablenotes}
	\end{threeparttable}
	\vspace{-4mm}
\end{table}
\subsection{Statistical Analysis of the Datasets}

Three representative datasets are selected: 1) Semantic3D \cite{hackel2017semantic3d}, the largest and most popular static dataset; 2) SemanticKITTI \cite{behley2019semantickitti}, the largest and most popular sequential dataset; and 3) SemanticPOSS \cite{pan2020semanticposs}, a new dataset that describes a dynamic urban scene with rich cars, people and riders.  These datasets are analyzed statistically on the aspects of size and scene diversity.

\subsubsection{Outline of the analysis}

A straightforward method for analyzing the dataset size is to count the point number and proportion.
Table \ref{tab:2} and Fig. \ref{fig:2} show such statistics with respect to the range distance of three datasets as examples.
It can be found that LiDAR points have a much higher density at near distances.
For example, Semantic3D and SemanticKITTI both have more than 60\% of their LiDAR points measured within 10 m and less than 10\% in [30m,70m].
SemanticPOSS is slightly different, as a LiDAR sensor (Pandora \cite{pandora}) of unevenly arranged scan lines is used, which has a higher resolution on horizontal LiDAR scans.
The spatial distribution of LiDAR points is very unbalanced in these datasets, which can be seen more visually in Fig. \ref{fig:3}.
Drawing a circle at 15m to the sensor's location, 72\% percent of the total 24,671,679 LiDAR points fall into the circle, where most are on a small section of natural terrain that has very similar properties.

This is a common phenomenon in current 3D LiDAR datasets, where close objects to the sensor's location are measured with much higher point densities than farther ones. \textcolor{red}{This fact causes it meaningless of counting point number directly, as large amount of points may bring few novel information and help little in model learning. 
On the other hand, the distribution of object categories is long-tailed. This is reflected in Fig. \ref{fig:3}(b-c), for a dataset measured from an on-road viewpoint, large portion of data points could belong to \textit{road}. Such long-tail problem could result in degraded performance of deep learning models on minority categories.}

In this research, re-sampling of LiDAR points, \textbf{voxelization}, is conducted to find datasets of uniform spatial resolution.
Tessellating the 3D space evenly into voxels and projecting LiDAR points into the voxels, a set of valid voxels $V=\{v_i\}$ is obtained that has at least one LiDAR point in each,
where $v_i$ is a $K$-dimensional vector, with each column $v_i^k$ denoting the proportion of LiDAR points of label $k$ in the voxel $i$.
\textcolor{red}{In this research, a grid size is set of $(0.5\,m)^3$. It should be noted that in such a quantization process, the voxel values may vary with phase change.} By counting the number and proportion of valid voxels, curves are plotted in Fig. \ref{fig:2}(a-c), which show more even distributions with respect to distance. 
Similar results can also be found in Table \ref{tab:2} and Fig. \ref{fig:3}. \textcolor{red}{By the way, LiDARs can only scan the surfaces of obstacles, as a result, nonempty voxels actually occupy few space volume compared to total voxels of the whole scene (see Fig. \ref{fig:2}(d)).}

A number of measures are subsequently defined on voxels to analyze the statistics of the datasets.
\textbf{Category proportion} ${\cal C}^k$ is the proportion of LiDAR points or voxels having label $k$. 
\textbf{Scene descriptor} ${\cal H}$ is a $K$-dimensional vector that characterizes the scene using category proportions.
Given a voxel set $V$ of a scene, a descriptor ${\cal H}=({\cal C}^0,{\cal C}^1...,{\cal C}^{K-1})$ is generated with each category proportion calculated as ${\cal C}^k= \sum_{v_i\in V} v_i^k / |V|$.
In this research, $V$ can be a voxel set of a single frame (Semantic3D), a sequence of frames (SemanticKITTI and SemanticPOSS) or a dataset. 
\textcolor{red}{Given a \textbf{scene} ${\cal S}$, the scene descriptor is denoted as ${\cal H}_{\cal S}$. Usually, we use ${\cal H}_{\cal S}$ to represent the results on a single scene/sequence ${\cal S}$, and ${\cal H}$ to represent the average results among the whole dataset.}
Dynamic objects such as vehicles, persons and riders have a much different nature than static objects such as buildings, trees and ground and are of special importance for autonomous driving applications. Thus, we define \textbf{dynamic scene descriptor} ${\cal O}$, which is a vector of the dynamic categories only.
\textcolor{red}{The dynamic categories is a subset of ${\cal H}$'s category list, but each column of ${\cal O}$ is the instance number of the category per frame.}
\textcolor{red}{Similar with ${\cal H}_{\cal S}$, given a scene ${\cal S}$, the dynamic scene descriptor is denoted as ${\cal O}_{\cal S}$}, and for a multiple frame scene, instance numbers are the per frame average.
\textbf{Scene diversity distance} ${\cal D}$ is used to measure the difference between two scenes on the scene descriptors. To balance the magnitude of different categories, we denote standardized scene descriptor $\tilde{{\cal H}}_{\cal S}=(\tilde{{\cal C}}^0,\tilde{{\cal C}}^1...,\tilde{{\cal C}}^{K-1})$, where z-score standardized category proportion $\tilde{{\cal C}}^k=({\cal C}^k - mean({\cal C}^k)) / std({\cal C}^k)$. Therein, the mean value $mean({\cal C}^k)$ and standard deviation $std({\cal C}^k)$ are calculated over all scenes of the three datasets.
Given two scenes $S_i$ and $S_j$, the scene diversity distance is estimated as ${\cal D}(S_i, S_j) = ||\tilde{{\cal H}}_i-\tilde{{\cal H}}_j||_2/K$. \textcolor{red}{Similarly, \textbf{dynamic scene diversity distance} is defined as ${\cal D_{\cal O}}(S_i, S_j) = ||\tilde{{\cal O}}_i-\tilde{{\cal O}}_j||_2/K_{\cal O}$, in which $\tilde{{\cal O}}$ includes $\tilde{{\cal C}}^k$ of only dynamic categories and $K_{\cal O}$ is the number of dynamic categories}. To balance the magnitude of different categories, standardization is also conducted on the values of category proportions to reduce dataset bias.

Each dataset has its own definition of labels/categories, which are much different. For comparison, merging of some labels is conducted as described in Fig. \ref{fig:4}.
\textcolor{red}{\textbf{Level 0} is some original label definitions. Some labels with the same linguistic expression are not merged for their biased semantic context. For example, 'road' in SemanticPOSS includes sidewalk area, but excludes it in SemanticKITTI.
\textbf{Level 1} is for per-dataset analysis, where moderately merge of some labels are conducted to keep the special characteristics of each dataset, while the results of different datasets can still be compared.
\textbf{Level 2} is for cross-dataset analysis, where the labels are largely merged to find a uniform definition of the three datasets.
}

\begin{figure*}[t!]
	\centering
	\includegraphics[scale=0.138]{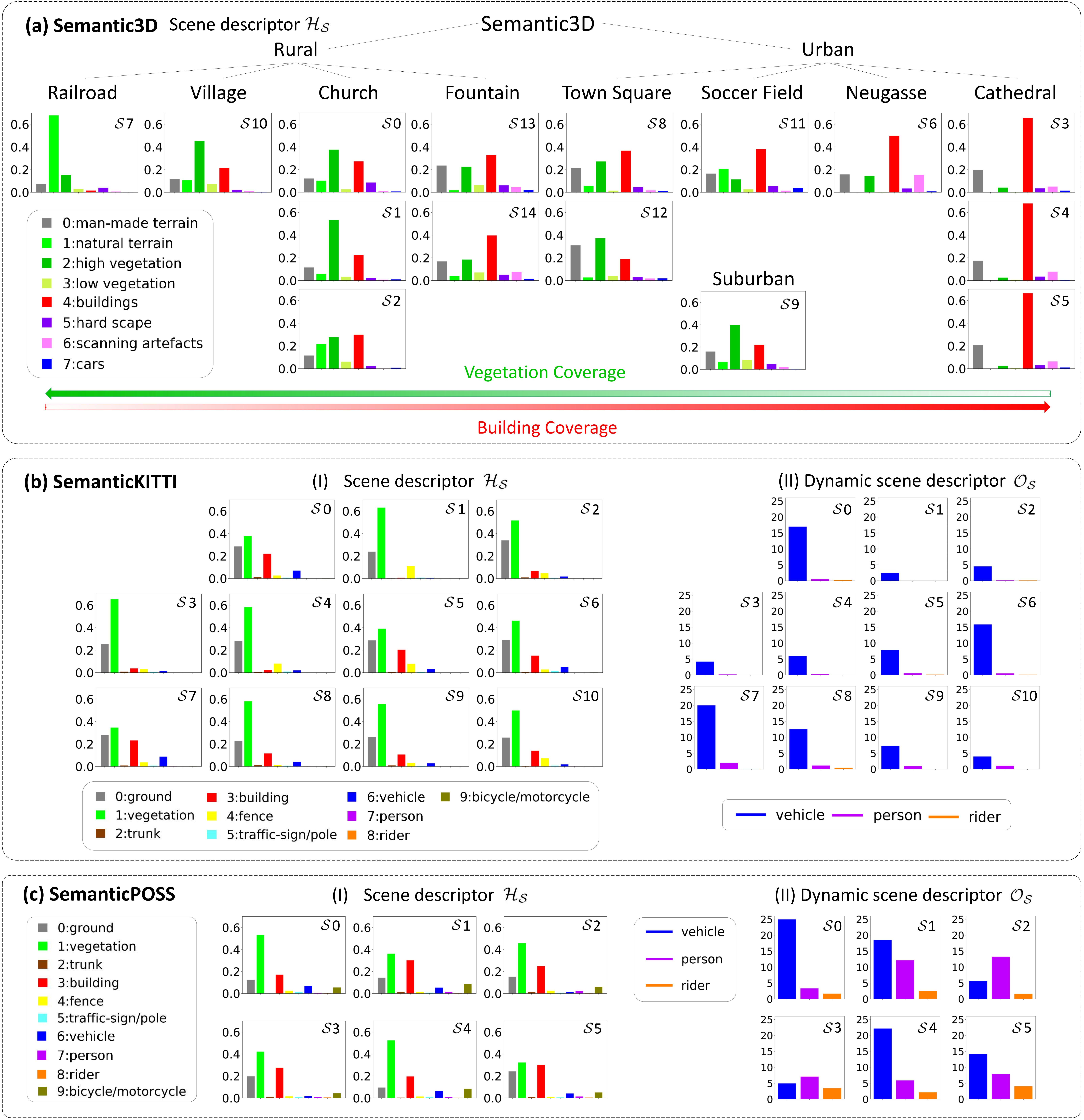}
	\caption{Per-scene analysis of each dataset. (a) Scene descriptors of 15 Semantic3D scenes, each is one frame with dense LiDAR points. (b) (Dynamic) Scene descriptors of 11 SemanticKITTI scenes, each is a sequence of LiDAR frames. (c) (Dynamic) Scene descriptors of 6 SemanticPOSS scenes, each is a sequence of LiDAR frames.}
	\label{fig:6}
	\vspace{-4mm}
\end{figure*}

Below, we first analyze each dataset on their features of scene description, then cross-scene and dataset comparison to statistically evaluate the difference in their scene diversity. Finally, we discuss the dataset concerning their representation of dynamic objects.

\subsubsection{Semantic3D}

Semantic3D contains 15 scenes in the training set. Each is a single frame that is measured using a terrestrial laser scanner from a fixed position.
A scene is visualized in Fig. \ref{fig:5}(a), with the whole dataset scene descriptor ${\cal H}$ plotted as a histogram.
It can be found that \textit{ground}, \textit{vegetation} and \textit{buildings} are the dominating categories, and the percentage of \textit{buildings} is significantly higher than the other two datasets. It has no moving objects, except for a few parking cars.

The scenes of Semantic3D are divided into three groups, i.e., urban, rural and suburban according to the geographic location of the data measurement.
As there is only one suburban scene ${\cal S}_9$, it is isolated from the tree structure in Fig. \ref{fig:6}(a).
Fig. \ref{fig:6}(a) shows scene descriptor ${\cal H}_{\cal S}$ using histograms.
It can be found that the ${\cal H}_{\cal S}$ of the same group could be very different, e.g., ${\cal S}_3$ and ${\cal S}_8$, whereas the scenes of different groups can be very similar, e.g., ${\cal S}_8$ and ${\cal S}_{13}$.
Regardless of whether the scenes are in the same or different groups, the category proportions of scene objects are very diversified. For example, ${\cal S}_7$ is a railroad scene. It is full of natural terrain, with almost no buildings. ${\cal S}_{3-5}$ are the opposite. They are cathedral scenes full of buildings but almost no vegetation.

Semantic3D has no moving objects such as \textit{person} and \textit{rider}. The dynamic scene descriptor ${\cal O}$ is not adaptive to Semantic3D, and the corresponding results of Semantic3D are absent in Fig. \ref{fig:5} and Fig. \ref{fig:6}.

In general, Semantic3D describes very diversified scenes. However, it describes static scenes with no moving object. Since each scene has only one LiDAR frame, this could create difficulty in training many deep learning methods.

\begin{figure*}[]
	\centering
	\includegraphics[scale=0.33]{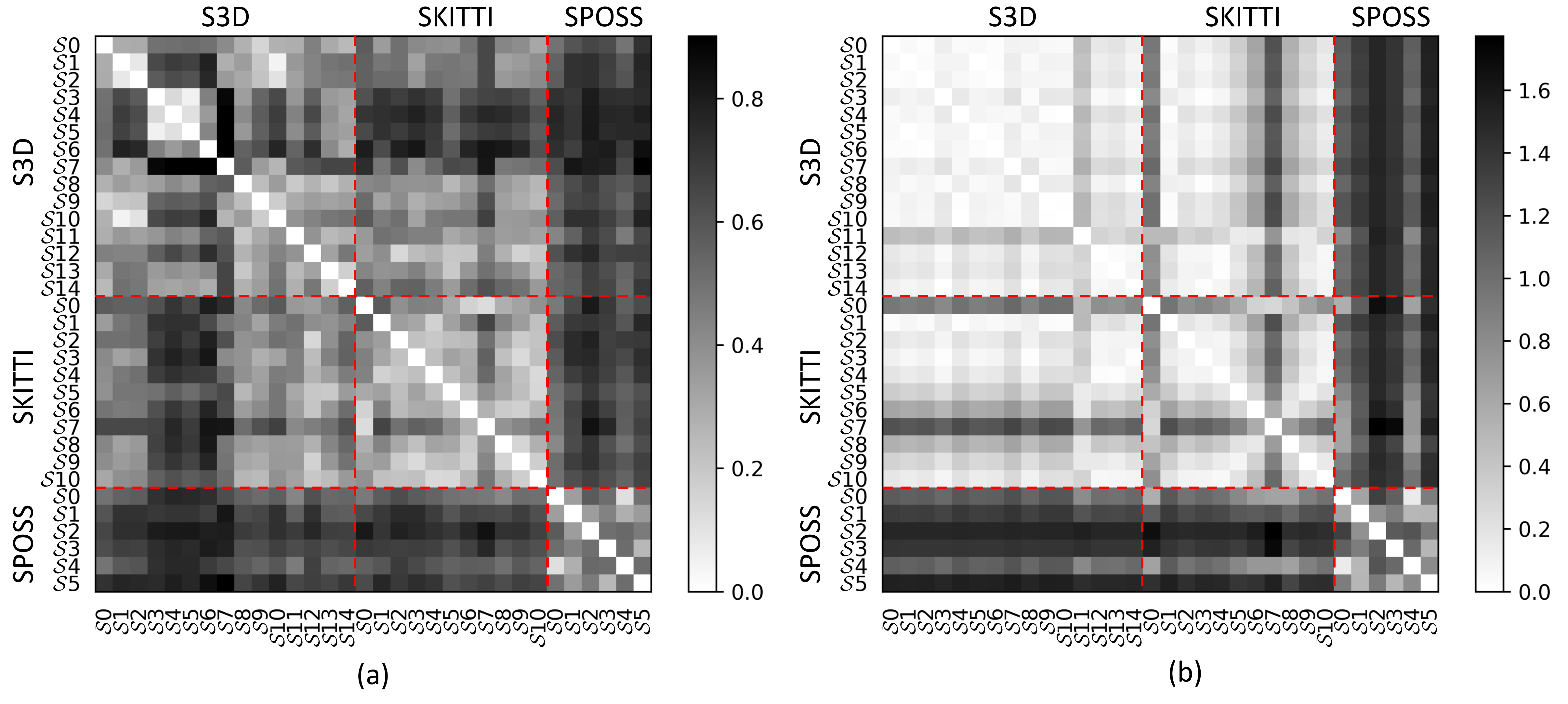}
	\caption{Cross-dataset scene diversity distance analysis. Confusion matrixes of scene (a) and dynamic scene (b) diversity distance cross the scenes of three datasets. The darker the more difference of the scenes. (S3D: Semantic3D, SKITTI: SemanticKITTI, SPOSS: SemanticPOSS)}
	\label{fig:7}
	\vspace{-4mm}
\end{figure*}
\begin{figure}[]
	\centering
	\includegraphics[scale=0.26]{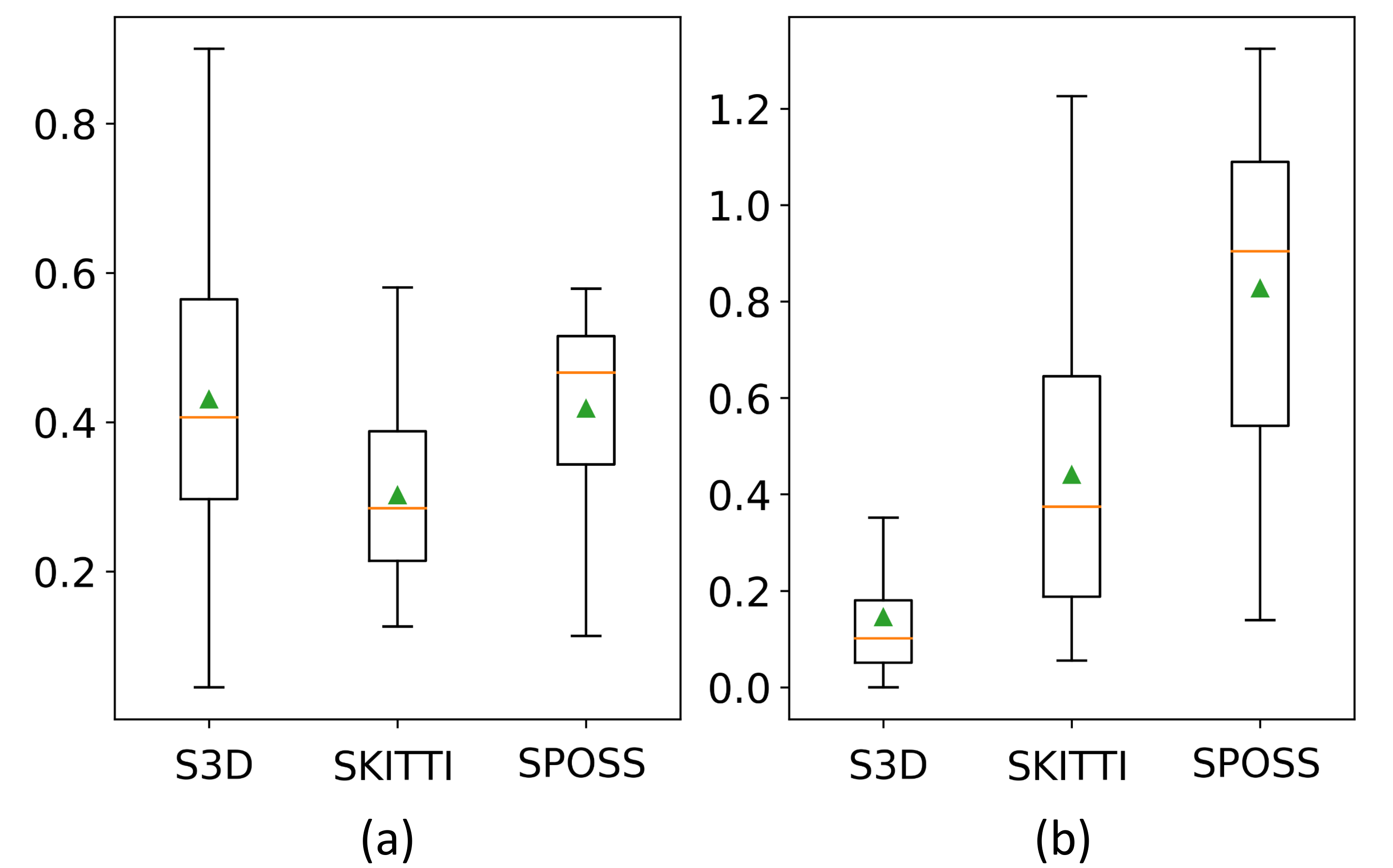}
	\vspace{-4mm}
	\caption{Inner-dataset scene diversity distance analysis. Mean and variance of scene (a) and dynamic scene (b) diversity distance of each dataset, and their comparison. (S3D: Semantic3D, SKITTI: SemanticKITTI, SPOSS: SemanticPOSS)}
	\label{fig:8}
\end{figure}

\subsubsection{SemanticKITTI}
\textcolor{red}{SemanticKITTI contains 11 sequences of 23, 201 LiDAR frames in training set that are measured continuously	from a moving vehicle on European streets.}
Each sequence is treated in this research as one scene; therefore, 11 scenes are analyzed.
One scene is visualized in Fig. \ref{fig:5}(b), with the whole dataset scene descriptor ${\cal H}$ plotted as a histogram.
Compared with Semantic3D, SemanticKITTI describes a wider street scene, where \textit{vegetation} and \textit{ground} are the two highest categories, possessing more than 50\% and 27\%, respectively. The proportion of \textit{buildings} is low compared to other datasets. 

SemanticKITTI provides instance labels of dynamic objects. The number of dynamic objects is an index to describe the complexity of a dynamic scene, which is analyzed by counting per frame instance number in Fig. \ref{fig:5}(d). SemanticKITTI has a good diversity of \textit{vehicle} distribution. However, \textit{persons} and \textit{riders} are scarce. Few scenes have more than 8 \textit{persons} or 4 \textit{riders}. This result is also confirmed by the dynamic scene descriptors ${\cal O}_{\cal S}$ of Fig. \ref{fig:6}(b). From Fig. \ref{fig:6}(b), the category proportions are not as diverse as Semantic3D.

The large data size makes SemanticKITTI very helpful for training deep learning models. However, the scenes are not as diversified as Semantic3D and have a limited number of dynamic objects.

\subsubsection{SemanticPOSS}
SemanticPOSS contains 6 sequences of LiDAR frames that were measured continuously from a moving vehicle on the campus of Peking University.
Compared to the other 3D LiDAR datasets collected on structured roads or highways,
SemanticPOSS describes scenes of abundant dynamic objects and mixed traffics.

Each sequence is treated as one scene; therefore, 6 scenes were analyzed.
In Fig. \ref{fig:5}(c), the whole dataset scene descriptor ${\cal H}$ is plotted as a histogram, where a more dynamic street scene is described.
From the scene descriptors of Fig. \ref{fig:6}(c), its general scene diversity is generally similar to SemanticKITTI, but very different to Semantic3D. 

The number of dynamic objects is analyzed in Fig. \ref{fig:5}(e). Much wider distributions can be found, compared to SemanticKITTI, where the average instances per frame are distributed from 0 to 32 for the vehicle, 24 for the person and 12 for the rider. The dynamic scene descriptors ${\cal O}_{\cal S}$ of Fig. \ref{fig:6}(c)
confirms these results too, i.e., SemanticPOSS describes scenes populated by different kinds of dynamic objects and at different crowded levels.

SemanticPOSS describes street scenes in a total of 2,998 LiDAR frames. The data size is limited with respect to SemanticKITTI, but it describes scenes of rich dynamics that are insufficient in other datasets.

\subsubsection{Cross-dataset Analysis}

A confusion matrix for scene diversity analysis is shown in Fig. \ref{fig:7}(a), where each value is the scene diversity distance ${\cal D}(S_i, S_j)$ of the pair of scenes; the whiter the less diversity, and the darker the more diversity.
For example, the first row compares the scene diversity distances of scene ${\cal S}_0$ in Semantic3D with the others. Additionally, from Semantic3D, ${\cal D}(S_0, S_1)$ and ${\cal D}(S_0, S_2)$ are light gray, but ${\cal D}(S_0, S_3)$, ${\cal D}(S_0, S_4)$ and ${\cal D}(S_0, S_5)$ are much darker. The answer can be found in Fig. \ref{fig:6}(a), where $S_{0,1,2}$ are rural churches, but $S_{3,4,5}$ are cathedral scenes. A similar confusion matrix is shown in Fig. \ref{fig:7}(b) to analyze the dynamic scene diversity. Here, each value is the dynamic scene diversity distance ${\cal D_{\cal O}}(S_i, S_j)$ of the pair of scenes.

The sub-matrices of datasets are visualized by boxplots in Fig. \ref{fig:8}(a) for inner-dataset scene diversity analysis. Semantic3D has the lowest minimum and the highest maximum, reflecting richer inner-dataset scene diversity. SemanticPOSS generally has a higher median scene diversity.


In general, scenes from different datasets tend to be more diverse than those inner-dataset. The non-diagonal blocks in Fig. \ref{fig:7}(a) are darker than diagonal blocks. 

SemanticPOSS provides the richest dynamic scene diversity.
Fig. \ref{fig:7}(b) reflects the large dynamic scene difference between SemanticPOSS and the others.
With values from the same dataset, three boxplots are drawn in Fig. \ref{fig:8}(b).
Because of the lack of moving objects in Semantic3D, its dynamic scene diversity is fairly low. From Table \ref{tab:3}, SemanticKITTI and SemanticPOSS both have many \textit{vehicles}, where the average instances per frame are 10.09 and 15.02, respectively. In addition, SemanticPOSS has more instances of \textit{person} and \textit{rider}. 

From the above analysis, we found that scenes could be very diverse that are not directly correlated with geographic location. \textcolor{red}{The existing 3D LiDAR datasets reflect only a very small set of real world scenes, whereas they exhibit insufficient inner-dataset diversity, while large cross-dataset difference. One question is how these data sets help training deep learning models, and how the test results on these datasets provide useful guidance for real-world applications.}

\begin{table}[]
	\setlength{\tabcolsep}{1.4mm}
	\renewcommand\arraystretch{1.5}
	\caption{DYNAMIC Scene OBJECTS with instance labels of the datasets}
	\label{tab:3}
	\centering
	\begin{threeparttable}
		\begin{tabular}{ccccccc} 
			\hline
			\multirow{2}{*}{dataset} & \multicolumn{3}{c}{average instance per frame}            & \multicolumn{3}{c}{category proportion (\%)}   \\ 
			\cline{2-7}
			& vehicle         & person         & rider          & vehicle       & person        & rider          \\ 
			\hline
			Semantic3D               & /               & /              & /              & 1.17          & /             & /              \\
			SemanticKITTI            & 10.09           & 0.63           & 0.18           & 3.47          & 0.05          & 0.02           \\
			SemanticPOSS             & \textbf{15.02 } & \textbf{8.29 } & \textbf{2.57 } & \textbf{4.28} & \textbf{1.20} & \textbf{0.29}  \\
			\hline
		\end{tabular}
	\end{threeparttable}
	\vspace{-4mm}
\end{table}

\section{Methods of 3D Semantic Segmentation}	\label{sec:3}

In this section, we provide a \textcolor{red}{brief} and systematic review of the representative methods of 3D semantic segmentation.

\subsection{Traditional and Deep Learning Methods}

\begin{figure*}[ht]
	\centering
	\includegraphics[scale=0.27]{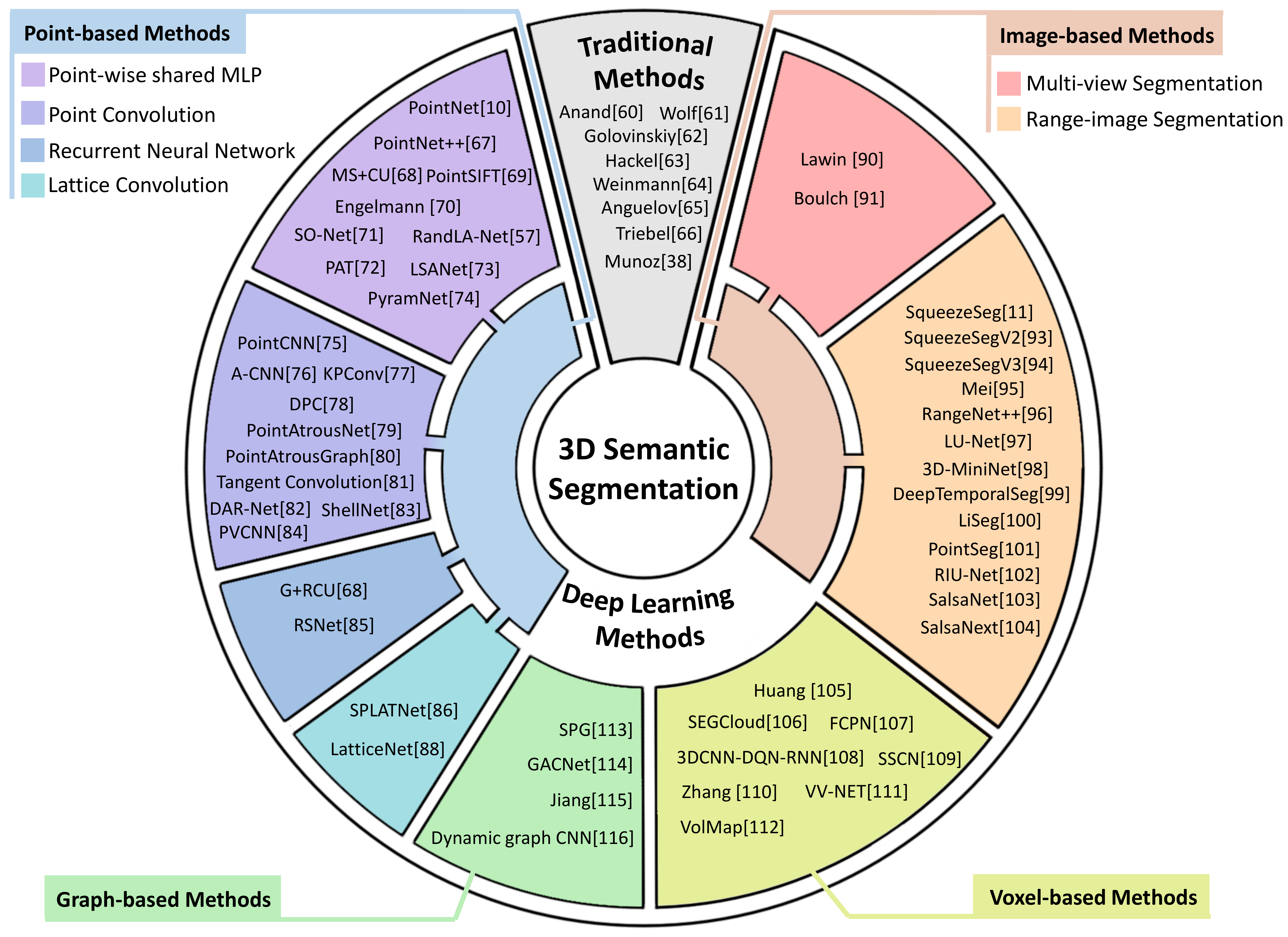}
	\caption{Overview of 3D semantic segmentation methods.}
	\label{fig:9}
	\vspace{-4mm}
\end{figure*}

Methods of 3D semantic segmentation have been widely studied for decades. As illustrated in Fig. \ref{fig:9}, they are divided into traditional and deep learning methods depending on feature representation and processing flow.

{\it Traditional methods} of 3D semantic segmentation \textcolor{red}{often use handcrafted features to extract geometric information of points and output point labels from a classifier such as Support Vector Machine (SVM) or Random Forest (RF).}

One common process of traditional methods is: over-segmenting point clouds followed by feature extraction and semantic classifiers.
\cite{anand2013contextually} and \cite{wolf2015fast} \cite{golovinskiy2009shape} are representative methods using this process.
The other common process is directly designing feature vectors of each point without prior over-segmentation, such as \cite{hackel2016fast} and \cite{weinmann2015semantic}.
\textcolor{red}{ And on this basis, some methods \cite{munoz2009contextual}\cite{wolf2015fast}\cite{anguelov2005discriminative}\cite{triebel2006robust}  use Conditional Random Fields (CRF) to aggregate contextual information.}

{\it Deep Learning Methods} use deep neural networks to learn a feature representation and directly map input data to semantic labels through an end-to-end procedure. Recently, a number of studies on 3D LiDAR semantic segmentation have been developed using deep neural networks, which can be broadly divided into four groups, as illustrated in Fig. \ref{fig:9} according to the formats of input data: 1) point-based methods, 2) image-based methods, 3) voxel-based methods, and 4) graph-based methods. Below, we provide a more detailed review of these groups of methods.

\subsection{Point-based Methods}
Point-based methods take raw point cloud as input directly and output point-wise labels. These methods can process arbitrary unstructured point clouds. The main difficulty of raw point cloud processing is how to extract local contextual features from the unstructured point cloud.

\subsubsection{Point-wise Shared MLP}
PointNet \cite{qi2017pointnet} is the pioneer of point-based deep networks for unstructured point cloud processing. It uses shared Multi-Layer Perceptrons (MLP) to extract point-wise features and aggregates global features by symmetry max pooling operation. PointNet++ \cite{qi2017pointnet++} improved PointNet \cite{qi2017pointnet} by introducing multi-scale grouping of neighboring points to extract local contextual features. 

Inspired by PointNet++ \cite{qi2017pointnet++}, many methods seek improvements of local feature extraction in different ways, \textcolor{red}{ such as considering different definitions of 'neighbor' \cite{engelmann2017exploring}\cite{jiang2018pointsift}\cite{engelmann2018know}, using different sampling approaches \cite{li2018so}\cite{hu2019randla}\cite{yang2019modeling}, and designing specific layers \cite{chen2019lsanet}\cite{zhiheng2019pyramnet}. }

\subsubsection{Point Convolution}
Convolution is the core operation for feature extraction in 2D image semantic segmentation tasks, which requests ordered inputs for local contextual information extraction. Several methods contributed to constructing an ordered feature sequence from unordered 3D LiDAR data, and then convolutional deep networks were transferred to 3D LiDAR semantic segmentation. PointCNN \cite{li2018pointcnn} ordered K-nearest points by their spatial distance to the centers, which is called the $\chi$-Conv operator for point convolution. 
\textcolor{red}{ In order to improve the convolution performance and efficiency, many specific convolutional networks were proposed, such as A-CNN \cite{komarichev2019cnn}, KPConv \cite{thomas2019kpconv}, DPC \cite{engelmann2019dilated}, PointAtrousNet \cite{pan2019pointatrousnet}, PointAtrousGraph \cite{pan2019pointatrousgraph}, tangent convolution \cite{tatarchenko2018tangent}, DAR-Net \cite{zhao2019dar}, ShellNet \cite{zhang2019shellnet}, and PVCNN \cite{liu2019point}.}

\subsubsection{Recurrent Neural Network}
Recurrent Neural Networks (RNN) are often used to extract contextual information of a sequence. For 3D semantic segmentation, RNN can extract spatial context by feeding ordered feature vectors in space.
Engelmann et al. \cite{engelmann2017exploring} proposed Grid (G) and Recurrent Consolidation Unit (RCU), which divide space into several grids as the network input. 
RSNet \cite{huang2018recurrent} transfers unordered points into an ordered sequence of feature vectors with a slice pooling layer. Then, RNN takes the sequence as input and aggregates spatial context information.

\subsubsection{Lattice Convolution}
A sparse permutohedral lattice is suitable for sparse data processing such as point clouds. SPLATNet \cite{su2018splatnet} applies the Bilateral Convolution Layer (BCL) \cite{jampani2016learning} to provide a transformation between point clouds and sparse lattices, which performs convolutions efficiently. 
And LatticeNet \cite{rosu2019latticenet} introduced a novel slicing operator for lattice processing to obtain a better local feature representation. 

\subsection{Image-based Methods}
Image-based methods project 3D LiDAR data onto a surface to generate 2D images as deep model inputs. These methods are usually derived from image semantic segmentation models, such as Fully Convolutional Network (FCN) \cite{long2015fully} and U-Net \cite{ronneberger2015u}. The output predictions with pixel-wise labels are reprojected to origin 3D LiDAR points. 

\subsubsection{Multi-view Segmentation}
A simple projection strategy is choosing several positions for taking photos of given point clouds. Lawin et al. \cite{lawin2017deep} rotated a virtual camera around a fixed vertical axis to generate multi-view synthetic images, which were processed by a FCN-based multi-stream architecture. 
Boulch et al. \cite{boulch2017unstructured} generated a mesh of 3D LiDAR data, and then produced images by randomly choosing virtual camera positions. 
For these multi-view methods, it is important to choose appropriate camera positions and projection strategies to reduce information loss.

\subsubsection{Range Image Segmentation} 
Range images are usually generated by projecting one frame of 3D LiDAR data onto a spherical surface. SqueezeSeg \cite{wu2018squeezeseg} is a typical end-to-end network for range image semantic segmentation based on SqueezeNet \cite{iandola2016squeezenet} and CRF. SqueezeSegV2 \cite{wu2019squeezesegv2} and \textcolor{red}{SqueezeSegV3 \cite{xu2020squeezesegv3}} are improved versions of SqueezeSeg.

Range image segmentation methods are usually implemented on sequential datasets, while spatial and temporal information can be incorporated. Mei et al. \cite{mei2019semantic} and RangeNet++ \cite{milioto2019rangenet++} introduced spatial constraints for predictions with more region consistency.
\textcolor{red}{ LU-Net \cite{biasutti2019lu} and 3D-MiniNet \cite{alonso20203d} introduced 3D spatial features to the projected range images.}
DeepTemporalSeg \cite{dewan2019deeptemporalseg} introduced temporal constraints based on a Bayes filter to make predictions more temporally consistent.

Some range image segmentation methods have focused on real-time performance, which is essential for applications such as  autonomous driving and unmanned detectors,
\textcolor{red}{ such as LiSeg \cite{zhang2018liseg}, PointSeg \cite{wang2018pointseg}, RIU-Net \cite{biasutti2019riu}, SalsaNet \cite{aksoy2019salsanet} and SalsaNext \cite{cortinhal2020salsanext}.}

\subsection{Voxel-based Methods}
Voxel-based methods transfer 3D LiDAR data into voxels for structured data representation. These methods usually take voxels as input and predict each voxel with one semantic label. 

A number of voxel-based methods \cite{huang2016point}\cite{tchapmi2017segcloud}\cite{rethage2018fully} are based on 3D Convolutional Neural Network (3D CNN). 
However, it is challenging for voxel-based methods to find a proper voxel size that balances precision and computational efficiency. Some methods have contributed to reducing computational cost of 3D convolution on sparse data while maintaining acceptable accuracy,
\textcolor{red}{ such as 3DCNN-DQN-RNN \cite{liu20173dcnn}, Submanifold Sparse Convolution \cite{graham20183d}, efficient convolution \cite{zhang2018efficient}, VV-NET \cite{meng2019vv}, and VolMap \cite{radi2019volmap}. }

\subsection{Graph-based Methods}
Graph-based methods construct a graph from 3D LiDAR data. A vertex usually represents a point or a group of points, and edges represent adjacency relationships between vertexes. Graph construction and graph convolution are two key operations of these methods.

Super-Point Graph (SPG) \cite{landrieu2018large} is a representative work. This network employs a PointNet \cite{qi2017pointnet} to encode vertex features and graph convolutions to extract contextual information. 
GACNet \cite{wang2019graph} proposed a novel graph convolution operation, Graph Attention Convolution (GAC), to consider structural relations between points of the same object. 
In order to learn more valid local features, 
\textcolor{red}{ some methods \cite{jiang2019hierarchical} \cite{wang2019dynamic} try to construct graph dynamically instead of fixed graph. }

\begin{table*}[b]
	\centering
	\renewcommand{\arraystretch}{1.2}
	\caption{Design of experiments.}
	\begin{tabular}{m{7em}|m{17em}|m{17em}|m{17em}}
		\hline \rule{0pt}{20pt} 
		& \makebox[15em][s]{\shortstack[l]{Experiment 1\\ cross-scene generalization evaluation.}}
		& \makebox[15em][s]{\shortstack[l]{Experiment 2\\ cross-dataset generalization evaluation.}}
		& \makebox[15em][s]{\shortstack[l]{Experiment 3\\ Dataset size effects evaluation.}}
		\\ \hline
		Scope & Scene diversity & Scene diversity & Dataset size \\ \hline
		Purpose & Train models on a single dataset with different scene diversity, and examine how the data hunger problem of scene diversity affects models’ performances. & Train models on different datasets, and examine how the data hunger problem of scene diversity affects models’ performances. & Examine how the data hunger problem of dataset size affects models’ performances, and whether the models are hungry for dataset size. \\ \hline
		Dataset & Semantic3D & SemanticKITTI, SemanticPOSS & SemanticKITTI \\ \hline
		Model & PointNet++, SPG & PointNet++, SqueezeSegV2, SPG & PointNet++, SqueezeSegV2, SPG \\ \hline
		Method & Three sub-datasets are made on Semantic3D, 1) urban: a dataset contains urban scene only, 2) rural: a dataset contains rural scenes only, 3) mix: a dataset contains both rural and urban scenes. Each sub-dataset is divided randomly into two parts for training and testing. The selected models are trained and tested crosswise on these sub-datasets. & Three datasets are used, SemanticKITTI, SemanticPOSS, and a mixed dataset, which contains both SemanticKITTI and SemanticPOSS data. Similar to Experiment 1, selected methods are trained and tested crosswise on these datasets. & Evaluate the model performance using different amounts of training data. We use parts of SemanticKITTI data to train the models and compare mIoU of the model predictions. \\ \hline
		Label for testing & \textit{man-made terrain, natural terrain, high vegetation, low vegetation, building, hard scape, car} & \textit{person, rider, vehicle, traffic sign/pole, trunk, vegetation, fence, building, bicycle/motorcycle, ground} & \textit{person, rider, vehicle, traffic sign/pole, trunk, vegetation, fence, building, bicycle/motorcycle, ground} \\ \hline
		Other details & Weights of all categories are the same for training. & Weights of all categories are the same for training. Single frame of point clouds is used as input, not overlapped frames. & Weights of all categories are the same for training. Single frame of point clouds is used as input, not overlapped frames. \\ \hline
		Result & Table \ref{tab:5}, Fig. \ref{fig:12}(a) & Table \ref{tab:6}, Fig. \ref{fig:12}(b) & Table \ref{tab:7}, Fig. \ref{fig:14}(a) \\ \hline
	\end{tabular}
	\label{tab:4}
\end{table*}

\section{Data Hungry or Not? Experiments} \label{sec:4}

As addressed in the previous sections, three representative datasets, Semantic3D \cite{hackel2017semantic3d}, SemanticKITTI \cite{behley2019semantickitti} and SemanticPOSS \cite{pan2020semanticposs}, are analyzed statistically. In this section, we design three experiments to answer the following questions: \textcolor{red}{How do scene diversity and training dataset size influence the model performance? Does the data hunger problem in scene diversity and dataset size exist for 3D LiDAR datasets?  Do different models have different sensitivity to the data hunger effect?}

\textcolor{red}{More specifically, as reviewed in previous section, 3D semantic segmentation methods using deep learning techniques can be broadly divided into four groups according to their input data format.
The experiments below explore whether the models of different input data formats face different degree of data hunger.}

\subsection{Selected Methods in Experiments}

Three methods are selected in the experiments, PointNet++ \cite{qi2017pointnet++}, SqueezeSegV2 \cite{wu2019squeezesegv2}, and SPG \cite{landrieu2018large}, 
\textcolor{red}{representing the fundamental architectures of point-, image- and graph-based methods respectively, which are of broad awareness. Since voxel-based method share the features of both point- and image-based ones, it is absent in this experiment.}

\textbf{PointNet++} is a typical point-based method taking raw point clouds as input. 
PointNet++ is a hierarchical encode-decode structure based on shared MLP. The sampling, grouping, and PointNet layer are used to learn local contextual features. Many point-based methods are derived from the PointNet++ architecture.

\textbf{SqueezeSegV2} is a typical image-based method taking range images as input. The architecture of SqueezeSegV2 is a typical convolutional neural network.
It is chosen as a deputy of CNN-based architectures, which is similar to most image-based methods.

\textbf{SPG} is a typical graph-based method taking the super-point graph as input. A segmentation algorithm is used to partition point clouds into several groups as vertexes of the graph. Edges are constructed to represent contextual relationships between vertexes by comparing the shape and size of the adjacent point groups. 
The PointNet layer and Gated Recurrent Unit (GRU) are used to learn local contextual features and implement graph convolution.

\subsection{Outline of the Experiments}

For the 3D semantic segmentation task, deep learning models need to give semantic predictions to every point of the given point cloud. To evaluate the model performance, we use the Intersection over Union (IoU) given by

\vspace{-2mm}
\begin{equation}
IoU_{c} = \frac{TP_{c}}{TP_{c}+FP_{c}+FN_{c}}
\end{equation}

where $TP_{c}$,$FP_{c}$,$FN_{c}$ denote the number of true positive, false positive, false negative predictions of category $c$. Let $N$ be the number of categories used for measurement, the mean IoU (mIoU) is defined as the arithmetic mean of IoU, namely,

\vspace{-2mm}
\begin{equation}
mIoU = \frac{1}{N}\sum_{c=1}^{N}IoU_{c}
\end{equation}

To analyze the data hunger effect of scene diversity and data size, three experiments shown in Table \ref{tab:4}, are designed.

\begin{table}[t] 
	\setlength{\tabcolsep}{1.8mm}
	\centering
	\begin{threeparttable}[b]
	\renewcommand\arraystretch{1.4}
	\caption{Result of Experiment 1: cross-scene generation evaluation.}
	\label{tab:5}
	\begin{tabular}{c|c|ccc|ccc}
		\hline
		\rowcolor{r00}Category & Model                           & \multicolumn{3}{c|}{\textbf{PointNet++}} & \multicolumn{3}{c}{\textbf{SPG}} \\ \hline
		& \diagbox[width=1.8cm, height=0.6cm]{Test}{Train} & urban     & rural    & mix      & urban  & rural  & mix    \\ \hline
		\multirow{3}{*}{\rotatebox{90}{\shortstack[l]{man-made \\ terrain}}}  
		& urban     & \cellcolor{r99}\textbf{95.3}     & \cellcolor{r00}90.0    & \cellcolor{r81}94.3    & \cellcolor{r99}\textbf{99.7}  & \cellcolor{r00}99.6  & \cellcolor{r00}99.6  \\
		& rural     & \cellcolor{r20}89.0     & \cellcolor{r99}\textbf{91.8}    & \cellcolor{r00}88.3    & \cellcolor{r00}80.0  & \cellcolor{r97}96.1  & \cellcolor{r99}\textbf{96.5}  \\
		& mix       & \cellcolor{r99}\textbf{92.2}     & \cellcolor{r00}90.5    & \cellcolor{r52}91.4    & \cellcolor{r00}97.7  & \cellcolor{r94}99.3  & \cellcolor{r99}\textbf{99.4}  \\ \hline
		\multirow{3}{*}{\rotatebox{90}{\shortstack[l]{natural \\ terrain}}}  
		& urban     & \cellcolor{r99}\textbf{92.1}     & \cellcolor{r00}68.0    & \cellcolor{r73}85.6    & \cellcolor{r99}\textbf{93.8}  & \cellcolor{r00}79.3  & \cellcolor{r36}84.6  \\
		& rural     & \cellcolor{r24}80.0     & \cellcolor{r99}\textbf{85.3}    & \cellcolor{r00}78.3    & \cellcolor{r00}32.8  & \cellcolor{r99}\textbf{92.4}  & \cellcolor{r97}91.1  \\
		& mix       & \cellcolor{r99}\textbf{82.1}     & \cellcolor{r00}79.3    & \cellcolor{r10}79.6    & \cellcolor{r00}51.2  & \cellcolor{r97}88.1  & \cellcolor{r99}\textbf{89.1}  \\ \hline
		\multirow{3}{*}{\rotatebox{90}{\shortstack[l]{high \\ vegetation}}}  
		& urban     & \cellcolor{r00}88.5     & \cellcolor{r99}\textbf{91.3}    & \cellcolor{r85}90.9    & \cellcolor{r00}91.4  & \cellcolor{r89}93.2  & \cellcolor{r99}\textbf{93.4}  \\
		& rural     & \cellcolor{r47}90.2     & \cellcolor{r99}\textbf{93.3}    & \cellcolor{r00}87.4    & \cellcolor{r00}11.2  & \cellcolor{r99}\textbf{85.9}  & \cellcolor{r31}34.7  \\
		& mix       & \cellcolor{r42}90.2     & \cellcolor{r99}\textbf{93.1}    & \cellcolor{r00}88.1    & \cellcolor{r00}49.0  & \cellcolor{r99}\textbf{89.1}  & \cellcolor{r29}60.8  \\ \hline
		\multirow{3}{*}{\rotatebox{90}{building}}  
		& urban     & \cellcolor{r37}96.5     & \cellcolor{r00}95.9    & \cellcolor{r99}\textbf{97.5}    & \cellcolor{r99}\textbf{95.3}  & \cellcolor{r00}87.3  & \cellcolor{r84}94.1  \\
		& rural     & \cellcolor{r34}89.8     & \cellcolor{r99}\textbf{92.3}    & \cellcolor{r00}88.5    & \cellcolor{r00}76.9  & \cellcolor{r99}\textbf{95.1}  & \cellcolor{r87}92.8  \\
		& mix       & \cellcolor{r16}94.0     & \cellcolor{r99}\textbf{94.5}    & \cellcolor{r00}93.9    & \cellcolor{r36}90.7  & \cellcolor{r00}88.9  & \cellcolor{r99}\textbf{93.8}  \\ \hline \hline
		\multirow{3}{*}{\rotatebox{90}{mIoU}}
		& urban     & \cellcolor{r99}\textbf{69.6}     & \cellcolor{r00}62.0    & \cellcolor{r86}68.6    & \cellcolor{r99}\textbf{80.5}  & \cellcolor{r00}68.4  & \cellcolor{r86}78.9  \\
		& rural     & \cellcolor{r00}71.9     & \cellcolor{r99}\textbf{78.9}    & \cellcolor{r22}73.5    & \cellcolor{r00}38.3  & \cellcolor{r99}\textbf{86.6}  & \cellcolor{r68}71.4  \\
		& mix       & \cellcolor{r00}71.7     & \cellcolor{r99}\textbf{72.1}    & \cellcolor{r00}71.7    & \cellcolor{r00}62.6  & \cellcolor{r96}75.3  & \cellcolor{r99}\textbf{75.7}  \\ \hline
	\end{tabular}
	\begin{tablenotes}
		\item[1] IoU of some dominant categories. Deeper color means the better performance on a specific test scene using a model.
	\end{tablenotes}
	\end{threeparttable}
\end{table}

\begin{figure}[t]
	\centering
	\includegraphics[scale=0.4]{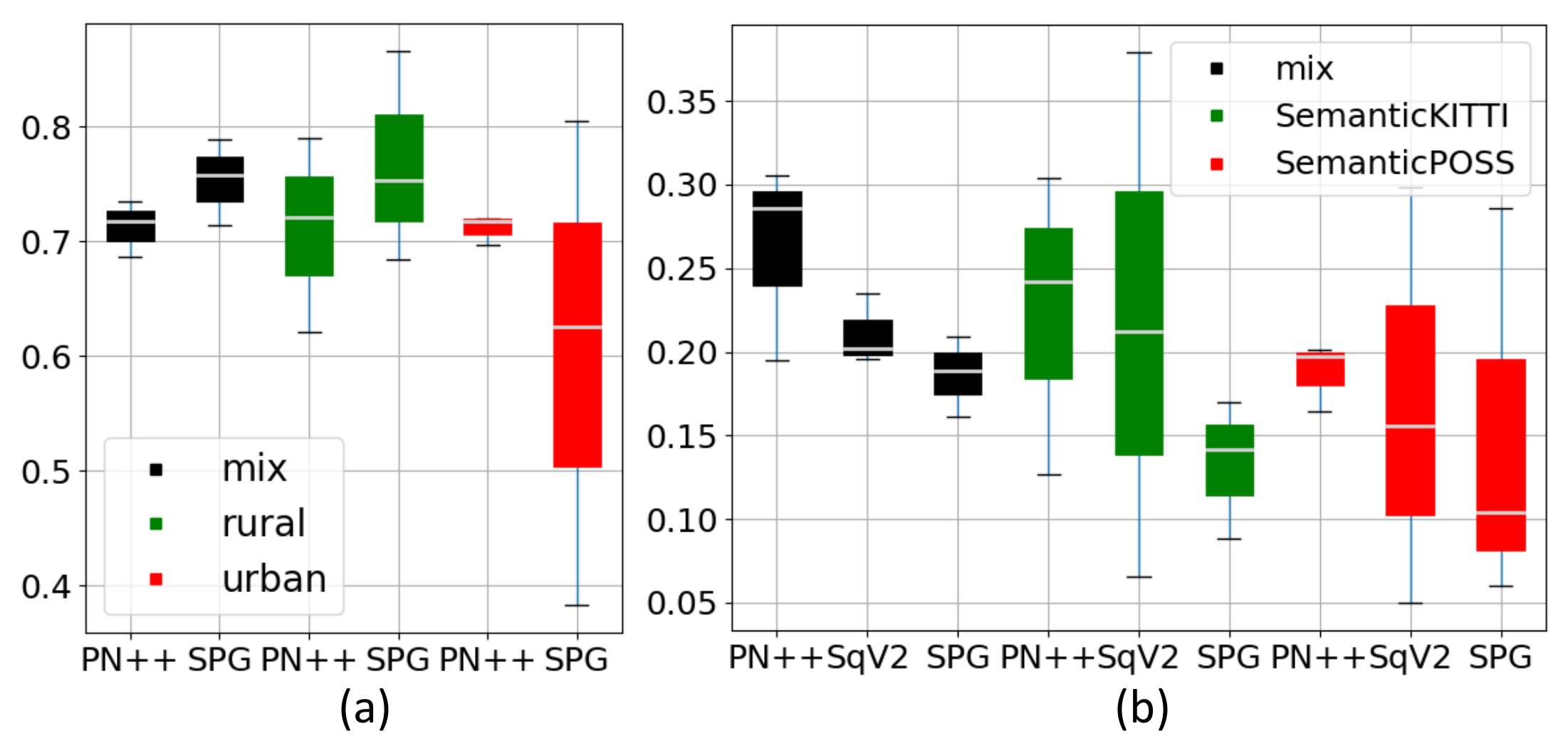}
	\caption{mIoU of the models trained on different scenes (PN++: PointNet++, SqV2: SqueezeSegV2). (a) Result of Experiment 1. (b) Result of Experiment 2. The different color of box means different training scenes. The values of vertical axis means mIoU. Each box shows the maximum, minimum and median of the mIoU on different test set.}
	\label{fig:12}
	\vspace{-4mm}
\end{figure}

\begin{table*}[!h] 
	\setlength{\tabcolsep}{2.5mm}   	\centering
	\begin{threeparttable}[b]
		\renewcommand\arraystretch{1.3}
		\caption{Result of Experiment 2: cross-dataset generation evaluation.}
		\label{tab:6}
		\begin{tabular}{c|c|ccc|ccc|ccc}
			\hline
			Category & Model                           & \multicolumn{3}{c|}{\textbf{PointNet++}} & \multicolumn{3}{c|}{\textbf{SqueezeSegV2}} & \multicolumn{3}{c}{\textbf{SPG}}\\ \hline
			& \diagbox[width=2.0cm, height=0.6cm]{Test}{Train} & SKITTI     & SPOSS    & mix      & SKITTI     & SPOSS  & mix    & SKITTI     & SPOSS  & mix  \\ \hline
			\multirow{3}{*}{\rotatebox{90}{person}} 
			& SKITTI  & \cellcolor{r00}0.7  & \cellcolor{r99}\textbf{6.4}  & \cellcolor{r19}1.8  & \cellcolor{r99}\textbf{15.4} & \cellcolor{r00}2.9  & \cellcolor{r22}5.7  & \cellcolor{r63}2.8  & \cellcolor{r00}0.9  & \cellcolor{r99}\textbf{3.9}  \\
			& SPOSS   & \cellcolor{r00}0.0  & \cellcolor{r99}\textbf{20.8} & \cellcolor{r89}18.7 & \cellcolor{r00}0.0  & \cellcolor{r99}\textbf{18.4} & \cellcolor{r85}15.8 & \cellcolor{r00}4.2  & \cellcolor{r94}17.2 & \cellcolor{r99}\textbf{18.0} \\
			& mix     & \cellcolor{r00}0.0  & \cellcolor{r99}\textbf{18.2} & \cellcolor{r96}17.4 & \cellcolor{r00}2.6  & \cellcolor{r99}\textbf{16.9}& \cellcolor{r83}14.5 & \cellcolor{r00}4.0  & \cellcolor{r38}9.1  & \cellcolor{r99}\textbf{17.4} \\ \hline
			\multirow{3}{*}{\rotatebox{90}{vehicle}} 
			& SKITTI  & \cellcolor{r99}\textbf{53.2} & \cellcolor{r00}16.6 & \cellcolor{r99}53.1 & \cellcolor{r99}\textbf{68.5} & \cellcolor{r00}8.0  & \cellcolor{r36}30.1 & \cellcolor{r99}\textbf{49.0}  & \cellcolor{r00}15.9 & \cellcolor{r92}46.2 \\
			& SPOSS   & \cellcolor{r00}4.1  & \cellcolor{r99}\textbf{8.9}  & \cellcolor{r97}8.8  & \cellcolor{r00}1.6  & \cellcolor{r99}\textbf{34.9} & \cellcolor{r42}15.7 & \cellcolor{r99}\textbf{15.3}  & \cellcolor{r00}11.5 & \cellcolor{r73}14.3 \\
			& mix     & \cellcolor{r99}\textbf{44.7} & \cellcolor{r00}17.4 & \cellcolor{r92}42.5 & \cellcolor{r99}\textbf{57.3} & \cellcolor{r00}8.6 & \cellcolor{r39}27.5 & \cellcolor{r99}\textbf{42.1}  & \cellcolor{r00}15.3 & \cellcolor{r91}39.6 \\ \hline
			\multirow{3}{*}{\rotatebox{90}{vegetation}} 
			& SKITTI  & \cellcolor{r98}64.0 & \cellcolor{r00}48.2 & \cellcolor{r99}\textbf{64.3} & \cellcolor{r99}\textbf{79.9} & \cellcolor{r00}0.2 & \cellcolor{r50}40.0 & \cellcolor{r97}51.2  & \cellcolor{r00}33.8 & \cellcolor{r99}\textbf{51.7} \\
			& SPOSS   & \cellcolor{r00}46.3 & \cellcolor{r99}\textbf{51.2} & \cellcolor{r75}50.0 & \cellcolor{r00}8.9  & \cellcolor{r99}\textbf{56.3}  & \cellcolor{r29}22.7 & \cellcolor{r00}49.2  & \cellcolor{r99}\textbf{52.8} & \cellcolor{r88}52.4 \\
			& mix     & \cellcolor{r76}56.3 & \cellcolor{r00}49.7 & \cellcolor{r99}\textbf{58.4} & \cellcolor{r99}\textbf{53.9} & \cellcolor{r00}22.2 & \cellcolor{r39}34.6 & \cellcolor{r86}50.4  & \cellcolor{r00}40.2 & \cellcolor{r99}\textbf{52.0} \\ \hline
			\multirow{3}{*}{\rotatebox{90}{building}} 
			& SKITTI  & \cellcolor{r95}61.8 & \cellcolor{r00}41.5 & \cellcolor{r99}\textbf{62.7} & \cellcolor{r99}\textbf{69.7} & \cellcolor{r00}9.6 & \cellcolor{r62}47.1 & \cellcolor{r99}\textbf{38.7}  & \cellcolor{r00}16.2 & \cellcolor{r77}33.7 \\
			& SPOSS   & \cellcolor{r00}18.9 & \cellcolor{r99}\textbf{42.7} & \cellcolor{r74}36.7 & \cellcolor{r00}9.2  & \cellcolor{r99}\textbf{47.0}  & \cellcolor{r99}\textbf{47.0} & \cellcolor{r00}34.3  & \cellcolor{r99}\textbf{55.3} & \cellcolor{r82}51.6 \\
			& mix     & \cellcolor{r00}40.3 & \cellcolor{r19}42.0 & \cellcolor{r99}\textbf{49.2} & \cellcolor{r54}37.9 & \cellcolor{r00}27.3 & \cellcolor{r99}\textbf{47.1} & \cellcolor{r34}36.7  & \cellcolor{r00}33.7 & \cellcolor{r99}\textbf{42.6} \\ \hline
			\multirow{3}{*}{\rotatebox{90}{ground}} 
			& SKITTI  & \cellcolor{r99}\textbf{80.9} & \cellcolor{r00}57.2 & \cellcolor{r99}80.8 & \cellcolor{r99}\textbf{88.5} & \cellcolor{r00}28.9 & \cellcolor{r43}66.3 & \cellcolor{r99}51.3  & \cellcolor{r00}36.8 & \cellcolor{r99}\textbf{51.4} \\
			& SPOSS   & \cellcolor{r00}40.0 & \cellcolor{r99}\textbf{62.2} & \cellcolor{r99}62.0 & \cellcolor{r00}45.1 & \cellcolor{r99}\textbf{71.3} & \cellcolor{r44}56.5 & \cellcolor{r00}36.3  & \cellcolor{r99}\textbf{75.6} & \cellcolor{r94}73.4 \\
			& mix     & \cellcolor{r60}68.6 & \cellcolor{r00}58.8 & \cellcolor{r99}\textbf{75.0} & \cellcolor{r99}\textbf{70.4} & \cellcolor{r00}41.7 & \cellcolor{r76}63.5 & \cellcolor{r00}46.2  & \cellcolor{r21}48.8 & \cellcolor{r99}\textbf{58.8} \\ \hline \hline
			\multirow{3}{*}{\rotatebox{90}{mIoU}} 
			& SKITTI  & \cellcolor{r99}30.4 & \cellcolor{r00}16.4 & \cellcolor{r99}\textbf{30.5} & \cellcolor{r99}\textbf{37.9} & \cellcolor{r00}5.0  & \cellcolor{r56}23.5 & \cellcolor{r99}\textbf{17.0} & \cellcolor{r00}6.0  & \cellcolor{r91}16.1 \\
			& SPOSS   & \cellcolor{r00}12.7 & \cellcolor{r99}\textbf{20.1} & \cellcolor{r91}19.5 & \cellcolor{r00}6.6  & \cellcolor{r99}\textbf{29.8} & \cellcolor{r56}19.6 & \cellcolor{r00}8.8  & \cellcolor{r99}\textbf{28.6} & \cellcolor{r61}20.9 \\
			& mix     & \cellcolor{r50}24.2 & \cellcolor{r00}19.7 & \cellcolor{r99}\textbf{28.6} & \cellcolor{r99}\textbf{21.2} & \cellcolor{r00}15.6 & \cellcolor{r82}20.2 & \cellcolor{r45}14.2 & \cellcolor{r00}10.4 & \cellcolor{r99}\textbf{18.9} \\ \hline
		\end{tabular}
		\begin{tablenotes}
			\item[1] SKITTI denotes SemanticKITTI, SPOSS denotes SemanticPOSS.
			\item[2] IoU of some dominant categories. Deeper color means the better performance on a specific test scene using a model.
		\end{tablenotes}
	\end{threeparttable}
	\vspace{-4mm}
\end{table*}



\subsection{Results}
The results of Experiment 1,2 are shown in Table \ref{tab:5} and Table \ref{tab:6}. Because SemanticKITTI is much larger than SemanticPOSS, we add different weights when calculating the mIoU of the mixed dataset to balance the data size bias. 
\textcolor{red}{ The IoU of the mixed dataset is given by
\vspace{-1mm}
\begin{equation}
IoU_{mix} = \frac{w_{SK}\cdot TP_{SK}+w_{SP}\cdot TP_{SP}}{w_{SK}(|SK|-TN_{SK})+w_{SP}(|SP|-TN_{SP})}
\end{equation}
where $TP_{SK}$,$TN_{SK}$ denote the number of true positive, true negative predictions of SemanticKITTI data, and $|SK|$,$|SP|$ denote the number of LiDAR scans of SemanticKITTI and SemanticPOSS. The weights are inversely proportional to the dataset size.
\vspace{-2mm}
\begin{equation}
w_{SK} = \frac{|SP|}{|SK|+|SP|}, w_{SP} = \frac{|SK|}{|SK|+|SP|}
\end{equation}
}

In Table \ref{tab:5} and Table \ref{tab:6}, experimental performances are colorized in units of 3$\times$3 blocks. In each 3$\times$3 block, the best result in each row is marked as the deepest red, and the worst is white. The medium results are colored depending on their distance to the best one. Let us see the left bottom 3$\times$3 block in Table \ref{tab:5} as an example, i.e., mIoU of different models based on PointNet++. In the first row, 69.6 indicates that the model trained on the urban set performs the best on the urban test set, and the model trained on the rural set achieves the worst mIoU (62.0) on the urban test set. For both Table \ref{tab:5} and Table \ref{tab:6}, we can see a specific model’s performance on different test scenes from a column view, and the performances of different models on a specific test scene from a row view.

To compare the general performance and robustness of a specific model, we use the box plot shown in Fig. \ref{fig:12}. Each box shows the maximum, minimum and median of the mIoU on different test sets. A higher position of a box indicates that the model performs relatively better, and a shorter length of a box indicates that the model performs relatively robustly. 

The results of Experiment 3 are shown in Table \ref{tab:7} and Fig. \ref{fig:14}(a). The curve of Fig. \ref{fig:14}(a) shows the performance trends of the models changing along with the training data size.

\subsection{Discussions}
We will answer the question of whether deep learning methods are data hungry for 3D semantic segmentation from two aspects, scene diversity and dataset size.

\subsubsection{Scene diversity}
From the results of Experiment 1 and Experiment 2, we can summarize our findings as follows:

{\bf - Performance decrease occurs when testing a model at scenes much different from the training scenes.} As shown in Table \ref{tab:5}, all models trained on rural scenes have a performance decrease when testing on urban scenes, and vice versa. This is probably caused by the high scene diversity between urban and rural scenes.
A similar phenomenon appears in experiments with SemanticKITTI and SemanticPOSS, as shown in Table \ref{tab:6}. Apart from the mixed dataset, mIoU always decreases when testing on different datasets with the training dataset. Both Table \ref{tab:5} and Table \ref{tab:6} show deeper color on the diagonal of each 3$\times$3 block, which indicates better performance on similar scenes.
	
{\bf - Preponderant categories are easier to distinguish.} A specific example is the \textit{high vegetation} in Table \ref{tab:5}. Models trained on rural scenes are good at classifying \textit{high vegetation} because of its preponderance in rural scenes.
Table \ref{tab:6} shows a similar phenomenon. For example, SemanticPOSS has a higher density of \textit{person} than SemanticKITTI but fewer samples of other categories. As a result, models trained on SemanticPOSS achieve better performance on \textit{person}, but are generally weaker on other categories.

{\bf - High scene diversity of a training set can improve the robustness of the model.} As shown in Table \ref{tab:5} and Table \ref{tab:6}, models training on the mixed scenes obtain acceptable predictions regardless of the test scenes. Fig. \ref{fig:12} summarizes different model performances using boxplots. Obviously, the mixed model has shorter boxes, which means a narrow distribution of minimum/maximum mIoU in general, showing its more stable performance and robustness. 

\textcolor{red}{{\bf - Mixing multiple datasets in training may not improve model performance.} On the other hand, through Table \ref{tab:5} and Table \ref{tab:6} we can find that the performances of models training on the mixed scenes are not the best on most test scenes. Simply merging more datasets to obtain better scene diversity will face great resistance, and even lead to model performance decreases due to the large domain gap between datasets, which are further discussed in Section \ref{sec:6}. }

{\bf - In summary, the data hunger problem in scene diversity currently exists for 3D LiDAR datasets.} Lack of some specific category or biased category distribution are common phenomena for datasets. For example, Semantic3D does not have dynamic categories such as \textit{person}, making it unsuitable for applications such as autonomous driving systems. 
And a single dataset usually does not have enough scene diversity to obtain a well-generalized model.
\textcolor{red}{ Furthermore, some other factors also affect model performance, such as the sensor difference. The image-based approaches suffer from this effect because of the difference in vertical resolution, as we can find SqueezeSegV2 performances drop more drastically while testing on another dataset in Table \ref{tab:6}.}
Therefore, the data hunger problem in scene diversity is still challenging for 3D LiDAR semantic segmentation models to improve their generalization ability.

\subsubsection{Dataset size}

\begin{table}[h]
	\centering
	\begin{threeparttable}
	\caption{Result of Experiment 3 – Dataset size effects evaluation.}
	\setlength{\tabcolsep}{1.7mm}
	\renewcommand\arraystretch{1.2}
	\label{tab:7}
	\begin{tabular}{c|c|ccccc}
		\hline
		Category & \backslashbox{Model}{size} & \textbf{12.50\%} & \textbf{25\%}  & \textbf{50\%}  & \textbf{75\%}  & \textbf{100\%} \\ \hline
		\multirow{3}{*}{\rotatebox{0}{person}}
		& PointNet++   & 0.3  & 0.3  & 0.4  & 0.5  & 0.7  \\
		& SPG          & 0.6  & 0.4  & 1.1  & 1.0  & 2.3  \\
		& SqueezeSegV2 & 1.3  & 3.0  & 6.9  & 7.5  & 15.4 \\ \hline
		\multirow{3}{*}{\rotatebox{0}{vehicle}}
		& PointNet++   & 52.2 & 54.4 & 54.5 & 54.5 & 54.9 \\
		& SPG          & 44.9 & 46.2 & 47.0 & 48.8 & 50.1 \\
		& SqueezeSegV2 & 61.6 & 68.4 & 71.6 & 74.7 & 78.4 \\ \hline
		\multirow{3}{*}{\rotatebox{0}{vegetation}}
		& PointNet++   & 63.3 & 64.7 & 64.1 & 62.8 & 64.0 \\
		& SPG          & 47.4 & 47.9 & 53.1 & 52.6 & 52.7 \\
		& SqueezeSegV2 & 60.2 & 66.6 & 70.8 & 71.7 & 72.6 \\ \hline
		\multirow{3}{*}{\rotatebox{0}{building}}
		& PointNet++   & 59.2 & 61.7 & 61.8 & 60.6 & 61.9 \\
		& SPG          & 33.2 & 36.7 & 34.1 & 36.1 & 38.0 \\
		& SqueezeSegV2 & 57.6 & 64.7 & 68.4 & 69.0 & 69.7 \\ \hline
		\multirow{3}{*}{\rotatebox{0}{ground}}
		& PointNet++   & 63.0 & 63.5 & 64.4 & 65.0 & 65.3 \\
		& SPG          & 45.0 & 44.9 & 50.7 & 51.9 & 52.2 \\
		& SqueezeSegV2 & 78.7 & 79.7 & 86.2 & 88.2 & 88.2 \\ \hline \hline
		\multirow{3}{*}{\rotatebox{0}{mIoU}}
		& PointNet++   & 27.6 & 29.7 & 30.0 & 30.1 & 30.4 \\
		& SPG          & 13.2 & 14.5 & 15.2 & 15.8 & 17.1 \\
		& SqueezeSegV2 & 24.4 & 28.2 & 32.6 & 34.5 & 37.2 \\ \hline
	\end{tabular}
	\begin{tablenotes}
		\item[1] IoU of some dominant categories using different training dataset size. Experiment on SemanticKITTI dataset.
	\end{tablenotes}
	\end{threeparttable}
	\vspace{-4mm}
\end{table}

\begin{figure}[]
	\centering
	\includegraphics[scale=0.26]{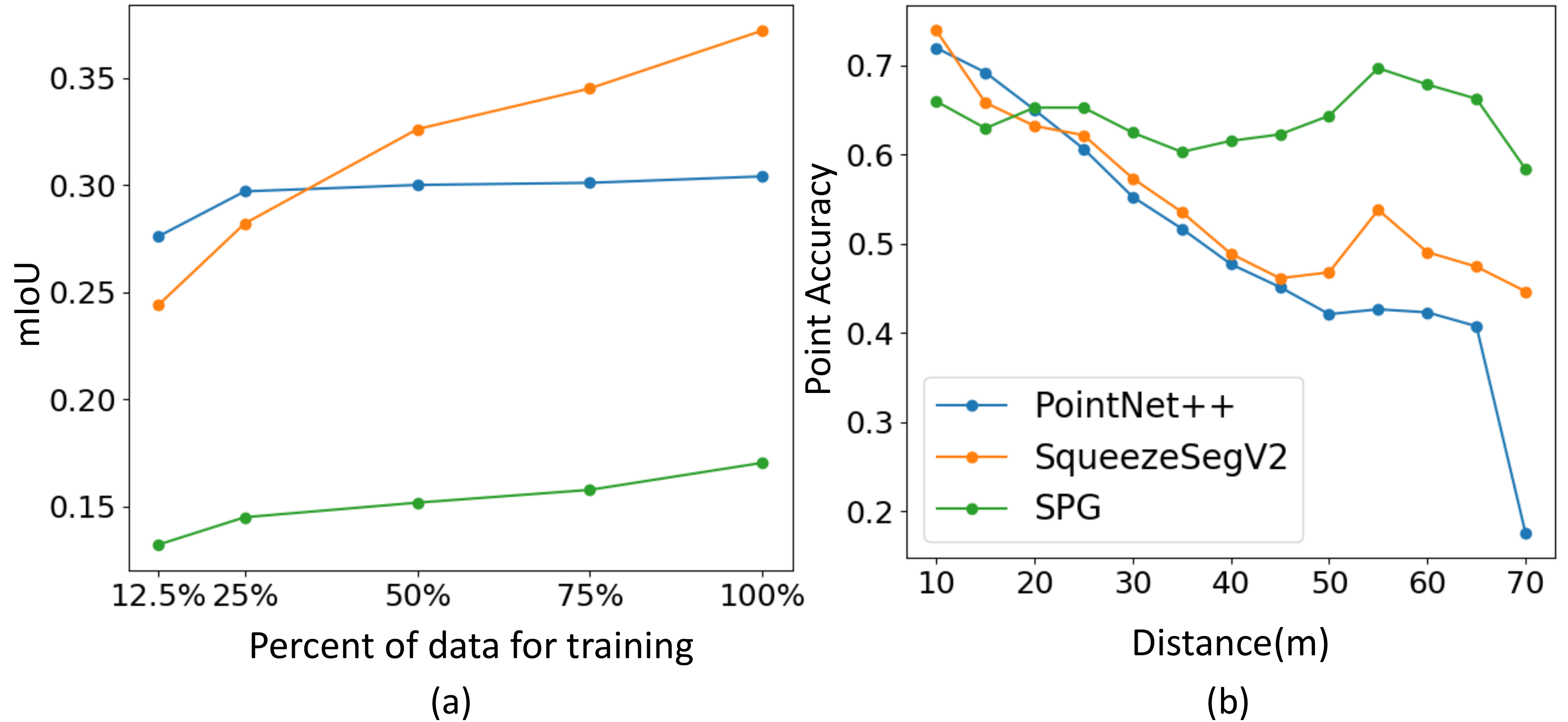}
	\caption{(a) Plots of training DATASET SIZE v.s. MODELS IoU. Experiment on SemanticKITTI dataset. (b) Plots of MODELS accuracy with respect to distance. Experiment on SemanticPOSS dataset.}
	\label{fig:14}
	\vspace{-4mm}
\end{figure}

The results of Experiment 3 illustrate the facts as follows:

{\bf - Increasing training data improves model performance.} All three models show uptrends with increasing training data. It is easy to understand that incremental data provides more features and information for models. 
	
{\bf Second, different models have different sensitivities to the quantity of training data.} As shown in Fig. \ref{fig:14}(a), the uptrend of SqueezeSegV2 is more significant than that of PointNet++. It seems that the data requirements of different models are different. 
SqueezeSegV2 takes range images as input, which are sensitive to the LiDAR's position. Incremental LiDAR frames in a scene captured at different viewpoints may provide more information for range image inputs than point cloud inputs. Therefore, the curve of PointNet++ seems to be saturated with 25\% training data, while the curve of SqueezeSegV2 maintains its growing trend.
	
{\bf - In summary, the data hunger problem for dataset size exists for current 3D LiDAR datasets.} The continuous uptrend of the mIoU-size curve indicates that the model does not reach limit of its ability and requires more data for improvement. It can be predicted that the mIoU will continue increasing if more training data are used. 
All three models show a continuous uptrend to different degrees when adding training data. Therefore, for most deep learning models, existing datasets are not sufficient.

\begin{figure}[t]
	\centering
	\includegraphics[scale=0.27]{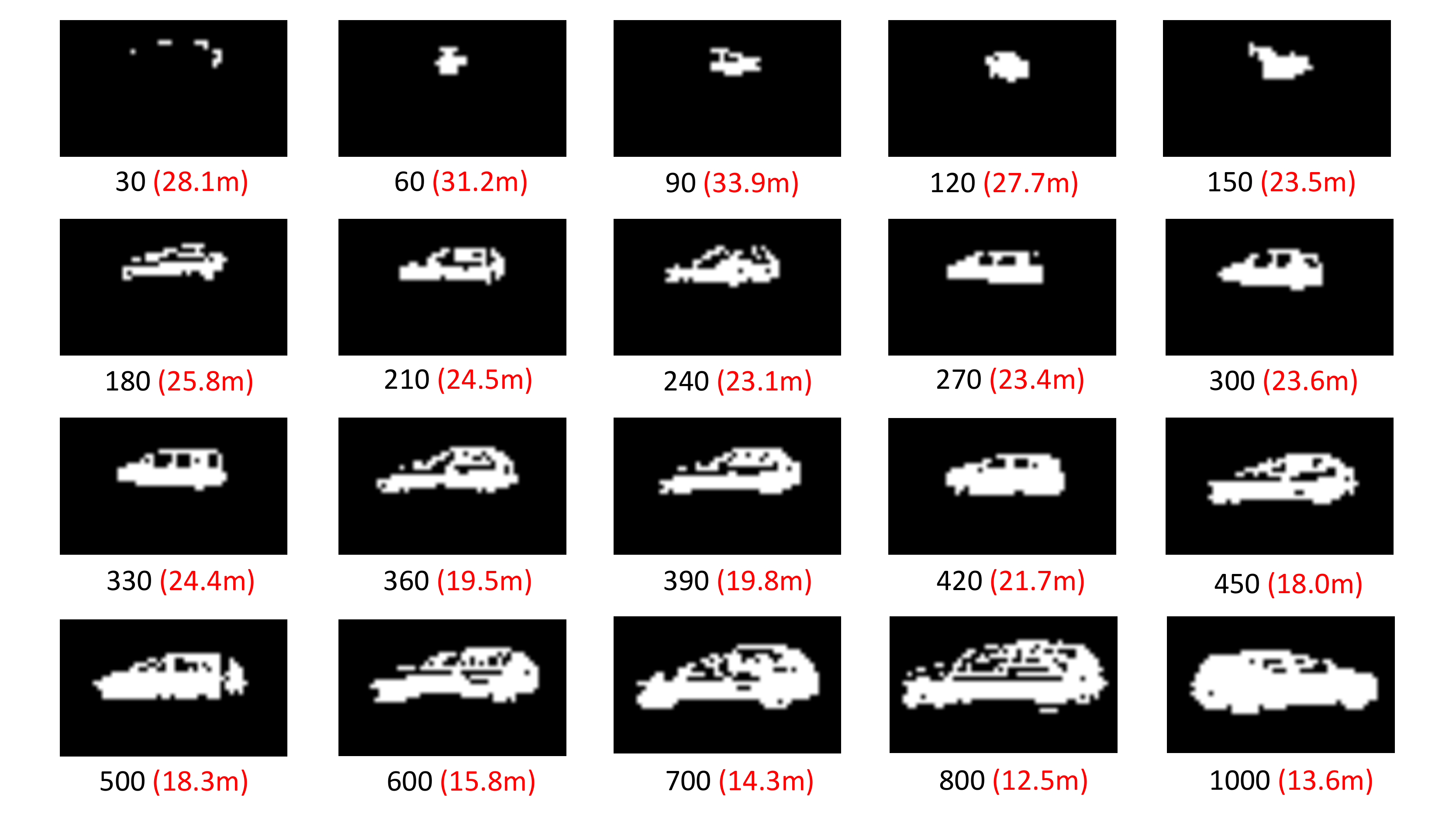}
	\caption{Car instances with different point number in range image. The point numbers of instances are shown in black text and corresponding distances are shown in red text.}
	\label{fig:15}
	\vspace{-4mm}
\end{figure}

\begin{figure}[t]
	\centering
	\includegraphics[scale=0.26]{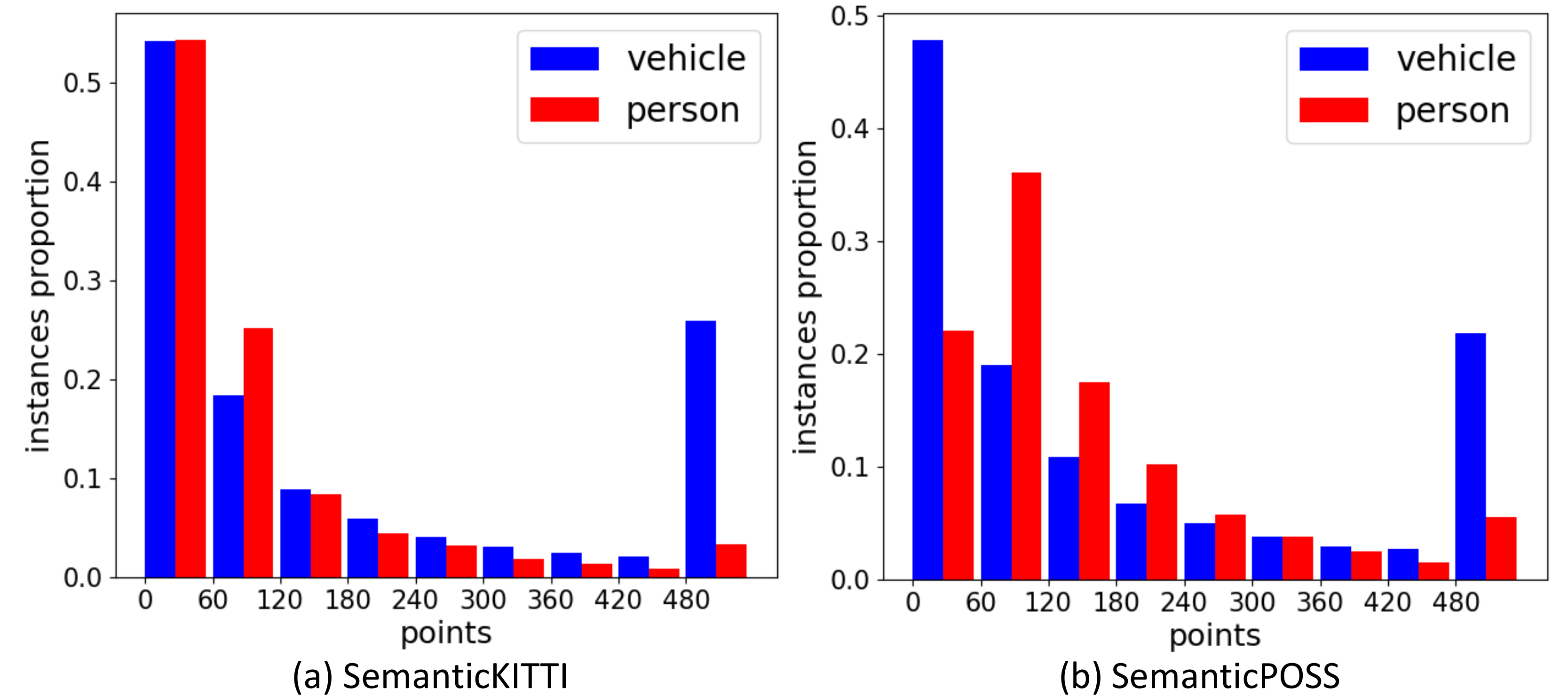}
	\caption{Instances with different point number distribution in SemanticKITTI and SemanticPOSS.}
	\label{fig:16}
	\vspace{-4mm}
\end{figure}

\subsubsection{Instance Distance and Quality}

In 3D datasets, point clouds become sparser with increasing distance to the sensor. Therefore, the points far away from the sensor are hard to be correctly classified. As shown in Fig. \ref{fig:14}(b), the model prediction accuracy decreases with increasing distance, but their downtrends are different. PointNet++ and SqueezeSegV2 show an obvious downtrend, but SPG does not clearly show an accuracy drop.

For an object, the further away from the sensor, the fewer number of points it contains and the higher possibility it will be occluded. Because the features of an object with too few points are vague and confusing, even for a human, it is difficult to definitely distinguish them. Fig. \ref{fig:15} shows some car instances with different points in the range image view. It is difficult to recognize instances with fewer than 150 points or more than 25m. The car features become clear with the increase in points. Therefore, it is reasonable to use the point number as a measurement of instance quality.

We calculate statistics of the point number distribution of \textit{person} and \textit{vehicle} instances in SemanticKITTI and SemanticPOSS, as shown in Fig. \ref{fig:16}. More than 50\% of instances contain fewer than 120 points, which makes no significant contribution to model training. Although it is inevitable for 3D LiDAR datasets to contain these instances, they truly enhance the data hunger problem for data size in the 3D LiDAR semantic segmentation task.

\section{Efforts to Solve the Data Hunger Problem} \label{sec:5}

\begin{figure*}[t]
	\centering
	\includegraphics[scale=0.23]{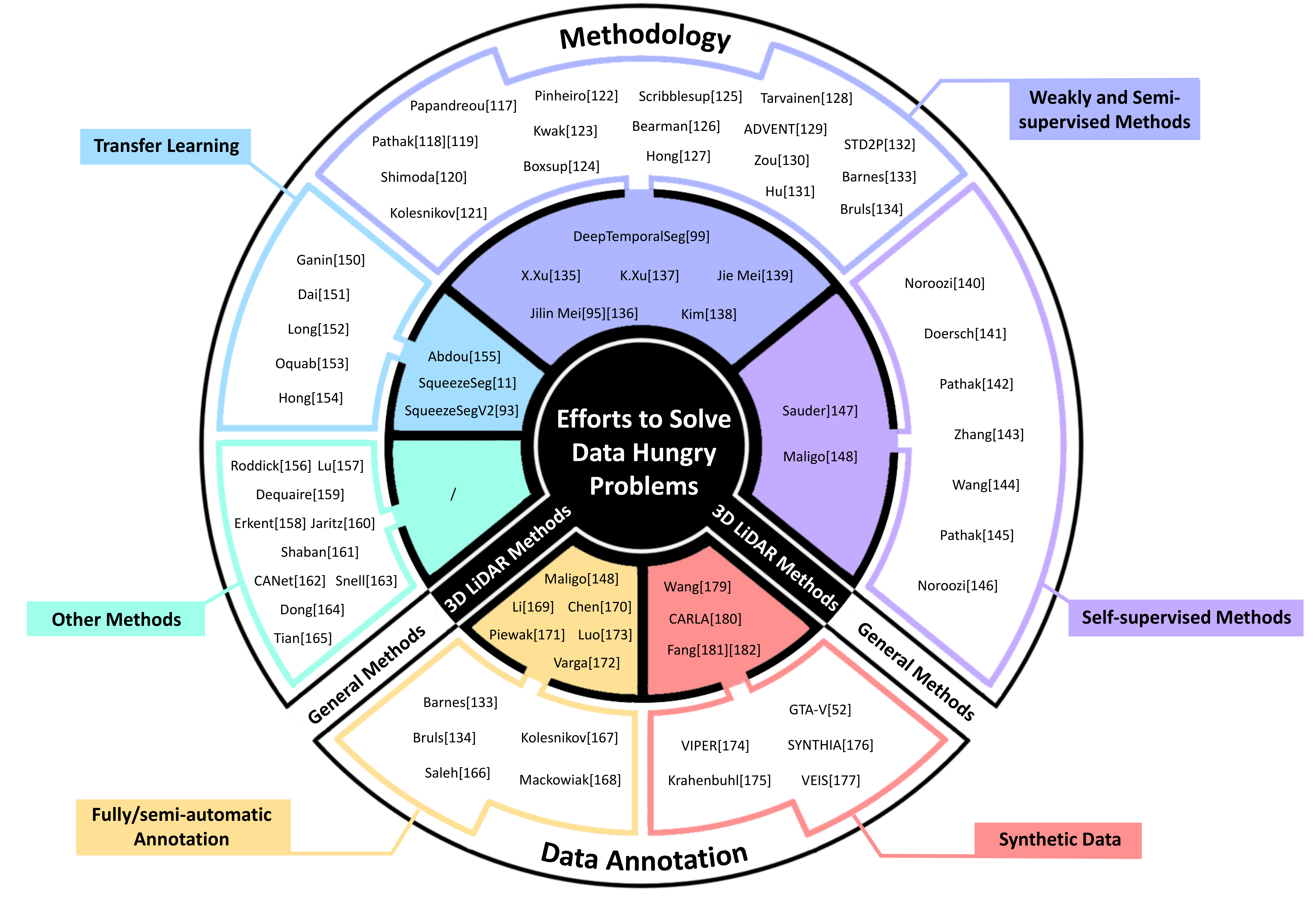}
	\caption{Overview of efforts to solve the data hunger problem. (\textit{General methods}: ideas come from computer vision or machine learning studies but can be heuristic or generalized to solve the data hunger problem of 3D LiDAR semantic segmentation.)}
	\label{fig:17}
	\vspace{-4mm}
\end{figure*}

The data hunger problem is currently a general challenge of deep learning systems \cite{torralba2011unbiased}, where large research efforts have been made for solutions in the fields of machine learning, including computer vision, intelligent vehicles, and robotics. These efforts can be broadly divided into two groups: 1) developing new methodologies that do not require a large quantity of fine annotated data and 2) developing new data annotation methods that are less human intensive. Both efforts can be further divided into two groups: 1) incorporating domain knowledge for 3D LiDAR data processing and 2) general purposes that have proven to be useful on other kinds of data yet have not been adapted on 3D LiDAR applications. Hereinafter, we refer to these two groups as {\bf 3D LiDAR methods} and {\bf general methods} for conciseness. Fig. \ref{fig:17} illustrates the state-of-the-art of these efforts.

\subsection{Methodology}
\subsubsection{Weakly and semi-supervised methods}

Weakly and semi-supervised methods are widely used to solve the data hunger problem with data size. They aim to mine the value of weak supervision as much as possible.

\paragraph{General methods}
Most studies on weakly and semi-supervised semantic segmentation are investigated in the image domain. Due to the existence of large image classification datasets such as ImageNet \cite{deng2009imagenet}, image-level annotations are easy to obtain as weak supervision of semantic segmentation \cite{papandreou2015weakly} \cite{pathak2014fully}. Sometimes, image-level weak supervision is integrated with additional information for better performance, such as prior knowledge of object size \cite{pathak2015constrained}, saliency models indicating object regions \cite{shimoda2016distinct} \cite{kolesnikov2016seed} and super-pixels \cite{pinheiro2015image} \cite{kwak2017weakly}. In addition to image-level labels, bounding boxes \cite{papandreou2015weakly} \cite{dai2015boxsup}, scribbles \cite{lin2016scribblesup} and point supervision \cite{bearman2016s} are considered as weak supervision.
\textcolor{red}{Prior knowledge can provide useful constraints of objectness \cite{pinheiro2015image}, class-agnostic shape \cite{hong2015decoupled} or combinations of several priors \cite{pathak2015constrained}.}
Pseudo labeling instinctively uses model predictions to annotate unlabeled data, which is universally applicable for classification \cite{tarvainen2017mean} and semantic segmentation \cite{vu2019advent} \cite{zou2018unsupervised}.
In addition, there are various ideas for weakly and semi-supervised learning, such as spatiotemporal constraints from videos \cite{hu2019learning}, optical flow \cite{he2017std2p} or other modalities, such as GPS \cite{barnes2017find} and LiDAR \cite{bruls2018mark}.

\paragraph{3D LiDAR methods}

For 3D LiDAR semantic segmentation tasks, available weak supervisions are not as abundant as image-related tasks. Xu et al. \cite{Xu2020} used a tiny fraction of points as weak supervision. Mei et al. \cite{mei2019incorporating} automatically generated weak annotations based on prior human domain geometrical knowledge. Pseudo labeling was implemented by Xu et al. \cite{xu2019semantic}, which also introduced spatial relationships to assist the semi-supervised framework.

Additional multi-modal information is helpful for data hunger problem. Kim et al. \cite{kim2018season} focused on a season-invariant semantic segmentation task by fusing images and 3D LiDAR information.

In addition, spatiotemporal constraints could help transit weak supervisions from adjacent points or LiDAR frames. \textcolor{red}{Constraints\cite{mei2018joint}\cite{mei2019semantic} can be integrated to help the model simultaneously consider intraclass compactness and interclass separability.} Dewan et al. \cite{dewan2019deeptemporalseg} proposed a Bayes filter based method using knowledge from previous scans, which makes the sequential predictions more temporally consistent.

\subsubsection{Self-supervised methods}

It is a common choice to pretrain a deep network with large-scale datasets such as ImageNet \cite{deng2009imagenet} before fine-tuning it to specific visual tasks. \textcolor{red}{However, when facing the data hunger of large-scale datasets, self-supervised learning methods can play a role}.

\paragraph{General methods}
Generally, models are trained on pretext tasks to learn meaningful representations related to the target task without any human annotations. 
Some typical pretext tasks include context prediction \cite{noroozi2016unsupervised} \cite{doersch2015unsupervised}, inpainting \cite{pathak2016context}, colorization \cite{zhang2016colorful} and temporal correlation \cite{wang2015unsupervised} \cite{pathak2017learning}.
\textcolor{red}{Although researchers have designed various pretext tasks, self-supervision performance is still not equal to pretraining. Several studies\cite{noroozi2018boosting} have been made to overcome this gap.}

\paragraph{3D LiDAR methods}
Inspired by the context prediction pretext task for image semantic segmentation, Sauder et al. \cite{sauder2019self} attempted to learn from the spatial distribution by predicting randomly rearranged voxels. 

In addition to learning from pretext tasks, self-supervised methods can also be implemented for clustering points with similar semantic information or common features. Maligo and Lacroix \cite{maligo2016classification} classified point clouds into a large set of categories through self-supervised Gaussian mixture models and annotators can simply assign semantic labels to these categories instead of point-level annotation.

\subsubsection{Transfer learning}

\textcolor{red}{The data hunger problem is not only reflected on datasets size. Diversity between different application scenarios also prevents models generalization performance. Transfer learning\cite{10.1007/978-3-030-01424-7_27} is one approach to handle this problem.}
\textcolor{red}{
\paragraph{General methods}
Transfer learning methods utilize knowledge from a known source domain to new target domains. Based on the techniques, these methods can be categorized to several groups, such as adversarial-based methods\cite{ganin2016domain}, instances-based methods\cite{dai2007boosting}, mapping-based methods\cite{long2015learning} and network-based methods\cite{oquab2014learning}.
It has been applied to many visual applications including semantic segmentation\cite{hong2016learning}.
}

\paragraph{3D LiDAR methods}

Transfer learning can help transfer knowledge from other domains to reduce the data demand of 3D LiDAR. Wu et al. \cite{wu2018squeezeseg} attempted to obtain extra training data from a LiDAR simulator GTA-V. To make the synthetic data more realistic, they transferred the noise distribution of KITTI data to synthesized data.
In \cite{wu2019squeezesegv2}, they proposed an upgrade version for domain shift. With knowledge transferred from the real world, models trained on synthetic data can even outperform baselines trained on real datasets.

Imbalanced categories distribution is one reflection of data hunger problem. It is usually difficult to classify non-dominant categories because of their rare appearance. Abdou et al. \cite{abdou2019end} proposed a weighted self-incremental transfer learning method, which re-weighted the loss function and trained non-dominant categories preferentially.

\textcolor{red}{
\subsubsection{Other methods}
Semantic segmentation can be regarded as one form of scene understanding. In addition, there are many studies tackling this problem from different view.
Semantic occupancy map is one of them that effectively combines 3D and 2D formats, and obtains pixel-wise scene understanding from map level\cite{roddick2020predicting}\cite{lu2019monocular}. By the way, utilizing Bayesian filter and combining temporal information\cite{erkent2018semantic}\cite{dequaire2018deep} could reduce the effect of data hunger.
Multi-modality is another way that fusing 2D and 3D data for usage\cite{erkent2018semantic}\cite{jaritz2019xmuda} to reduce the data hunger effect.
}
\textcolor{red}{
Few-shot learning aims to generalize to new tasks with only a few annotations and prior knowledge. It was introduced by Shaban et al. \cite{shaban2017one} for one-way semantic segmentation\cite{zhang2019canet}. After that, N-way semantic segmentation was explored by prototype framework \cite{snell2017prototypical}\cite{dong2018few} and optimization-based method \cite{tian2019differentiable}. 
}

\subsection{Data Annotation}
\subsubsection{Fully/semi-automatic annotation}
The high cost of pixel/point-level human annotations is one of the most important factors causing the data hunger problem. Many studies have focused on obtaining cheaper annotations by designing fully or semi-automatic annotation methods.

\paragraph{General methods}

Fully automatic annotation often uses labels from additional sensors, such as drivable paths from GPS \cite{barnes2017find} and road marking annotations from LiDAR’s intensity channel \cite{bruls2018mark}.

\textcolor{red}{Some semi-automatic annotation methods attempt to obtain dense segmentation masks through simple clicks based on objectness priors \cite{saleh2016built} and image regions \cite{kolesnikov2016improving}.} 
Mackowiak et al. \cite{mackowiak2018cereals} utilized the active learning idea, which makes the model select worthiest regions for hand labeling, which significantly reduces the annotation cost.

\paragraph{3D LiDAR methods}

Fine annotations in 3D space are much more time-consuming.
Methods that attempt to obtain fully automatic annotations usually depend on prior knowledge of geometry or labels from other sensors. Li et al. \cite{li2016classification} used a decision-tree model to integrate prior knowledge among different categories and generate initialized training labels of object segments. 
Besides the methods based on prior knowledge, several studies \cite{chen2019image} \cite{piewak2018boosting} \cite{varga2017super} introduced labels from camera data helping 3D LiDAR annotations.

In addition, some semi-automatic methods attempted to make 3D LiDAR annotation easier and faster. Luo et al. \cite{luo2018semantic} introduced an active learning framework incorporating neighbor-consistency priors to create a minimal manually annotated training set. As a result, only a few supervoxels need to be annotated. 
Maligo et al. \cite{maligo2016classification} tried to cluster points into a large set of categories before manual annotation. Human annotators only need to group these categories and give them semantic labels through only a few shots.

\subsubsection{Synthetic data}

With the impressive progress of computer graphics, synthetic data have become an alternative to expensive and time-consuming manual annotations.

\paragraph{General methods}

Video games are usually the first choice for synthetic data collection. Richter et al. built synthetic datasets called GTA-V \cite{richter2016playing} and VIPER \cite{richter2017playing} based on the commercial game engine Grand Theft Auto V. VIPER provides the ground truth of image semantic segmentation, instance segmentation, 3D scene layout, visual odometry and optical flow. Krahenbuhl \cite{krahenbuhl2018free} extended the data collection across three video games with more diverse scenarios.

Scenarios based on video games only provide limited freedom for customization. To overcome this drawback, SYNTHIA \cite{ros2016synthia} and VEIS \cite{saleh2018effective} used the Unity3D \cite{Unity3D} development platform to design urban structures and add objects optionally.

\paragraph{3D LiDAR methods}

Several attempts have been made to obtain point-level labels of 3D LiDAR data from simulators. Wang et al. \cite{wang2019automatic} proposed a pipeline for automatic generation of simulated 3D LiDAR with point-level labels. It is based on CARLA \cite{dosovitskiy2017carla}, a simulator for autonomous driving. There are some other general synthetic datasets, such as SYNTHIA \cite{ros2016synthia} and \cite{krahenbuhl2018free}, which are based on video games. They are not designed for obtaining synthetic 3D LiDAR data, but the annotations of depth images may be exploited for the acquisition of 3D LiDAR data.

Except for acquiring annotations from thorough synthetic scenes, Fang et al. \cite{fang2020augmented} proposed a novel LiDAR simulator that augments real scene points with synthetic obstacles. Furthermore, real traffic flows \cite{fang2018simulating} can be placed in different street view background, which is inspiring for generating synthetic data that looks real.

Synthetic data provide economical supplements for data hunger, but domain adaptation techniques are still needed when applied in the real world.

\section{Future Works and Discussion} \label{sec:6}

The "data hunger" problem is increasingly being recognized as a serious and widespread challenge for 3D LiDAR semantic segmentation. However, solutions for the problem have still been a largely underexplored domain compared with studies in computer vision and machine learning. Developing new methods that rely less on fine annotated 3D LiDAR data and developing more diversified 3D LiDAR datasets could become two main directions to focus on. Below, we elaborate on future works in these directions, followed by a discussion on the open questions, which leads to important, but until now, little studied topics.

\subsection{Methodology}
Compared to the vast number of methods using visual images, studies on 3D LiDAR data are very limited in both breadth and depth, and are usually sporadic, premature and unsystematic.
Some new tendencies, such as few-shot learning, to the best of our knowledge, have not been attempted on 3D LiDAR data for semantic segmentation or relevant tasks. Below, we discuss the potential future topics on these aspects.

\subsubsection{Bounding boxes}
The bounding box has been used as a weak supervision signal in developing many semantic segmentation methods of visual images  \cite{dai2015boxsup}\cite{papandreou2015weakly}, and bounding boxes are actually available in many open datasets \cite{geiger2012we} \cite{patil2019h3d} \cite{caesar2019nuscenes} \cite{sun2019scalability}, as reviewed in Table \ref{tab:1}. Can we make use of this information in processing 3D LiDAR data? The idea of solving the data hunger problem in 3D LiDAR seems to be absent in the efforts until now.

\subsubsection{Prior knowledge}
Different from visual images, 3D LiDAR captures real-world data of true physical size and spatial geometry. Many prior knowledge can be used. For example, an approximate elevation of the ground surface and rough size of objects. These prior knowledge can greatly help to save learning costs.

\subsubsection{Spatial-temporal constraints}

Semantic segmentation of video has been studied in computer vision and multimedia societies \cite{garcia2018survey}; however, 3D LiDAR data have been mostly processed frame by frame, which ignores temporal continuity and coherence cues. It can help to save computation time, improve accuracy and reduce the needs of fine annotated data.

\subsubsection{Self-supervised learning}
To make use of the large quantity of unlabeled data, self-supervised learning utilizes pretext tasks. Such an idea is still rare in 3D LiDAR processing, whereas various pretext tasks have been designed on images to learn meaningful representations. It may inspire new directions in the 3D LiDAR domain.

\begin{figure}[]
	\centering
	\includegraphics[scale=0.22]{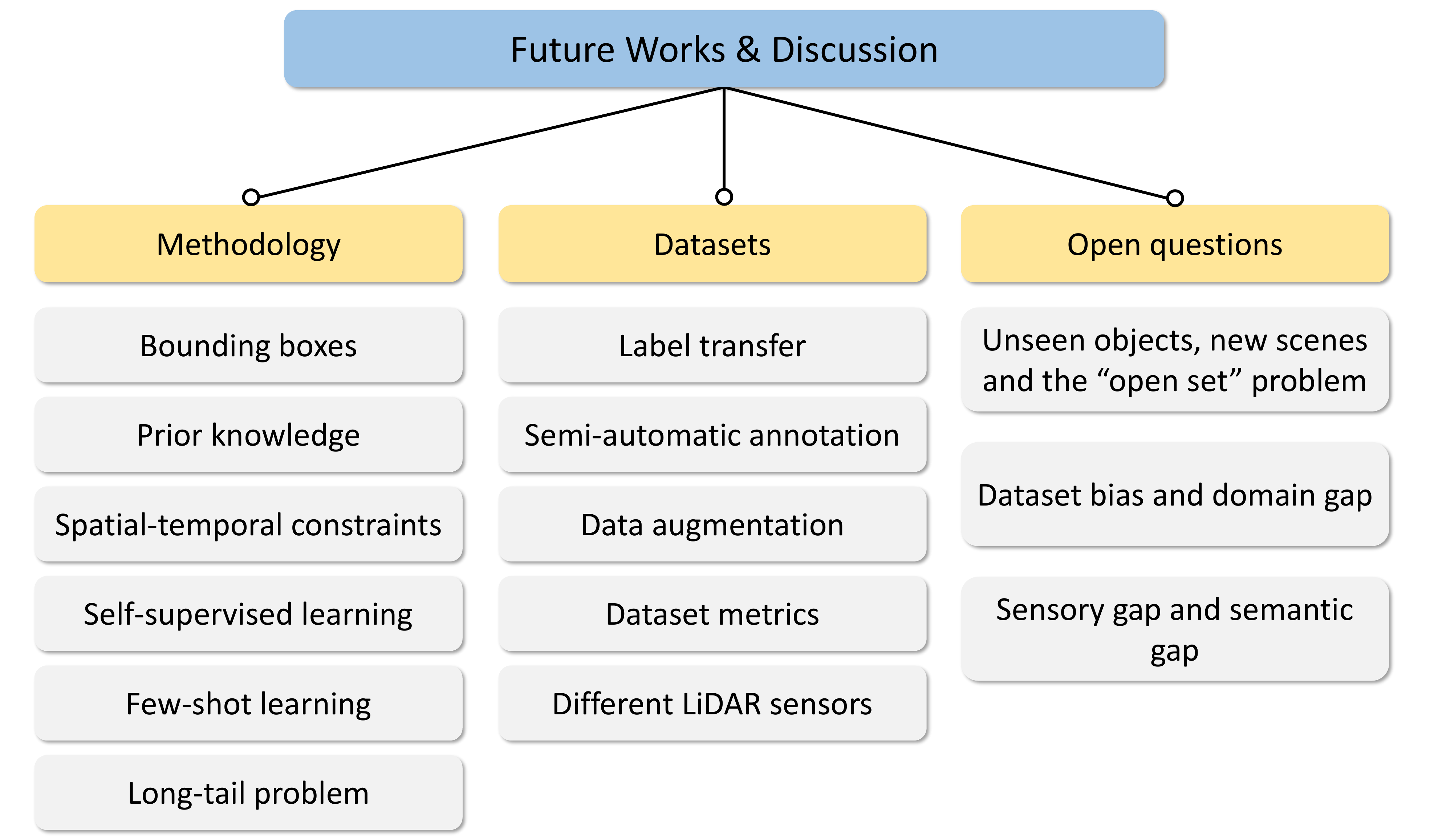}
	\caption{Overview of future works and open questions.}
	\label{fig:18}
	\vspace{0mm}
\end{figure}

\subsubsection{Few-shot learning}

Few-shot learning generalizes new tasks from a few supervisions and prior knowledge, as a human being does. It has even moved forward to zero-shot semantic segmentation\cite{bucher2019zero}, which transfers semantic similarities from linguistic features.
The idea of few-shot learning is inspiring, but the studies are still in the early stages and has not been applied to 3D LiDAR data.

\textcolor{red}{
\subsubsection{The long-tail problem}
3D LiDAR datasets are long-tailed.
As discussed in Section \ref{sec:2}, \textit{road}, \textit{building} and \textit{vegetation} occupy large proportions, whereas \textit{people}, \textit{car} and \textit{riders} are rare and thus difficult to model.
Testing on such datasets reveals the performance of the major categories, while those of the minor ones are often overlooked.  
Long-tail has been a general problem in machine learning that has been studied extensively \cite{wang2017learning}.
The problem is even severe of 3D LiDAR datasets due to the way of data acquisition and object occupation at natural scenes, whereas it has been few addressed in 3D semantic segmentation studies.
}

\subsection{Datasets}

The superior performance of deep learning methods is usually established on a large quantity of fine annotated datasets. However, currently, 3D LiDAR datasets are very limited in terms of both size and diversity.
\textcolor{red}{
How can data generation be more efficient and less labor intensive? More research is needed, where some potential topics are discussed below.}

\subsubsection{Label transfer}
Label transfer is a cheap solution to obtain more 3D LiDAR annotations, which borrows labels from other modalities, such as images\cite{piewak2018boosting} \cite{varga2017super} \cite{chen2019image}.
However, existing methods usually project image results to 3D LiDAR data, where both correct and incorrect annotations are transferred. Error detection mechanisms are needed in this procedure, where prior knowledge, such as the size, geometry, spatial and temporal coherence of 3D LiDAR data, could be important cues to filter false alarms.

\subsubsection{Semi-automatic annotation}
Semiautomatic annotation is an attempt to find a balance between the efficiency and quality of data generation. Some studies \cite{Zimmer2019} use semi-automatic methods to accelerate the labeling of 3D bounding boxes.
Semi-automatic annotation for point-level semantic segmentation is more complicated. Some CRF-based methods \cite{LIM2009701} are developed, but the cost improvements are still limited. In the future, techniques such as active learning \cite{Luo2018} and online learning may further boost the semi-automatic annotation process.

\subsubsection{Data augmentation}
Data augmentation is a commonly used trick for enriching data diversity, which has been proven to achieve gains compared to architectural improvements, both in 2D image \cite{Shorten2019} and 3D LiDAR object detection tasks \cite{Cheng2020}.
However, common augmentation operations such as random flip, rotation and scaling, which are useful for single objects, may cause discordance of context elements for semantic segmentation. 
To the best of our knowledge, there is no systematic research on point-wise data augmentation for 3D LiDAR semantic segmentation, so there is considerable room for growth.
\textcolor{red}{
\subsubsection{Dataset metrics}
It is needed to guide dataset generation, e.g. building datasets of new scenes but avoiding generating similar scenes repetition.
It is needed too to guide dataset-based testing more efficiently.
A dataset is composed of very limited snapshots of real-world scenes. 
It is very important for real-world applications such as autonomous driving to have the knowledge what performance could be achieved at a target scene, what difference of scenes can the model adapt to, 
and what dataset could lead a test result that reflects objectively the performance at real world.
Many metrics have been developed to evaluate model performance, 
while we have found no widely accepted metric of dataset that is able to answer the above needs.
}
\textcolor{red}{
\subsubsection{Different LiDAR sensors}
In recent years, many LiDAR sensors have been developed of different specifications by different manufacturers, and according to system configurations, 3D points are measured from viewpoints at different elevations and angles, which could vary object appearance and category distribution of the datasets largely.
As listed in Table \ref{tab:1}, most datasets are developed by a particular LiDAR sensor of a particular system configuration. 
The challenges from different sensors are not fully recognized now, which makes it hard to train a 3D semantic segmentation model on more datasets, 
and hard to transfer a model to the ones with other types of sensor input. 
These problems need to be addressed in future from both dataset and methodological aspects.
}
\subsection{Open Questions}

\subsubsection{\textcolor{red}{Unseen objects, new scenes, and the "open set" problem}}

Currently, studies mostly design algorithms and evaluate model performance in a "closed set" \cite{Scheirer2013}, which assumes that the testing set obeys the same distribution as the training one. However, real world applications are an "open set" problem, \textcolor{red}{which requires deep models to deal with unseen objects and new scenes.
It has been a general problem of deep models, which will always be data hungry in unseen categories and new scenarios. 
However, the problem is even severe for the applications like autonomous driving, where some categories are unseen or rare in datasets but need to be handled in real applications. 
Therefore data hunger may always exist, it is far more important of the model to be aware of when it is unsure.} 


\subsubsection{\textcolor{red}{Dataset bias and domain gap}}
\textcolor{red}{It is inevitable for actual datasets to be a biased sampling of
the real world. Dataset bias is a general problem of machine learning. 
Many studies have been addressed to explore the bias \cite{torralba2011unbiased} by cross-dataset experiments, undo the damage \cite{khosla2012undoing} by modeling bias vector,
and \cite{tommasi2017deeper} found that many existing ad hoc learning algorithms for undoing the dataset bias do not help for CNN features.
The most concerned dataset bias in 3D semantic segmentation task is domain gap, which exhibits in both model and dataset aspects.
The former is usually solved by transfer learning [ ] to adapt a model from one domain to another. 
However, domain gap among datasets is less explored. 
As discussed in Section \ref{sec:4}, mixing multiple data sets in training may not improve model accuracy, and testing model on a dataset with a large domain gap may also cause a significant performance degradation, 
limiting the usefulness of the test results.
Making use of different dataset needs rigorous study, where dataset metrics to evaluate domain gap quantitatively may be one of the keys.}

\textcolor{red}{
\subsubsection{Sensory gap and semantic gap}
Sensory gap exists between objects in the real world and in data due to sensory limitation, e.g. partial observation, occlusion and sparse point sampling.
Semantic gap means people understand an object based on the knowledge that may beyond those provided by the data, whereas a machine does not.
For example, a tree could be measured on its straight trunk, which looks very similar to a pole from the data appearance at the moment.
In addition, there are many types of bicycles and motorcycles, from small to large scales, 
where sensory data may not be distinguishable on functional categorization, 
and an object of the same functional category may have largely different appearance, and in different countries/regions.
These sensory and semantic gaps bring great difficulties in developing a general category list for dataset sharing, which needs scientific studies at the communities' level.}

\section{Conclusion}   \label{sec:7}
\textcolor{red}{In this research, a comprehensive survey and experimental study is conducted to seek answers to the question: 'Are we hungry for 3D LiDAR data for semantic segmentation?' The following studies are addressed. 
	First, a broad review to the main 3D LiDAR datasets is conducted, followed by a statistical analysis on three representative datasets to gain an in-depth view on the datasets' size and diversity, which are the critical factors in learning deep models. 
	Second, a systematic review to the state-of-the-art 3D semantic segmentation is conducted, followed by experiments and cross examinations of three representative deep learning methods to find out how the size and diversity of the datasets affect deep models' performance. 
	The major findings in the above studies are: }
\begin{enumerate}
	\item \textcolor{red}{Due to the uneven spatial distribution of 3D LiDAR data, large amount of points could be scanned on the same objects that bring few novel information, and large portion of data points belong to the category of the objects nearby sensor's viewpoint, e.g. \textit{road}, the datasets are therefore extremely long-tailed.}
	\item \textcolor{red}{Scenes could be very diverse that are not directly correlated with geographic location. The existing 3D LiDAR datasets reflect only a very small set of real world scenes, whereas they exhibit insufficient inner-dataset diversity, while large cross-dataset difference.}
	\item \textcolor{red}{Due to the large domain dap, mixing multiple datasets in training may not improve model accuracy, and testing model on a dataset with a large domain gap may also cause a significant performance degradation, limiting the usefulness of the test results. }
\end{enumerate}
\textcolor{red}{Finally, a systematic survey to the existing efforts to solve the data hunger problem is conducted on both methodological and dataset's viewpoints, followed by an insightful discussion of remaining problems and open questions leading to potential topics in future works.}

\textcolor{red}{Towards the robotic and autonomous driving applications at complex real-worlds, data hunger may always exist. On one hand, it is important to develop methods rely on less fine annotated data for training, and facing rare/unseen objects, it is far more important of the model to be aware when it's unsure. On the other hand, metrics are needed to measure the domain gap of datasets/scenes, which is needed to guide dataset generation and model testing more efficiently. A general definition of category list is also needed for dataset sharing, which requires scientific study through international collaboration.}

\ifCLASSOPTIONcaptionsoff
  \newpage
\fi

\bibliographystyle{IEEEtran}
\bibliographystyle{unsrt}
\bibliography{refs}

\begin{thebibliography}{100}
\providecommand{\url}[1]{#1}
\csname url@samestyle\endcsname
\providecommand{\newblock}{\relax}
\providecommand{\bibinfo}[2]{#2}
\providecommand{\BIBentrySTDinterwordspacing}{\spaceskip=0pt\relax}
\providecommand{\BIBentryALTinterwordstretchfactor}{4}
\providecommand{\BIBentryALTinterwordspacing}{\spaceskip=\fontdimen2\font plus
\BIBentryALTinterwordstretchfactor\fontdimen3\font minus
  \fontdimen4\font\relax}
\providecommand{\BIBforeignlanguage}[2]{{%
\expandafter\ifx\csname l@#1\endcsname\relax
\typeout{** WARNING: IEEEtran.bst: No hyphenation pattern has been}%
\typeout{** loaded for the language `#1'. Using the pattern for}%
\typeout{** the default language instead.}%
\else
\language=\csname l@#1\endcsname
\fi
#2}}
\providecommand{\BIBdecl}{\relax}
\BIBdecl

\bibitem{thrun2006stanley}
S.~Thrun, M.~Montemerlo, H.~Dahlkamp, D.~Stavens, A.~Aron, J.~Diebel, P.~Fong,
  J.~Gale, M.~Halpenny, G.~Hoffmann \emph{et~al.}, ``Stanley: The robot that
  won the {DARPA} grand challenge,'' \emph{Journal of Field Robotics}, vol.~23,
  no.~9, pp. 661--692, 2006.

\bibitem{patz2008practical}
B.~J. Patz, Y.~Papelis, R.~Pillat, G.~Stein, and D.~Harper, ``A practical
  approach to robotic design for the {DARPA} urban challenge,'' \emph{Journal
  of Field Robotics}, vol.~25, no.~8, pp. 528--566, 2008.

\bibitem{zhang2014loam}
J.~Zhang and S.~Singh, ``{LOAM}: {LiDAR} odometry and mapping in real-time.''
  in \emph{Robotics: Science and Systems}, vol.~2, no.~9, 2014.

\bibitem{hess2016real}
W.~Hess, D.~Kohler, H.~Rapp, and D.~Andor, ``Real-time loop closure in {2D}
  {LiDAR} slam,'' in \emph{2016 IEEE International Conference on Robotics and
  Automation}.\hskip 1em plus 0.5em minus 0.4em\relax IEEE, 2016, pp.
  1271--1278.

\bibitem{li2016vehicle}
B.~Li, T.~Zhang, and T.~Xia, ``Vehicle detection from {3D} {LiDAR} using fully
  convolutional network,'' \emph{arXiv preprint arXiv:1608.07916}, 2016.

\bibitem{chen2017multi}
X.~Chen, H.~Ma, J.~Wan, B.~Li, and T.~Xia, ``Multi-view {3D} object detection
  network for autonomous driving,'' in \emph{IEEE Conference on Computer Vision
  and Pattern Recognition}, 2017, pp. 1907--1915.

\bibitem{hackel2017semantic3d}
T.~Hackel, N.~Savinov, L.~Ladicky, J.~D. Wegner, K.~Schindler, and
  M.~Pollefeys, ``Semantic{3D}.net: A new large-scale point cloud
  classification benchmark,'' \emph{arXiv preprint arXiv:1704.03847}, 2017.

\bibitem{behley2019semantickitti}
J.~Behley, M.~Garbade, A.~Milioto, J.~Quenzel, S.~Behnke, C.~Stachniss, and
  J.~Gall, ``{SemanticKITTI}: A dataset for semantic scene understanding of
  {LiDAR} sequences,'' in \emph{IEEE International Conference on Computer
  Vision}, 2019, pp. 9297--9307.

\bibitem{rusu20113d}
R.~B. Rusu and S.~Cousins, ``{3D} is here: Point cloud library (pcl),'' in
  \emph{2011 IEEE International Conference on Robotics and Automation}.\hskip
  1em plus 0.5em minus 0.4em\relax IEEE, 2011, pp. 1--4.

\bibitem{qi2017pointnet}
C.~R. Qi, H.~Su, K.~Mo, and L.~J. Guibas, ``{PointNet}: Deep learning on point
  sets for {3D} classification and segmentation,'' in \emph{IEEE Conference on
  Computer Vision and Pattern Recognition}, 2017, pp. 652--660.

\bibitem{wu2018squeezeseg}
B.~Wu, A.~Wan, X.~Yue, and K.~Keutzer, ``{SqueezeSeg}: Convolutional neural
  nets with recurrent {CRF} for real-time road-object segmentation from {3D}
  {LiDAR} point cloud,'' in \emph{IEEE International Conference on Robotics and
  Automation}.\hskip 1em plus 0.5em minus 0.4em\relax IEEE, 2018, pp.
  1887--1893.

\bibitem{garcia2017review}
A.~Garcia-Garcia, S.~Orts-Escolano, S.~Oprea, V.~Villena-Martinez, and
  J.~Garcia-Rodriguez, ``A review on deep learning techniques applied to
  semantic segmentation,'' \emph{arXiv preprint arXiv:1704.06857}, 2017.

\bibitem{yu2018methods}
H.~Yu, Z.~Yang, L.~Tan, Y.~Wang, W.~Sun, M.~Sun, and Y.~Tang, ``Methods and
  datasets on semantic segmentation: A review,'' \emph{Neurocomputing}, vol.
  304, pp. 82--103, 2018.

\bibitem{yuxing2019review}
Y.~Xie, J.~Tian, and X.~X. Zhu, ``A review of point cloud semantic
  segmentation,'' \emph{arXiv preprint arXiv:1908.08854}, 2019.

\bibitem{yulan2019review}
Y.~Guo, H.~Wang, Q.~Hu, H.~Liu, L.~Liu, and M.~Bennamoun, ``Deep learning for
  {3D} point clouds: A survey,'' \emph{arXiv preprint arXiv:1912.12033}, 2019.

\bibitem{zhu2016beyond}
H.~Zhu, F.~Meng, J.~Cai, and S.~Lu, ``Beyond pixels: A comprehensive survey
  from bottom-up to semantic image segmentation and cosegmentation,''
  \emph{Journal of Visual Communication and Image Representation}, vol.~34, pp.
  12--27, 2016.

\bibitem{lecun2015deep}
Y.~LeCun, Y.~Bengio, and G.~Hinton, ``Deep learning,'' \emph{Nature}, vol. 521,
  no. 7553, pp. 436--444, 2015.

\bibitem{schmidhuber2015deep}
J.~Schmidhuber, ``Deep learning in neural networks: An overview,'' \emph{Neural
  Networks}, vol.~61, pp. 85--117, 2015.

\bibitem{long2015fully}
J.~Long, E.~Shelhamer, and T.~Darrell, ``Fully convolutional networks for
  semantic segmentation,'' in \emph{IEEE Conference on Computer Vision and
  Pattern Recognition}, 2015, pp. 3431--3440.

\bibitem{chen2014big}
X.-W. Chen and X.~Lin, ``Big data deep learning: challenges and perspectives,''
  \emph{IEEE access}, vol.~2, pp. 514--525, 2014.

\bibitem{torralba2011unbiased}
A.~Torralba and A.~A. Efros, ``Unbiased look at dataset bias,'' in \emph{IEEE
  Conference on Computer Vision and Pattern Recognition}.\hskip 1em plus 0.5em
  minus 0.4em\relax IEEE, 2011, pp. 1521--1528.

\bibitem{sun2017revisiting}
C.~Sun, A.~Shrivastava, S.~Singh, and A.~Gupta, ``Revisiting unreasonable
  effectiveness of data in deep learning era,'' in \emph{IEEE International
  Conference on Computer Vision}, 2017, pp. 843--852.

\bibitem{marcus2018deep}
G.~Marcus, ``Deep learning: A critical appraisal,'' \emph{arXiv preprint
  arXiv:1801.00631}, 2018.

\bibitem{armeni20163d}
I.~Armeni, O.~Sener, A.~R. Zamir, H.~Jiang, I.~Brilakis, M.~Fischer, and
  S.~Savarese, ``{3D} semantic parsing of large-scale indoor spaces,'' in
  \emph{IEEE Conference on Computer Vision and Pattern Recognition}, 2016, pp.
  1534--1543.

\bibitem{everingham2015pascal}
M.~Everingham, S.~A. Eslami, L.~Van~Gool, C.~K. Williams, J.~Winn, and
  A.~Zisserman, ``The {PASCAL} visual object classes challenge: A
  retrospective,'' \emph{International Journal of Computer Vision}, vol. 111,
  no.~1, pp. 98--136, 2015.

\bibitem{deng2009imagenet}
J.~Deng, W.~Dong, R.~Socher, L.-J. Li, K.~Li, and L.~Fei-Fei, ``{ImageNet}: A
  large-scale hierarchical image database,'' in \emph{IEEE Conference on
  Computer Vision and Pattern Recognition}.\hskip 1em plus 0.5em minus
  0.4em\relax Ieee, 2009, pp. 248--255.

\bibitem{nguyen20133d}
A.~Nguyen and B.~Le, ``{3D} point cloud segmentation: A survey,'' in \emph{IEEE
  Conference on Robotics, Automation and Mechatronics}.\hskip 1em plus 0.5em
  minus 0.4em\relax IEEE, 2013, pp. 225--230.

\bibitem{grilli2017review}
E.~Grilli, F.~Menna, and F.~Remondino, ``A review of point clouds segmentation
  and classification algorithms,'' \emph{The International Archives of
  Photogrammetry, Remote Sensing and Spatial Information Sciences}, vol.~42, p.
  339, 2017.

\bibitem{lateef2019survey}
F.~Lateef and Y.~Ruichek, ``Survey on semantic segmentation using deep learning
  techniques,'' \emph{Neurocomputing}, vol. 338, pp. 321--348, 2019.

\bibitem{vodrahalli20173d}
K.~Vodrahalli and A.~K. Bhowmik, ``{3D} computer vision based on machine
  learning with deep neural networks: A review,'' \emph{Journal of the Society
  for Information Display}, vol.~25, no.~11, pp. 676--694, 2017.

\bibitem{ioannidou2017deep}
A.~Ioannidou, E.~Chatzilari, S.~Nikolopoulos, and I.~Kompatsiaris, ``Deep
  learning advances in computer vision with {3D} data: A survey,'' \emph{ACM
  Computing Surveys (CSUR)}, vol.~50, no.~2, pp. 1--38, 2017.

\bibitem{griffiths2019review}
D.~Griffiths and J.~Boehm, ``A review on deep learning techniques for {3D}
  sensed data classification,'' \emph{Remote Sensing}, vol.~11, no.~12, p.
  1499, 2019.

\bibitem{Saifullahi2020review}
S.~A. Bello, S.~Yu, and C.~Wang, ``Review: deep learning on {3D} point
  clouds,'' \emph{arXiv preprint arXiv:2001.06280}, 2020.

\bibitem{feng2020deep}
D.~Feng, C.~Haase-Sch{\"u}tz, L.~Rosenbaum, H.~Hertlein, C.~Glaeser, F.~Timm,
  W.~Wiesbeck, and K.~Dietmayer, ``Deep multi-modal object detection and
  semantic segmentation for autonomous driving: Datasets, methods, and
  challenges,'' \emph{IEEE Transactions on Intelligent Transportation Systems},
  2020.

\bibitem{roynard2018paris}
X.~Roynard, J.-E. Deschaud, and F.~Goulette, ``{Paris-Lille-3D}: A large and
  high-quality ground-truth urban point cloud dataset for automatic
  segmentation and classification,'' \emph{The International Journal of
  Robotics Research}, vol.~37, no.~6, pp. 545--557, 2018.

\bibitem{geiger2012we}
A.~Geiger, P.~Lenz, and R.~Urtasun, ``Are we ready for autonomous driving? the
  {KITTI} vision benchmark suite,'' in \emph{IEEE Conference on Computer Vision
  and Pattern Recognition}.\hskip 1em plus 0.5em minus 0.4em\relax IEEE, 2012,
  pp. 3354--3361.

\bibitem{griffiths2019synthcity}
D.~Griffiths and J.~Boehm, ``Synthcity: A large scale synthetic point cloud,''
  \emph{arXiv preprint arXiv:1907.04758}, 2019.

\bibitem{munoz2009contextual}
D.~Munoz, J.~A. Bagnell, N.~Vandapel, and M.~Hebert, ``Contextual
  classification with functional max-margin markov networks,'' in \emph{IEEE
  Conference on Computer Vision and Pattern Recognition}.\hskip 1em plus 0.5em
  minus 0.4em\relax IEEE, 2009, pp. 975--982.

\bibitem{serna2014paris}
A.~Serna, B.~Marcotegui, F.~Goulette, and J.-E. Deschaud, ``Paris-rue-madame
  database,'' in \emph{International Conference on Pattern Recognition
  Applications and Methods}, ser. ICPRAM 2014.\hskip 1em plus 0.5em minus
  0.4em\relax Setubal, PRT: SCITEPRESS - Science and Technology Publications,
  Lda, 2014, p. 819–824.

\bibitem{vallet2015terramobilita}
B.~Vallet, M.~Br{\'e}dif, A.~Serna, B.~Marcotegui, and N.~Paparoditis,
  ``Terramobilita/iqmulus urban point cloud analysis benchmark,''
  \emph{Computers \& Graphics}, vol.~49, pp. 126--133, 2015.

\bibitem{gehrung2017approach}
J.~Gehrung, M.~Hebel, M.~Arens, and U.~Stilla, ``An approach to extract moving
  objects from mls data using a volumetric background representation,''
  \emph{ISPRS Annals of the Photogrammetry, Remote Sensing and Spatial
  Information Sciences}, vol.~4, p. 107, 2017.

\bibitem{de2013unsupervised}
M.~De~Deuge, A.~Quadros, C.~Hung, and B.~Douillard, ``Unsupervised feature
  learning for classification of outdoor {3D} scans,'' in \emph{Australasian
  Conference on Robitics and Automation}, vol.~2, 2013, p.~1.

\bibitem{pan2020semanticposs}
Y.~Pan, B.~Gao, J.~Mei, S.~Geng, C.~Li, and H.~Zhao, ``{SemanticPOSS}: A point
  cloud dataset with large quantity of dynamic instances,'' \emph{arXiv
  preprint arXiv:2002.09147}, 2020.

\bibitem{geyer2020a2d2}
J.~Geyer, Y.~Kassahun, M.~Mahmudi, X.~Ricou, R.~Durgesh, A.~S. Chung,
  L.~Hauswald, V.~H. Pham, M.~M{\"u}hlegg, S.~Dorn \emph{et~al.}, ``{A2Ds2}:
  {Audi} autonomous driving dataset,'' \emph{arXiv preprint arXiv:2004.06320},
  2020.

\bibitem{patil2019h3d}
A.~Patil, S.~Malla, H.~Gang, and Y.-T. Chen, ``The {H3D} dataset for
  full-surround {3D} multi-object detection and tracking in crowded urban
  scenes,'' in \emph{International Conference on Robotics and
  Automation}.\hskip 1em plus 0.5em minus 0.4em\relax IEEE, 2019, pp.
  9552--9557.

\bibitem{caesar2019nuscenes}
H.~Caesar, V.~Bankiti, A.~H. Lang, S.~Vora, V.~E. Liong, Q.~Xu, A.~Krishnan,
  Y.~Pan, G.~Baldan, and O.~Beijbom, ``{nuScenes}: A multimodal dataset for
  autonomous driving,'' \emph{arXiv preprint arXiv:1903.11027}, 2019.

\bibitem{lyft2019}
R.~Kesten, M.~Usman, J.~Houston, T.~Pandya, K.~Nadhamuni, A.~Ferreira, M.~Yuan,
  B.~Low, A.~Jain, P.~Ondruska, S.~Omari, S.~Shah, A.~Kulkarni, A.~Kazakova,
  C.~Tao, L.~Platinsky, W.~Jiang, and V.~Shet, ``Lyft level 5 perception
  dataset 2020,'' \url{https://level5.lyft.com/dataset/}, 2019.

\bibitem{chang2019argoverse}
M.-F. Chang, J.~Lambert, P.~Sangkloy, J.~Singh, S.~Bak, A.~Hartnett, D.~Wang,
  P.~Carr, S.~Lucey, D.~Ramanan \emph{et~al.}, ``Argoverse: {3D} tracking and
  forecasting with rich maps,'' in \emph{IEEE Conference on Computer Vision and
  Pattern Recognition}, 2019, pp. 8748--8757.

\bibitem{sun2019scalability}
P.~Sun, H.~Kretzschmar, X.~Dotiwalla, A.~Chouard, V.~Patnaik, P.~Tsui, J.~Guo,
  Y.~Zhou, Y.~Chai, B.~Caine \emph{et~al.}, ``Scalability in perception for
  autonomous driving: Waymo open dataset,'' \emph{arXiv}, pp. arXiv--1912,
  2019.

\bibitem{pham20203d}
Q.-H. Pham, P.~Sevestre, R.~S. Pahwa, H.~Zhan, C.~H. Pang, Y.~Chen, A.~Mustafa,
  V.~Chandrasekhar, and J.~Lin, ``{A*3D} dataset: Towards autonomous driving in
  challenging environments,'' in \emph{IEEE International Conference on
  Robotics and Automation}.\hskip 1em plus 0.5em minus 0.4em\relax IEEE, 2020,
  pp. 2267--2273.

\bibitem{bijelic2020seeing}
M.~Bijelic, T.~Gruber, F.~Mannan, F.~Kraus, W.~Ritter, K.~Dietmayer, and
  F.~Heide, ``Seeing through fog without seeing fog: Deep multimodal sensor
  fusion in unseen adverse weather,'' in \emph{IEEE Conference on Computer
  Vision and Pattern Recognition}, 2020, pp. 11\,682--11\,692.

\bibitem{richter2016playing}
S.~R. Richter, V.~Vineet, S.~Roth, and V.~Koltun, ``Playing for data: Ground
  truth from computer games,'' in \emph{European Conference on Computer
  Vision}.\hskip 1em plus 0.5em minus 0.4em\relax Springer, 2016, pp. 102--118.

\bibitem{cordts2016cityscapes}
M.~Cordts, M.~Omran, S.~Ramos, T.~Rehfeld, M.~Enzweiler, R.~Benenson,
  U.~Franke, S.~Roth, and B.~Schiele, ``The {Cityscapes} dataset for semantic
  urban scene understanding,'' in \emph{IEEE Conference on Computer Vision and
  Pattern Recognition}, 2016, pp. 3213--3223.

\bibitem{silberman2012indoor}
N.~Silberman, D.~Hoiem, P.~Kohli, and R.~Fergus, ``Indoor segmentation and
  support inference from {RGBD} images,'' in \emph{European Conference on
  Computer Vision}.\hskip 1em plus 0.5em minus 0.4em\relax Springer, 2012, pp.
  746--760.

\bibitem{dai2017scannet}
A.~Dai, A.~X. Chang, M.~Savva, M.~Halber, T.~Funkhouser, and M.~Nie{\ss}ner,
  ``{ScanNet}: Richly-annotated {3D} reconstructions of indoor scenes,'' in
  \emph{IEEE Conference on Computer Vision and Pattern Recognition}, 2017, pp.
  5828--5839.

\bibitem{huang2018apolloscape}
X.~Huang, P.~Wang, X.~Cheng, D.~Zhou, Q.~Geng, and R.~Yang, ``The {ApolloScape}
  open dataset for autonomous driving and its application,'' \emph{arXiv
  preprint arXiv:1803.06184}, 2018.

\bibitem{hu2019randla}
Q.~Hu, B.~Yang, L.~Xie, S.~Rosa, Y.~Guo, Z.~Wang, N.~Trigoni, and A.~Markham,
  ``{RandLA-Net}: Efficient semantic segmentation of large-scale point
  clouds,'' \emph{arXiv preprint arXiv:1911.11236}, 2019.

\bibitem{kirillov2019panoptic}
A.~Kirillov, K.~He, R.~Girshick, C.~Rother, and P.~Doll{\'a}r, ``Panoptic
  segmentation,'' in \emph{IEEE Conference on Computer Vision and Pattern
  Recognition}, 2019, pp. 9404--9413.

\bibitem{pandora}
Hesaitech.com, ``Pandora-{HESAI},'' \url{https://www.hesaitech.com/en/Pandora}.

\bibitem{anand2013contextually}
A.~Anand, H.~S. Koppula, T.~Joachims, and A.~Saxena, ``Contextually guided
  semantic labeling and search for three-dimensional point clouds,'' \emph{The
  International Journal of Robotics Research}, vol.~32, no.~1, pp. 19--34,
  2013.

\bibitem{wolf2015fast}
D.~Wolf, J.~Prankl, and M.~Vincze, ``Fast semantic segmentation of {3D} point
  clouds using a dense {CRF} with learned parameters,'' in \emph{IEEE
  International Conference on Robotics and Automation}.\hskip 1em plus 0.5em
  minus 0.4em\relax IEEE, 2015, pp. 4867--4873.

\bibitem{golovinskiy2009shape}
A.~Golovinskiy, V.~G. Kim, and T.~Funkhouser, ``Shape-based recognition of 3d
  point clouds in urban environments,'' in \emph{IEEE International Conference
  on Computer Vision}.\hskip 1em plus 0.5em minus 0.4em\relax IEEE, 2009, pp.
  2154--2161.

\bibitem{hackel2016fast}
T.~Hackel, J.~D. Wegner, and K.~Schindler, ``Fast semantic segmentation of 3d
  point clouds with strongly varying density,'' \emph{ISPRS annals of the
  photogrammetry, remote sensing and spatial information sciences}, vol.~3, pp.
  177--184, 2016.

\bibitem{weinmann2015semantic}
M.~Weinmann, B.~Jutzi, S.~Hinz, and C.~Mallet, ``Semantic point cloud
  interpretation based on optimal neighborhoods, relevant features and
  efficient classifiers,'' \emph{ISPRS Journal of Photogrammetry and Remote
  Sensing}, vol. 105, pp. 286--304, 2015.

\bibitem{anguelov2005discriminative}
D.~Anguelov, B.~Taskarf, V.~Chatalbashev, D.~Koller, D.~Gupta, G.~Heitz, and
  A.~Ng, ``Discriminative learning of markov random fields for segmentation of
  {3D} scan data,'' in \emph{IEEE Conference on Computer Vision and Pattern
  Recognition}, vol.~2.\hskip 1em plus 0.5em minus 0.4em\relax IEEE, 2005, pp.
  169--176.

\bibitem{triebel2006robust}
R.~Triebel, K.~Kersting, and W.~Burgard, ``Robust {3D} scan point
  classification using associative markov networks,'' in \emph{IEEE
  International Conference on Robotics and Automation}.\hskip 1em plus 0.5em
  minus 0.4em\relax IEEE, 2006, pp. 2603--2608.

\bibitem{qi2017pointnet++}
C.~R. Qi, L.~Yi, H.~Su, and L.~J. Guibas, ``{PointNet++}: Deep hierarchical
  feature learning on point sets in a metric space,'' in \emph{Advances in
  Neural Information Processing Systems}, 2017, pp. 5099--5108.

\bibitem{engelmann2017exploring}
F.~Engelmann, T.~Kontogianni, A.~Hermans, and B.~Leibe, ``Exploring spatial
  context for {3D} semantic segmentation of point clouds,'' in \emph{IEEE
  International Conference on Computer Vision Workshops}, 2017, pp. 716--724.

\bibitem{jiang2018pointsift}
M.~Jiang, Y.~Wu, T.~Zhao, Z.~Zhao, and C.~Lu, ``{PointSIFT}: A {SIFT}-like
  network module for {3D} point cloud semantic segmentation,'' \emph{arXiv
  preprint arXiv:1807.00652}, 2018.

\bibitem{engelmann2018know}
F.~Engelmann, T.~Kontogianni, J.~Schult, and B.~Leibe, ``Know what your
  neighbors do: {3D} semantic segmentation of point clouds,'' in \emph{European
  Conference on Computer Vision}, 2018, pp. 0--0.

\bibitem{li2018so}
J.~Li, B.~M. Chen, and G.~Hee~Lee, ``{SO-Net}: Self-organizing network for
  point cloud analysis,'' in \emph{IEEE Conference on Computer Vision and
  Pattern Recognition}, 2018, pp. 9397--9406.

\bibitem{yang2019modeling}
J.~Yang, Q.~Zhang, B.~Ni, L.~Li, J.~Liu, M.~Zhou, and Q.~Tian, ``Modeling point
  clouds with self-attention and gumbel subset sampling,'' in \emph{IEEE
  Conference on Computer Vision and Pattern Recognition}, 2019, pp. 3323--3332.

\bibitem{chen2019lsanet}
L.-Z. Chen, X.-Y. Li, D.-P. Fan, M.-M. Cheng, K.~Wang, and S.-P. Lu,
  ``{LSANet}: Feature learning on point sets by local spatial attention,''
  \emph{arXiv preprint arXiv:1905.05442}, 2019.

\bibitem{zhiheng2019pyramnet}
K.~Zhiheng and L.~Ning, ``{PyramNet}: Point cloud pyramid attention network and
  graph embedding module for classification and segmentation,'' \emph{arXiv
  preprint arXiv:1906.03299}, 2019.

\bibitem{li2018pointcnn}
Y.~Li, R.~Bu, M.~Sun, W.~Wu, X.~Di, and B.~Chen, ``{PointCNN}: Convolution on
  x-transformed points,'' in \emph{Advances in neural information processing
  systems}, 2018, pp. 820--830.

\bibitem{komarichev2019cnn}
A.~Komarichev, Z.~Zhong, and J.~Hua, ``{A-CNN}: Annularly convolutional neural
  networks on point clouds,'' in \emph{IEEE Conference on Computer Vision and
  Pattern Recognition}, 2019, pp. 7421--7430.

\bibitem{thomas2019kpconv}
H.~Thomas, C.~R. Qi, J.-E. Deschaud, B.~Marcotegui, F.~Goulette, and L.~J.
  Guibas, ``{KPConv}: Flexible and deformable convolution for point clouds,''
  in \emph{IEEE International Conference on Computer Vision}, 2019, pp.
  6411--6420.

\bibitem{engelmann2019dilated}
F.~Engelmann, T.~Kontogianni, and B.~Leibe, ``Dilated point convolutions: On
  the receptive field of point convolutions,'' \emph{arXiv preprint
  arXiv:1907.12046}, 2019.

\bibitem{pan2019pointatrousnet}
L.~Pan, P.~Wang, and C.-M. Chew, ``{PointAtrousNet}: Point atrous convolution
  for point cloud analysis,'' \emph{IEEE Robotics and Automation Letters},
  vol.~4, no.~4, pp. 4035--4041, 2019.

\bibitem{pan2019pointatrousgraph}
L.~Pan, C.-M. Chew, and G.~H. Lee, ``{PointAtrousGraph}: Deep hierarchical
  encoder-decoder with atrous convolution for point clouds,'' \emph{arXiv
  preprint arXiv:1907.09798}, 2019.

\bibitem{tatarchenko2018tangent}
M.~Tatarchenko, J.~Park, V.~Koltun, and Q.-Y. Zhou, ``Tangent convolutions for
  dense prediction in {3D},'' in \emph{IEEE Conference on Computer Vision and
  Pattern Recognition}, 2018, pp. 3887--3896.

\bibitem{zhao2019dar}
Z.~Zhao, M.~Liu, and K.~Ramani, ``{DAR-Net}: Dynamic aggregation network for
  semantic scene segmentation,'' \emph{arXiv preprint arXiv:1907.12022}, 2019.

\bibitem{zhang2019shellnet}
Z.~Zhang, B.-S. Hua, and S.-K. Yeung, ``{ShellNet}: Efficient point cloud
  convolutional neural networks using concentric shells statistics,'' in
  \emph{IEEE International Conference on Computer Vision}, 2019, pp.
  1607--1616.

\bibitem{liu2019point}
Z.~Liu, H.~Tang, Y.~Lin, and S.~Han, ``Point-voxel cnn for efficient {3D} deep
  learning,'' in \emph{Advances in Neural Information Processing Systems},
  2019, pp. 965--975.

\bibitem{huang2018recurrent}
Q.~Huang, W.~Wang, and U.~Neumann, ``Recurrent slice networks for {3D}
  segmentation of point clouds,'' in \emph{IEEE Conference on Computer Vision
  and Pattern Recognition}, 2018, pp. 2626--2635.

\bibitem{su2018splatnet}
H.~Su, V.~Jampani, D.~Sun, S.~Maji, E.~Kalogerakis, M.-H. Yang, and J.~Kautz,
  ``{SplatNet}: Sparse lattice networks for point cloud processing,'' in
  \emph{IEEE Conference on Computer Vision and Pattern Recognition}, 2018, pp.
  2530--2539.

\bibitem{jampani2016learning}
V.~Jampani, M.~Kiefel, and P.~V. Gehler, ``Learning sparse high dimensional
  filters: Image filtering, dense {CRFs} and bilateral neural networks,'' in
  \emph{IEEE Conference on Computer Vision and Pattern Recognition}, 2016, pp.
  4452--4461.

\bibitem{rosu2019latticenet}
R.~A. Rosu, P.~Sch{\"u}tt, J.~Quenzel, and S.~Behnke, ``{LatticeNet}: Fast
  point cloud segmentation using permutohedral lattices,'' \emph{arXiv preprint
  arXiv:1912.05905}, 2019.

\bibitem{ronneberger2015u}
O.~Ronneberger, P.~Fischer, and T.~Brox, ``{U-Net}: Convolutional networks for
  biomedical image segmentation,'' in \emph{International Conference on Medical
  Image Computing and Computer-Assisted Intervention}.\hskip 1em plus 0.5em
  minus 0.4em\relax Springer, 2015, pp. 234--241.

\bibitem{lawin2017deep}
F.~J. Lawin, M.~Danelljan, P.~Tosteberg, G.~Bhat, F.~S. Khan, and M.~Felsberg,
  ``Deep projective {3D} semantic segmentation,'' in \emph{International
  Conference on Computer Analysis of Images and Patterns}.\hskip 1em plus 0.5em
  minus 0.4em\relax Springer, 2017, pp. 95--107.

\bibitem{boulch2017unstructured}
A.~Boulch, B.~L. Saux, and N.~Audebert, ``Unstructured point cloud semantic
  labeling using deep segmentation networks,'' in \emph{Workshop on 3D Object
  Retrieval}, ser. 3DOR '17.\hskip 1em plus 0.5em minus 0.4em\relax Goslar,
  DEU: Eurographics Association, 2017, p. 17–24.

\bibitem{iandola2016squeezenet}
F.~N. Iandola, S.~Han, M.~W. Moskewicz, K.~Ashraf, W.~J. Dally, and K.~Keutzer,
  ``{SqueezeNet}: {AlexNet}-level accuracy with 50x fewer parameters and $<$
  0.5{MB} model size,'' \emph{arXiv preprint arXiv:1602.07360}, 2016.

\bibitem{wu2019squeezesegv2}
B.~Wu, X.~Zhou, S.~Zhao, X.~Yue, and K.~Keutzer, ``{SqueezeSegV2}: Improved
  model structure and unsupervised domain adaptation for road-object
  segmentation from a {LiDAR} point cloud,'' in \emph{International Conference
  on Robotics and Automation}.\hskip 1em plus 0.5em minus 0.4em\relax IEEE,
  2019, pp. 4376--4382.

\bibitem{xu2020squeezesegv3}
C.~Xu, B.~Wu, Z.~Wang, W.~Zhan, P.~Vajda, K.~Keutzer, and M.~Tomizuka,
  ``{SqueezeSegV3}: Spatially-adaptive convolution for efficient point-cloud
  segmentation,'' \emph{arXiv preprint arXiv:2004.01803}, 2020.

\bibitem{mei2019semantic}
J.~Mei, B.~Gao, D.~Xu, W.~Yao, X.~Zhao, and H.~Zhao, ``Semantic segmentation of
  {3D} {LiDAR} data in dynamic scene using semi-supervised learning,''
  \emph{IEEE Transactions on Intelligent Transportation Systems}, 2019.

\bibitem{milioto2019rangenet++}
A.~Milioto, I.~Vizzo, J.~Behley, and C.~Stachniss, ``{RangeNet++}: Fast and
  accurate {LiDAR} semantic segmentation,'' in \emph{IEEE/RSJ International
  Conference on Intelligent Robots and Systems}, 2019.

\bibitem{biasutti2019lu}
P.~Biasutti, V.~Lepetit, J.-F. Aujol, M.~Br{\'e}dif, and A.~Bugeau, ``{LU-Net}:
  An efficient network for {3D} {LiDAR} point cloud semantic segmentation based
  on end-to-end-learned {3D} features and {U-Net},'' in \emph{IEEE
  International Conference on Computer Vision Workshops}, 2019, pp. 0--0.

\bibitem{alonso20203d}
I.~Alonso, L.~Riazuelo, L.~Montesano, and A.~C. Murillo, ``{3D-MiniNet}:
  Learning a {2D} representation from point clouds for fast and efficient {3D}
  {LiDAR} semantic segmentation,'' \emph{arXiv preprint arXiv:2002.10893},
  2020.

\bibitem{dewan2019deeptemporalseg}
A.~Dewan and W.~Burgard, ``{DeepTemporalSeg}: Temporally consistent semantic
  segmentation of {3D} {LiDAR} scans,'' \emph{arXiv preprint arXiv:1906.06962},
  2019.

\bibitem{zhang2018liseg}
W.~Zhang, C.~Zhou, J.~Yang, and K.~Huang, ``{LiSeg}: Lightweight road-object
  semantic segmentation in {3D} {LiDAR} scans for autonomous driving,'' in
  \emph{IEEE Intelligent Vehicles Symposium}.\hskip 1em plus 0.5em minus
  0.4em\relax IEEE, 2018, pp. 1021--1026.

\bibitem{wang2018pointseg}
Y.~Wang, T.~Shi, P.~Yun, L.~Tai, and M.~Liu, ``{PointSeg}: Real-time semantic
  segmentation based on {3D} {LiDAR} point cloud,'' \emph{arXiv preprint
  arXiv:1807.06288}, 2018.

\bibitem{biasutti2019riu}
P.~Biasutti, A.~Bugeau, J.-F. Aujol, and M.~Br{\'e}dif, ``{RIU-Net}:
  Embarrassingly simple semantic segmentation of {3D} {LiDAR} point cloud,''
  \emph{arXiv preprint arXiv:1905.08748}, 2019.

\bibitem{aksoy2019salsanet}
E.~E. Aksoy, S.~Baci, and S.~Cavdar, ``{SalsaNet}: Fast road and vehicle
  segmentation in {LiDAR} point clouds for autonomous driving,'' \emph{arXiv
  preprint arXiv:1909.08291}, 2019.

\bibitem{cortinhal2020salsanext}
T.~Cortinhal, G.~Tzelepis, and E.~E. Aksoy, ``{SalsaNext}: Fast,
  uncertainty-aware semantic segmentation of {LiDAR} point clouds for
  autonomous driving,'' \emph{arXiv preprint arXiv:2003.03653}, 2020.

\bibitem{huang2016point}
J.~Huang and S.~You, ``Point cloud labeling using {3D} convolutional neural
  network,'' in \emph{International Conference on Pattern Recognition}.\hskip
  1em plus 0.5em minus 0.4em\relax IEEE, 2016, pp. 2670--2675.

\bibitem{tchapmi2017segcloud}
L.~Tchapmi, C.~Choy, I.~Armeni, J.~Gwak, and S.~Savarese, ``Segcloud: Semantic
  segmentation of {3D} point clouds,'' in \emph{2017 international conference
  on {3D} vision}.\hskip 1em plus 0.5em minus 0.4em\relax IEEE, 2017, pp.
  537--547.

\bibitem{rethage2018fully}
D.~Rethage, J.~Wald, J.~Sturm, N.~Navab, and F.~Tombari, ``Fully-convolutional
  point networks for large-scale point clouds,'' in \emph{European Conference
  on Computer Vision}, 2018, pp. 596--611.

\bibitem{liu20173dcnn}
F.~Liu, S.~Li, L.~Zhang, C.~Zhou, R.~Ye, Y.~Wang, and J.~Lu, ``{3DCNN-DQN-RNN}:
  A deep reinforcement learning framework for semantic parsing of large-scale
  {3D} point clouds,'' in \emph{IEEE International Conference on Computer
  Vision}, 2017, pp. 5678--5687.

\bibitem{graham20183d}
B.~Graham, M.~Engelcke, and L.~van~der Maaten, ``{3D} semantic segmentation
  with submanifold sparse convolutional networks,'' in \emph{IEEE Conference on
  Computer Vision and Pattern Recognition}, 2018, pp. 9224--9232.

\bibitem{zhang2018efficient}
C.~Zhang, W.~Luo, and R.~Urtasun, ``Efficient convolutions for real-time
  semantic segmentation of {3D} point clouds,'' in \emph{International
  Conference on {3D} Vision}.\hskip 1em plus 0.5em minus 0.4em\relax IEEE,
  2018, pp. 399--408.

\bibitem{meng2019vv}
H.-Y. Meng, L.~Gao, Y.-K. Lai, and D.~Manocha, ``{VV-Net}: Voxel vae net with
  group convolutions for point cloud segmentation,'' in \emph{IEEE
  International Conference on Computer Vision}, 2019, pp. 8500--8508.

\bibitem{radi2019volmap}
H.~Radi and W.~Ali, ``{VolMap}: A real-time model for semantic segmentation of
  a {LiDAR} surrounding view,'' \emph{arXiv preprint arXiv:1906.11873}, 2019.

\bibitem{landrieu2018large}
L.~Landrieu and M.~Simonovsky, ``Large-scale point cloud semantic segmentation
  with superpoint graphs,'' in \emph{IEEE Conference on Computer Vision and
  Pattern Recognition}, 2018, pp. 4558--4567.

\bibitem{wang2019graph}
L.~Wang, Y.~Huang, Y.~Hou, S.~Zhang, and J.~Shan, ``Graph attention convolution
  for point cloud semantic segmentation,'' in \emph{IEEE Conference on Computer
  Vision and Pattern Recognition}, 2019, pp. 10\,296--10\,305.

\bibitem{jiang2019hierarchical}
L.~Jiang, H.~Zhao, S.~Liu, X.~Shen, C.-W. Fu, and J.~Jia, ``Hierarchical
  point-edge interaction network for point cloud semantic segmentation,'' in
  \emph{IEEE International Conference on Computer Vision}, 2019, pp.
  10\,433--10\,441.

\bibitem{wang2019dynamic}
Y.~Wang, Y.~Sun, Z.~Liu, S.~E. Sarma, M.~M. Bronstein, and J.~M. Solomon,
  ``Dynamic graph {CNN} for learning on point clouds,'' \emph{ACM Transactions
  On Graphics (tog)}, vol.~38, no.~5, pp. 1--12, 2019.

\bibitem{papandreou2015weakly}
G.~Papandreou, L.-C. Chen, K.~P. Murphy, and A.~L. Yuille, ``Weakly-and
  semi-supervised learning of a deep convolutional network for semantic image
  segmentation,'' in \emph{IEEE International Conference on Computer Vision},
  2015, pp. 1742--1750.

\bibitem{pathak2014fully}
D.~Pathak, E.~Shelhamer, J.~Long, and T.~Darrell, ``Fully convolutional
  multi-class multiple instance learning,'' \emph{arXiv preprint
  arXiv:1412.7144}, 2014.

\bibitem{pathak2015constrained}
D.~Pathak, P.~Krahenbuhl, and T.~Darrell, ``Constrained convolutional neural
  networks for weakly supervised segmentation,'' in \emph{IEEE International
  Conference on Computer Vision}, 2015, pp. 1796--1804.

\bibitem{shimoda2016distinct}
W.~Shimoda and K.~Yanai, ``Distinct class-specific saliency maps for weakly
  supervised semantic segmentation,'' in \emph{European Conference on Computer
  Vision}.\hskip 1em plus 0.5em minus 0.4em\relax Springer, 2016, pp. 218--234.

\bibitem{kolesnikov2016seed}
A.~Kolesnikov and C.~H. Lampert, ``Seed, expand and constrain: Three principles
  for weakly-supervised image segmentation,'' in \emph{European Conference on
  Computer Vision}.\hskip 1em plus 0.5em minus 0.4em\relax Springer, 2016, pp.
  695--711.

\bibitem{pinheiro2015image}
P.~O. Pinheiro and R.~Collobert, ``From image-level to pixel-level labeling
  with convolutional networks,'' in \emph{IEEE Conference on Computer Vision
  and Pattern Recognition}, 2015, pp. 1713--1721.

\bibitem{kwak2017weakly}
S.~Kwak, S.~Hong, and B.~Han, ``Weakly supervised semantic segmentation using
  superpixel pooling network,'' in \emph{AAAI Conference on Artificial
  Intelligence}, 2017.

\bibitem{dai2015boxsup}
J.~Dai, K.~He, and J.~Sun, ``{BoxSup}: Exploiting bounding boxes to supervise
  convolutional networks for semantic segmentation,'' in \emph{IEEE
  International Conference on Computer Vision}, 2015, pp. 1635--1643.

\bibitem{lin2016scribblesup}
D.~Lin, J.~Dai, J.~Jia, K.~He, and J.~Sun, ``{ScribbleSup}: Scribble-supervised
  convolutional networks for semantic segmentation,'' in \emph{IEEE Conference
  on Computer Vision and Pattern Recognition}, 2016, pp. 3159--3167.

\bibitem{bearman2016s}
A.~Bearman, O.~Russakovsky, V.~Ferrari, and L.~Fei-Fei, ``What's the point:
  Semantic segmentation with point supervision,'' in \emph{European Conference
  on Computer Vision}.\hskip 1em plus 0.5em minus 0.4em\relax Springer, 2016,
  pp. 549--565.

\bibitem{hong2015decoupled}
S.~Hong, H.~Noh, and B.~Han, ``Decoupled deep neural network for
  semi-supervised semantic segmentation,'' in \emph{Advances in neural
  information processing systems}, 2015, pp. 1495--1503.

\bibitem{tarvainen2017mean}
A.~Tarvainen and H.~Valpola, ``Mean teachers are better role models:
  Weight-averaged consistency targets improve semi-supervised deep learning
  results,'' in \emph{Advances in neural information processing systems}, 2017,
  pp. 1195--1204.

\bibitem{vu2019advent}
T.-H. Vu, H.~Jain, M.~Bucher, M.~Cord, and P.~P{\'e}rez, ``{ADVENT}:
  Adversarial entropy minimization for domain adaptation in semantic
  segmentation,'' in \emph{IEEE Conference on Computer Vision and Pattern
  Recognition}, 2019, pp. 2517--2526.

\bibitem{zou2018unsupervised}
Y.~Zou, Z.~Yu, B.~Vijaya~Kumar, and J.~Wang, ``Unsupervised domain adaptation
  for semantic segmentation via class-balanced self-training,'' in
  \emph{European Conference on Computer Vision}, 2018, pp. 289--305.

\bibitem{hu2019learning}
A.~Hu, A.~Kendall, and R.~Cipolla, ``Learning a spatio-temporal embedding for
  video instance segmentation,'' \emph{arXiv preprint arXiv:1912.08969}, 2019.

\bibitem{he2017std2p}
Y.~He, W.-C. Chiu, M.~Keuper, and M.~Fritz, ``Std2p: {RGBD} semantic
  segmentation using spatio-temporal data-driven pooling,'' in \emph{IEEE
  Conference on Computer Vision and Pattern Recognition}, 2017, pp. 4837--4846.

\bibitem{barnes2017find}
D.~Barnes, W.~Maddern, and I.~Posner, ``Find your own way: Weakly-supervised
  segmentation of path proposals for urban autonomy,'' in \emph{IEEE
  International Conference on Robotics and Automation}.\hskip 1em plus 0.5em
  minus 0.4em\relax IEEE, 2017, pp. 203--210.

\bibitem{bruls2018mark}
T.~Bruls, W.~Maddern, A.~A. Morye, and P.~Newman, ``Mark yourself: Road marking
  segmentation via weakly-supervised annotations from multimodal data,'' in
  \emph{IEEE International Conference on Robotics and Automation}.\hskip 1em
  plus 0.5em minus 0.4em\relax IEEE, 2018, pp. 1863--1870.

\bibitem{Xu2020}
X.~Xu and G.~H. Lee, ``Weakly supervised semantic point cloud segmentation:
  Towards 10x fewer labels,'' in \emph{IEEE Conference on Computer Vision and
  Pattern Recognition}, 2020, pp. 13\,706--13\,715.

\bibitem{mei2019incorporating}
J.~Mei and H.~Zhao, ``Incorporating human domain knowledge in {3D}
  {LiDAR}-based semantic segmentation,'' \emph{IEEE Transactions on Intelligent
  Vehicles}, 2019.

\bibitem{xu2019semantic}
K.~Xu, Y.~Yao, K.~Murasaki, S.~Ando, and A.~Sagata, ``Semantic segmentation of
  sparsely annotated {3D} point clouds by pseudo-labelling,'' in
  \emph{International Conference on {3D} Vision}.\hskip 1em plus 0.5em minus
  0.4em\relax IEEE, 2019, pp. 463--471.

\bibitem{kim2018season}
D.-K. Kim, D.~Maturana, M.~Uenoyama, and S.~Scherer, ``Season-invariant
  semantic segmentation with a deep multimodal network,'' in \emph{Field and
  Service Robotics}.\hskip 1em plus 0.5em minus 0.4em\relax Springer, 2018, pp.
  255--270.

\bibitem{mei2018joint}
J.~Mei, L.~Zhang, Y.~Wang, Z.~Zhu, and H.~Ding, ``Joint margin, cograph, and
  label constraints for semisupervised scene parsing from point clouds,''
  \emph{IEEE Transactions on Geoscience and Remote Sensing}, vol.~56, no.~7,
  pp. 3800--3813, 2018.

\bibitem{noroozi2016unsupervised}
M.~Noroozi and P.~Favaro, ``Unsupervised learning of visual representations by
  solving jigsaw puzzles,'' in \emph{European Conference on Computer
  Vision}.\hskip 1em plus 0.5em minus 0.4em\relax Springer, 2016, pp. 69--84.

\bibitem{doersch2015unsupervised}
C.~Doersch, A.~Gupta, and A.~A. Efros, ``Unsupervised visual representation
  learning by context prediction,'' in \emph{IEEE International Conference on
  Computer Vision}, 2015, pp. 1422--1430.

\bibitem{pathak2016context}
D.~Pathak, P.~Krahenbuhl, J.~Donahue, T.~Darrell, and A.~A. Efros, ``Context
  encoders: Feature learning by inpainting,'' in \emph{IEEE Conference on
  Computer Vision and Pattern Recognition}, 2016, pp. 2536--2544.

\bibitem{zhang2016colorful}
R.~Zhang, P.~Isola, and A.~A. Efros, ``Colorful image colorization,'' in
  \emph{European Conference on Computer Vision}.\hskip 1em plus 0.5em minus
  0.4em\relax Springer, 2016, pp. 649--666.

\bibitem{wang2015unsupervised}
X.~Wang and A.~Gupta, ``Unsupervised learning of visual representations using
  videos,'' in \emph{IEEE International Conference on Computer Vision}, 2015,
  pp. 2794--2802.

\bibitem{pathak2017learning}
D.~Pathak, R.~Girshick, P.~Doll{\'a}r, T.~Darrell, and B.~Hariharan, ``Learning
  features by watching objects move,'' in \emph{IEEE Conference on Computer
  Vision and Pattern Recognition}, 2017, pp. 2701--2710.

\bibitem{noroozi2018boosting}
M.~Noroozi, A.~Vinjimoor, P.~Favaro, and H.~Pirsiavash, ``Boosting
  self-supervised learning via knowledge transfer,'' in \emph{IEEE Conference
  on Computer Vision and Pattern Recognition}, 2018, pp. 9359--9367.

\bibitem{sauder2019self}
J.~Sauder and B.~Sievers, ``Self-supervised deep learning on point clouds by
  reconstructing space,'' in \emph{Advances in Neural Information Processing
  Systems}, 2019, pp. 12\,942--12\,952.

\bibitem{maligo2016classification}
A.~Maligo and S.~Lacroix, ``Classification of outdoor {3D} {LiDAR} data based
  on unsupervised gaussian mixture models,'' \emph{IEEE Transactions on
  Automation Science and Engineering}, vol.~14, no.~1, pp. 5--16, 2016.

\bibitem{10.1007/978-3-030-01424-7_27}
C.~Tan, F.~Sun, T.~Kong, W.~Zhang, C.~Yang, and C.~Liu, ``A survey on deep
  transfer learning,'' in \emph{Artificial Neural Networks and Machine
  Learning}, V.~K{\r{u}}rkov{\'a}, Y.~Manolopoulos, B.~Hammer, L.~Iliadis, and
  I.~Maglogiannis, Eds.\hskip 1em plus 0.5em minus 0.4em\relax Cham: Springer
  International Publishing, 2018, pp. 270--279.

\bibitem{ganin2016domain}
Y.~Ganin, E.~Ustinova, H.~Ajakan, P.~Germain, H.~Larochelle, F.~Laviolette,
  M.~Marchand, and V.~Lempitsky, ``Domain-adversarial training of neural
  networks,'' \emph{The Journal of Machine Learning Research}, vol.~17, no.~1,
  pp. 2096--2030, 2016.

\bibitem{dai2007boosting}
W.~Dai, Q.~Yang, G.-R. Xue, and Y.~Yu, ``Boosting for transfer learning,'' in
  \emph{International Conference on Machine Learning}, 2007, pp. 193--200.

\bibitem{long2015learning}
M.~Long, Y.~Cao, J.~Wang, and M.~Jordan, ``Learning transferable features with
  deep adaptation networks,'' in \emph{International Conference on Machine
  Learning}.\hskip 1em plus 0.5em minus 0.4em\relax PMLR, 2015, pp. 97--105.

\bibitem{oquab2014learning}
M.~Oquab, L.~Bottou, I.~Laptev, and J.~Sivic, ``Learning and transferring
  mid-level image representations using convolutional neural networks,'' in
  \emph{IEEE Conference on Computer Vision and Pattern Recognition}, 2014, pp.
  1717--1724.

\bibitem{hong2016learning}
S.~Hong, J.~Oh, H.~Lee, and B.~Han, ``Learning transferrable knowledge for
  semantic segmentation with deep convolutional neural network,'' in \emph{IEEE
  Conference on Computer Vision and Pattern Recognition}, 2016, pp. 3204--3212.

\bibitem{abdou2019end}
M.~Abdou, M.~Elkhateeb, I.~Sobh, and A.~Elsallab, ``End-to-end {3D}-pointcloud
  semantic segmentation for autonomous driving,'' \emph{arXiv preprint
  arXiv:1906.10964}, 2019.

\bibitem{roddick2020predicting}
T.~Roddick and R.~Cipolla, ``Predicting semantic map representations from
  images using pyramid occupancy networks,'' in \emph{IEEE Conference on
  Computer Vision and Pattern Recognition}, 2020, pp. 11\,138--11\,147.

\bibitem{lu2019monocular}
C.~Lu, M.~J.~G. van~de Molengraft, and G.~Dubbelman, ``Monocular semantic
  occupancy grid mapping with convolutional variational encoder--decoder
  networks,'' \emph{IEEE Robotics and Automation Letters}, vol.~4, no.~2, pp.
  445--452, 2019.

\bibitem{erkent2018semantic}
{\"O}.~Erkent, C.~Wolf, C.~Laugier, D.~S. Gonz{\'a}lez, and V.~R. Cano,
  ``Semantic grid estimation with a hybrid bayesian and deep neural network
  approach,'' in \emph{IEEE/RSJ International Conference on Intelligent Robots
  and Systems}.\hskip 1em plus 0.5em minus 0.4em\relax IEEE, 2018, pp.
  888--895.

\bibitem{dequaire2018deep}
J.~Dequaire, P.~Ondr{\'u}{\v{s}}ka, D.~Rao, D.~Wang, and I.~Posner, ``Deep
  tracking in the wild: End-to-end tracking using recurrent neural networks,''
  \emph{The International Journal of Robotics Research}, vol.~37, no. 4-5, pp.
  492--512, 2018.

\bibitem{jaritz2019xmuda}
M.~Jaritz, T.-H. Vu, R.~de~Charette, {\'E}.~Wirbel, and P.~P{\'e}rez,
  ``{xMUDA}: Cross-modal unsupervised domain adaptation for {3D} semantic
  segmentation,'' \emph{arXiv preprint arXiv:1911.12676}, 2019.

\bibitem{shaban2017one}
A.~Shaban, S.~Bansal, Z.~Liu, I.~Essa, and B.~Boots, ``One-shot learning for
  semantic segmentation,'' \emph{arXiv preprint arXiv:1709.03410}, 2017.

\bibitem{zhang2019canet}
C.~Zhang, G.~Lin, F.~Liu, R.~Yao, and C.~Shen, ``{CANet}: Class-agnostic
  segmentation networks with iterative refinement and attentive few-shot
  learning,'' in \emph{IEEE Conference on Computer Vision and Pattern
  Recognition}, 2019, pp. 5217--5226.

\bibitem{snell2017prototypical}
J.~Snell, K.~Swersky, and R.~Zemel, ``Prototypical networks for few-shot
  learning,'' in \emph{Advances in neural information processing systems},
  2017, pp. 4077--4087.

\bibitem{dong2018few}
N.~Dong and E.~Xing, ``Few-shot semantic segmentation with prototype
  learning,'' in \emph{British Machine Vision Conference}, vol.~3, no.~4, 2018,
  p.~79.

\bibitem{tian2019differentiable}
P.~Tian, Z.~Wu, L.~Qi, L.~Wang, Y.~Shi, and Y.~Gao, ``Differentiable
  meta-learning model for few-shot semantic segmentation,'' \emph{arXiv
  preprint arXiv:1911.10371}, 2019.

\bibitem{saleh2016built}
F.~Saleh, M.~S. Aliakbarian, M.~Salzmann, L.~Petersson, S.~Gould, and J.~M.
  Alvarez, ``Built-in foreground/background prior for weakly-supervised
  semantic segmentation,'' in \emph{European Conference on Computer
  Vision}.\hskip 1em plus 0.5em minus 0.4em\relax Springer, 2016, pp. 413--432.

\bibitem{kolesnikov2016improving}
A.~Kolesnikov and C.~H. Lampert, ``Improving weakly-supervised object
  localization by micro-annotation,'' \emph{arXiv preprint arXiv:1605.05538},
  2016.

\bibitem{mackowiak2018cereals}
R.~Mackowiak, P.~Lenz, O.~Ghori, F.~Diego, O.~Lange, and C.~Rother,
  ``Cereals-cost-effective region-based active learning for semantic
  segmentation,'' \emph{arXiv preprint arXiv:1810.09726}, 2018.

\bibitem{li2016classification}
Z.~Li, L.~Zhang, R.~Zhong, T.~Fang, L.~Zhang, and Z.~Zhang, ``Classification of
  urban point clouds: A robust supervised approach with automatically
  generating training data,'' \emph{IEEE Journal of Selected Topics in Applied
  Earth Observations and Remote Sensing}, vol.~10, no.~3, pp. 1207--1220, 2016.

\bibitem{chen2019image}
Z.~Chen, Q.~Liao, Z.~Wang, Y.~Liu, and M.~Liu, ``Image detector based automatic
  {3D} data labeling and training for vehicle detection on point cloud,'' in
  \emph{2019 IEEE Intelligent Vehicles Symposium}.\hskip 1em plus 0.5em minus
  0.4em\relax IEEE, 2019, pp. 1408--1413.

\bibitem{piewak2018boosting}
F.~Piewak, P.~Pinggera, M.~Schafer, D.~Peter, B.~Schwarz, N.~Schneider,
  M.~Enzweiler, D.~Pfeiffer, and M.~Zollner, ``Boosting {LiDAR}-based semantic
  labeling by cross-modal training data generation,'' in \emph{European
  Conference on Computer Vision}, 2018, pp. 0--0.

\bibitem{varga2017super}
R.~Varga, A.~Costea, H.~Florea, I.~Giosan, and S.~Nedevschi, ``Super-sensor for
  360-degree environment perception: Point cloud segmentation using image
  features,'' in \emph{IEEE International Conference on Intelligent
  Transportation Systems}.\hskip 1em plus 0.5em minus 0.4em\relax IEEE, 2017,
  pp. 1--8.

\bibitem{luo2018semantic}
H.~Luo, C.~Wang, C.~Wen, Z.~Chen, D.~Zai, Y.~Yu, and J.~Li, ``Semantic labeling
  of mobile {LiDAR} point clouds via active learning and higher order {MRF},''
  \emph{IEEE Transactions on Geoscience and Remote Sensing}, vol.~56, no.~7,
  pp. 3631--3644, 2018.

\bibitem{richter2017playing}
S.~R. Richter, Z.~Hayder, and V.~Koltun, ``Playing for benchmarks,'' in
  \emph{IEEE International Conference on Computer Vision}, 2017, pp.
  2213--2222.

\bibitem{krahenbuhl2018free}
P.~Kr{\"a}henb{\"u}hl, ``Free supervision from video games,'' in \emph{IEEE
  Conference on Computer Vision and Pattern Recognition}, 2018, pp. 2955--2964.

\bibitem{ros2016synthia}
G.~Ros, L.~Sellart, J.~Materzynska, D.~Vazquez, and A.~M. Lopez, ``The
  {SYNTHIA} dataset: A large collection of synthetic images for semantic
  segmentation of urban scenes,'' in \emph{IEEE Conference on Computer Vision
  and Pattern Recognition}, 2016, pp. 3234--3243.

\bibitem{saleh2018effective}
F.~S. Saleh, M.~S. Aliakbarian, M.~Salzmann, L.~Petersson, and J.~M. Alvarez,
  ``Effective use of synthetic data for urban scene semantic segmentation,'' in
  \emph{European Conference on Computer Vision}.\hskip 1em plus 0.5em minus
  0.4em\relax Springer, 2018, pp. 86--103.

\bibitem{Unity3D}
J.~K. Haas, ``A history of the unity game engine,'' 2014.

\bibitem{wang2019automatic}
F.~Wang, Y.~Zhuang, H.~Gu, and H.~Hu, ``Automatic generation of synthetic
  {LiDAR} point clouds for 3-d data analysis,'' \emph{IEEE Transactions on
  Instrumentation and Measurement}, vol.~68, no.~7, pp. 2671--2673, 2019.

\bibitem{dosovitskiy2017carla}
A.~Dosovitskiy, G.~Ros, F.~Codevilla, A.~Lopez, and V.~Koltun, ``{CARLA}: An
  open urban driving simulator,'' \emph{arXiv preprint arXiv:1711.03938}, 2017.

\bibitem{fang2020augmented}
J.~Fang, D.~Zhou, F.~Yan, T.~Zhao, F.~Zhang, Y.~Ma, L.~Wang, and R.~Yang,
  ``Augmented {LiDAR} simulator for autonomous driving,'' \emph{IEEE Robotics
  and Automation Letters}, vol.~5, no.~2, pp. 1931--1938, 2020.

\bibitem{fang2018simulating}
J.~Fang, F.~Yan, T.~Zhao, F.~Zhang, D.~Zhou, R.~Yang, Y.~Ma, and L.~Wang,
  ``Simulating {LiDAR} point cloud for autonomous driving using real-world
  scenes and traffic flows,'' \emph{arXiv preprint arXiv:1811.07112}, vol.~1,
  2018.

\bibitem{garcia2018survey}
A.~Garcia-Garcia, S.~Orts-Escolano, S.~Oprea, V.~Villena-Martinez,
  P.~Martinez-Gonzalez, and J.~Garcia-Rodriguez, ``A survey on deep learning
  techniques for image and video semantic segmentation,'' \emph{Applied Soft
  Computing}, vol.~70, pp. 41--65, 2018.

\bibitem{bucher2019zero}
M.~Bucher, T.-H. Vu, M.~Cord, and P.~Pérez, ``Zero-shot semantic
  segmentation,'' \emph{arXiv preprint arXiv:1906.00817}, 2019.

\bibitem{wang2017learning}
Y.-X. Wang, D.~Ramanan, and M.~Hebert, ``Learning to model the tail,'' in
  \emph{Advances in Neural Information Processing Systems}, 2017, pp.
  7029--7039.

\bibitem{Zimmer2019}
W.~Zimmer, A.~Rangesh, and M.~Trivedi, ``{{3D} BAT: A Semi-Automatic, Web-based
  {3D} annotation toolbox for full-surround, multi-modal data streams},''
  \emph{IEEE Intelligent Vehicles Symposium, Proceedings}, vol. 2019-June, pp.
  1816--1821, 2019.

\bibitem{LIM2009701}
E.~H. Lim and D.~Suter, ``{3D} terrestrial {LiDAR} classifications with
  super-voxels and multi-scale conditional random fields,''
  \emph{Computer-Aided Design}, vol.~41, no.~10, pp. 701 -- 710, 2009.

\bibitem{Luo2018}
H.~Luo, C.~Wang, C.~Wen, Z.~Chen, D.~Zai, Y.~Yu, and J.~Li, ``Semantic labeling
  of mobile {LiDAR} point clouds via active learning and higher order {MRF},''
  \emph{IEEE Transactions on Geoscience and Remote Sensing}, vol.~56, no.~7,
  pp. 3631--3644, 2018.

\bibitem{Shorten2019}
C.~Shorten and T.~M. Khoshgoftaar, ``{A survey on Image Data Augmentation for
  Deep Learning},'' \emph{Journal of Big Data}, vol.~6, no.~1, 2019.

\bibitem{Cheng2020}
S.~Cheng, Z.~Leng, E.~D. Cubuk, B.~Zoph, C.~Bai, J.~Ngiam, Y.~Song, B.~Caine,
  V.~Vasudevan, C.~Li \emph{et~al.}, ``Improving {3D} object detection through
  progressive population based augmentation,'' \emph{arXiv preprint
  arXiv:2004.00831}, 2020.

\bibitem{Scheirer2013}
W.~J. Scheirer, A.~{De Rezende Rocha}, A.~Sapkota, and T.~E. Boult, ``{Toward
  open set recognition},'' \emph{IEEE Transactions on Pattern Analysis and
  Machine Intelligence}, vol.~35, no.~7, pp. 1757--1772, 2013.

\bibitem{khosla2012undoing}
A.~Khosla, T.~Zhou, T.~Malisiewicz, A.~A. Efros, and A.~Torralba, ``Undoing the
  damage of dataset bias,'' in \emph{European Conference on Computer
  Vision}.\hskip 1em plus 0.5em minus 0.4em\relax Springer, 2012, pp. 158--171.

\bibitem{tommasi2017deeper}
T.~Tommasi, N.~Patricia, B.~Caputo, and T.~Tuytelaars, ``A deeper look at
  dataset bias,'' in \emph{Domain Adaptation in Computer Vision
  Applications}.\hskip 1em plus 0.5em minus 0.4em\relax Springer, 2017, pp.
  37--55.

\end{thebibliography}

\vspace{-10 mm}
\begin{IEEEbiography}
	[{\vspace{-5 mm}\includegraphics[width=1in,height=1in,clip,keepaspectratio]{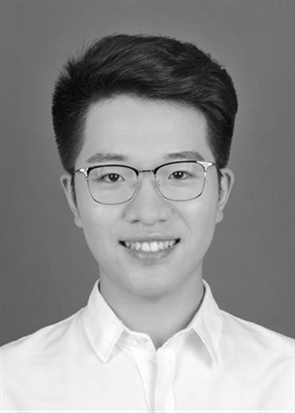}}]{Biao Gao}
	received B.S. degree in computer science (machine intelligence) from Peking University, Beijing, China, in 2017, where he is currently pursuing the Ph.D. degree with the Key Laboratory of Machine Perception (MOE), Peking University.
	His research interests include intelligent vehicles, 3D LiDAR perception, computer vision, and machine learning.
\end{IEEEbiography}
\vspace{-15 mm}
\begin{IEEEbiography}
	[{\vspace{-10 mm}\includegraphics[width=1in,height=1in,clip,keepaspectratio]{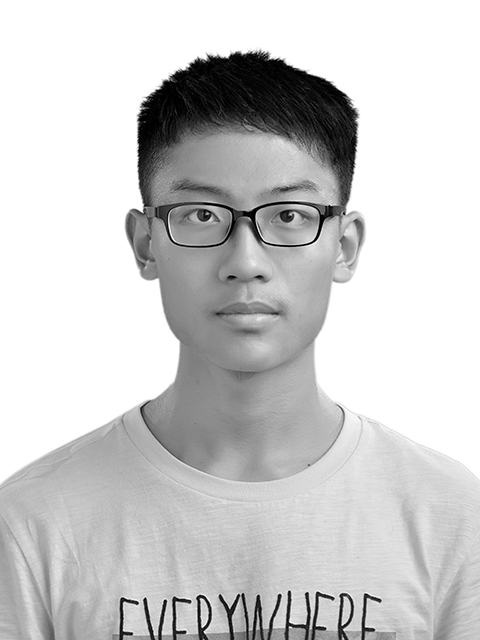}}]{Yancheng Pan}
	is currently pursuing the B.S. degree in computer science (machine intelligence) from Peking University, Beijing, China.
	He will continue pursuing the M.S. degree with the Key Laboratory of Machine Perception (MOE), Peking University.	
	His research interests include computer vision, 3D LiDAR perception and intelligent vehicles.
\end{IEEEbiography}
\vspace{-15 mm}
\begin{IEEEbiography}
	[{\vspace{-10 mm}\includegraphics[width=0.9in,height=0.9in,clip,keepaspectratio]{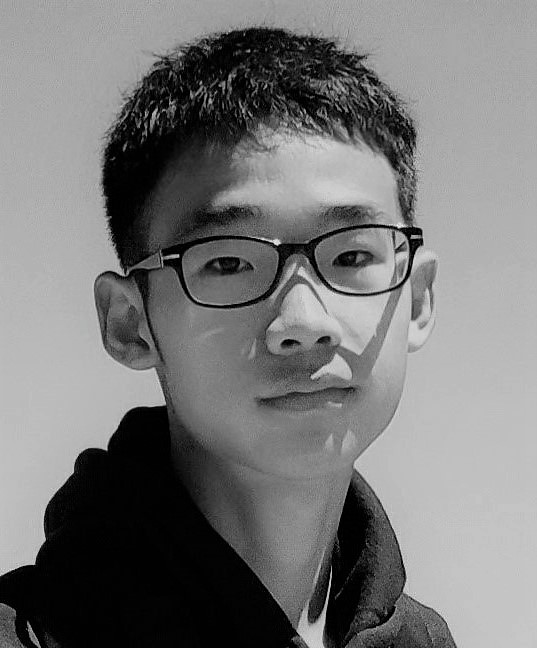}}]{Chengkun Li}
	is currently pursuing the B.S. degree in Automation from Beijing Institute of Technology, Beijing, China. He is presently an intern student in the Key Laboratory of Machine Perception (MOE), Peking University.	
	His research interests include computer vision, control and intelligent vehicles.
\end{IEEEbiography}
\vspace{-20 mm}
\begin{IEEEbiography}
	[{\vspace{-8 mm}\includegraphics[width=1in,height=1in,clip,keepaspectratio]{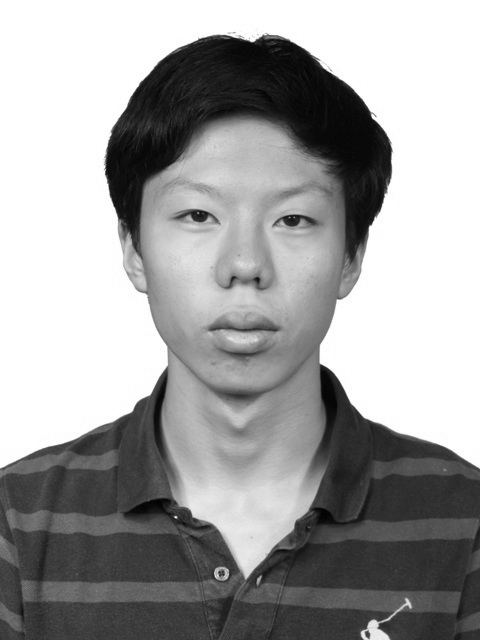}}]{Sibo Geng}
	is currently pursuing the B.S. degree in data science and big data from Yuanpei College, Peking University, Beijing, China. He is presently an intern student in the Key Laboratory of Machine Perception (MOE), Peking University, Beijing, China. His research interests include machine learning and intelligent vehicles.
\end{IEEEbiography}
\vspace{-20 mm}
\begin{IEEEbiography}
	[{\includegraphics[width=1in,height=1in,clip,keepaspectratio]{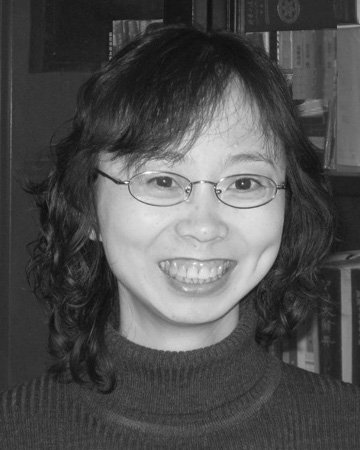}}]{Huijing Zhao}
	received B.S. degree in computer science in 1991 from
	Peking University, China. From 1991 to 1994, she was recruited by Peking
	University in a project of developing a GIS platform. She obtained M.E.
	degree in 1996 and Ph.D. degree in 1999 in civil engineering from the
	University of Tokyo, Japan. After post-doctoral research
	as the same university, in 2003, she was promoted to be a visiting
	associate professor in Center for Spatial Information Science, the
	University of Tokyo, Japan. In 2007, she joined Peking Univ as an associate
	professor at the School of Electronics Engineering and Computer Science.
	Her research interest covers intelligent vehicle, machine perception and mobile robot.
\end{IEEEbiography}

\end{document}